\documentclass{article} 
\usepackage{iclr2027_conference,times}

\iclrfinalcopy          

\usepackage{amsmath,amsfonts,bm}

\def\eqref#1{equation~\ref{#1}}

\def\1{\bm{1}}

\DeclareMathAlphabet{\mathsfit}{\encodingdefault}{\sfdefault}{m}{sl}
\SetMathAlphabet{\mathsfit}{bold}{\encodingdefault}{\sfdefault}{bx}{n}

\usepackage{hyperref}
\usepackage{url}
\usepackage{booktabs}
\usepackage{multirow}
\usepackage{graphicx}
\usepackage{array}
\usepackage{booktabs}
\usepackage{makecell}
\usepackage{amssymb}
\usepackage{graphicx}
\usepackage[table]{xcolor}

\usepackage[most]{tcolorbox}

\usepackage{seqsplit}

\usepackage{tabularx}
\usepackage{array}
\usepackage{booktabs}
\usepackage{makecell}

\usepackage{makecell}
\usepackage{pifont}
\usepackage[table]{xcolor}
\newcommand{\cmark}{\ding{51}}
\newcommand{\nomark}{--}

\usepackage[most]{tcolorbox}
\usepackage{listings}

\newtcblisting{promptbox}[1]{
    enhanced,
    breakable,
    colback=gray!3,
    colframe=gray!60,
    boxrule=0.6pt,
    arc=2pt,
    title={#1},
    fonttitle=\bfseries,
    listing only,
    listing engine=listings,
    listing options={
        basicstyle=\ttfamily\footnotesize,
        breaklines=true,
        breakatwhitespace=false,
        columns=fullflexible,
        keepspaces=true,
        showstringspaces=false,
        upquote=true,
        aboveskip=0pt,
        belowskip=0pt
    },
    left=6pt,
    right=6pt,
    top=5pt,
    bottom=5pt
}

\title{
\makebox[0pt][r]{%
    \raisebox{-0.7\height}[0pt][0pt]{%
        \includegraphics[height=1.9cm]{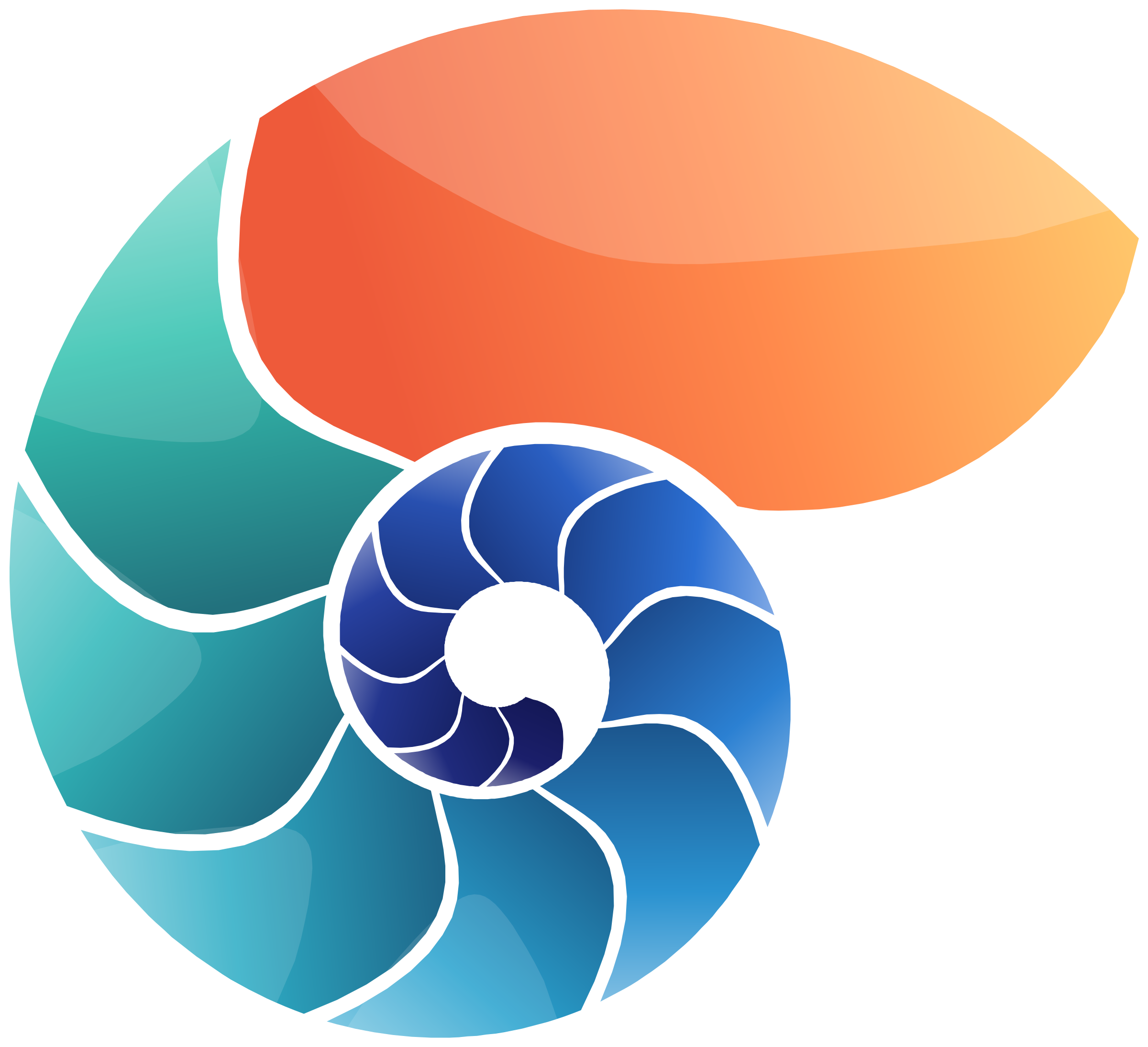}
    }\hspace{0.6cm}%
}%
Know What You Don’t Know: Agentic Data Synthesis by Evolving Generation Space for Closed-Loop Learning
}

\author{%
  \textbf{Wei Yang$^{1}$\thanks{Work was done during an internship at AWS AI.},
  Qing Ping$^{2}$, Youzhi Luo$^{2}$, Yawei Wang$^{2}$, Abhishek Kumar$^{2}$} \\
  \textbf{Panpan Xu$^{2}$, Jun Huan$^{2}$} \\
  $^{1}$University of Southern California \\
  $^{2}$AWS AI
}
\begin{document}

\maketitle
\lhead{Preprint}

\begin{abstract}
The continued improvement of large language models increasingly depends on synthetic experience, yet effective data synthesis for interactive agents requires not only executable and verifiable trajectories, but also interactions that provide learning value for the current task agent.
Existing methods increasingly incorporate environment feedback, data validation, or learner failures, but still face a \emph{feedback-to-decision gap}: observed weaknesses do not directly specify how future synthesis should change, or whether they can be represented by the current synthesis space.
To address this gap, we propose \textsc{EvoSyn}, a learning-feedback-driven multi-agent framework that organizes synthesis around a revisable state specifying \emph{what} interaction factors can be deliberately generated and \emph{where} synthesis effort should be allocated.
A role-specialized multi-agent system translates learner feedback into synthesis revisions, while a Designer allocates synthesis directions and a Planner grounds them into executable experience.
This process enables \emph{within-space adaptation} toward weak or under-covered regions and \emph{synthesis-space expansion} that promotes newly discovered factors into explicit generation controls.
Experiments on $\tau^2$-Bench and CAR-Bench show that \textsc{EvoSyn} improves average task success by $13.3$ points over the strongest synthesis baseline, while producing higher-quality and structurally broader synthetic experience and discovering controllable factors missed without learner feedback. 
\end{abstract}

\section{Introduction}

Synthetic data has become a central mechanism for pushing large language models beyond the limits of human annotation~\citep{wang2023self,xu2024wizardlm,shao2024deepseekmath}. 
In reasoning domains, a synthetic example can often be formulated as a self-contained problem with a verifiable answer or solution trace, allowing synthesis to focus on concepts or reasoning patterns the model has not yet mastered~\citep{gou2024tora,li2024numinamath}. 
Interactive agents impose an additional challenge~\citep{pahuja2025explorer,wang2026trajectory2task}. 
A valid agentic example must be feasible in an environment, executable through tools, consistent with domain policies, and verifiable through state changes or task outcomes~\citep{chen2026eigendata,wang2026trajectory2task}. 
Yet validity alone does not imply \textbf{learning utility}. 
An executable and correct task may simply repeat behavior the current agent already handles, while interactions that expose its remaining weaknesses remain absent from training~\citep{huang2025r,liang2026sws}. 
Thus, the central question in agentic data synthesis is not simply how to generate valid tasks, but \textit{what interactions are most useful for the current learner to experience next}.

Recent work has made substantial progress toward feedback-driven synthesis~\citep{chen2025scaling,liang2026sws}. 
Environment-grounded methods such as AutoPlay~\citep{ramrakhya2025scaling} and Trajectory2Task~\citep{wang2026trajectory2task} use interaction feedback to construct executable tasks; data-centric methods such as EigenData~\citep{chen2026eigendata} audit and repair generated samples; and learner-adaptive methods such as R-Zero~\citep{huang2025r} and Socratic-Zero~\citep{wang2025socratic} generate new problems from model failures or capability boundaries. 
These methods improve feasibility, data quality, and learner adaptivity, but using feedback is not the same as knowing what synthesis decision it should support. 
A common implicit assumption is that the learner's remaining weaknesses are already expressible within the current synthesis space, such as predefined task categories or interaction dimensions. 
Yet observed failures are not themselves actionable synthesis coordinates: they reveal where the learner struggles without necessarily identifying which interaction factors should be controlled in future generation~\citep{yang2026coevolve,kwan2026scope}. 
We refer to this missing bridge as the \textbf{feedback-to-decision gap}: translating learner feedback into actionable synthesis decisions while determining whether the current synthesis space can represent the learner's remaining weaknesses.

\begin{figure*}[t]
    \centering
    \includegraphics[width=\textwidth]{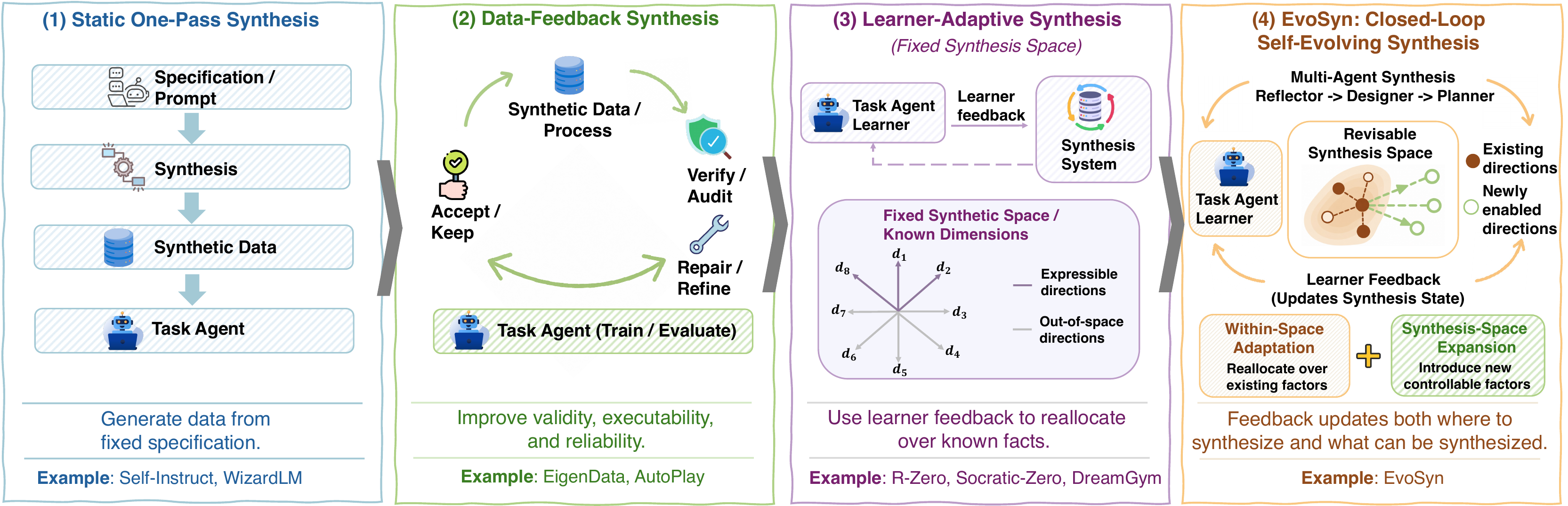}
    \caption{
From static synthesis to closed-loop self-evolving agentic synthesis.
Existing paradigms improve data quality or adapt within a fixed synthesis space. In contrast, \textsc{EvoSyn} uses learner feedback to revise both the allocation over known factors and the synthesis space itself by introducing new controllable interaction factors.
}
    \label{fig:intro_paradigm}
\end{figure*}

As illustrated in Figure~\ref{fig:intro_paradigm}, closing this gap requires the synthesis process itself to evolve together with the task agent~\citep{cheng2025evolving}. 
Learner feedback can expose two fundamentally different forms of mismatch. 
When the learner's weakness is already expressible in the current synthesis space, the problem is one of \textbf{within-space adaptation}: synthesis effort should be reallocated toward under-covered or challenging regions~\citep{yue2026dr}. 
When the weakness is not captured by existing synthesis coordinates, reallocation alone is insufficient; the system must instead perform \textbf{synthesis-space expansion}, promoting previously unrepresented but actionable interaction factors into explicit controls for future generation~\citep{li2026less}. 
For example, failures on refund or flight-change tasks may actually stem from numerical computation, policy-dependent action ordering, or the coordination of multiple state-changing operations. 
If such factors are absent from the synthesis space, generating more examples from the same task categories may increase data volume without addressing the underlying capability gap. 
Self-evolving agentic synthesis should therefore adapt not only the distribution over what can already be generated, but also the representation of what can be generated at all.

To address these challenges, we propose \textsc{EvoSyn}, a learning-feedback-driven multi-agent framework for self-evolving agentic data synthesis. 
\textsc{EvoSyn} organizes generation around a structured and revisable synthesis space over controllable interaction factors, which serves both as a coordinate system for learner diagnosis and as a control surface for future generation. 
Within this space, a Designer Agent selects synthesis directions and allocates generation budget, while a Planner Agent grounds these directions into environment-consistent task blueprints. 
A Generator Agent and rollout teacher then instantiate executable tasks and successful trajectories, which are further filtered by a Verifier Agent to ensure valid supervision. 
After task-agent training, a Reflector Agent analyzes performance, coverage, and failure signals and maps them back to the synthesis space. 
This feedback drives either \textbf{within-space adaptation}, which reallocates synthesis effort over existing coordinates, or \textbf{synthesis-space expansion}, which introduces new evidence-grounded and actionable factors as future generation controls. 
Through this closed loop, \textsc{EvoSyn} enables the task agent and its synthesis space to co-evolve through learning feedback.

Our main contributions are as follows:
\begin{itemize}
    \item We identify a \textbf{feedback-to-decision gap} in self-evolving agentic data synthesis and formulate self-evolving synthesis over a \textbf{revisable synthesis space} that adapts with the learner.

    \item We propose \textsc{EvoSyn}, which combines \textbf{within-space adaptation} to reallocate synthesis effort over existing factors with \textbf{synthesis-space expansion} to introduce new actionable factors revealed by learner feedback.

   \item Experiments on $\tau^2$-Bench and CAR-Bench improve average task success by $13.3$ points over the strongest synthesis baseline, while analyses show that learner feedback drives allocation shifts and reveals useful synthesis factors missed without it.
\end{itemize}

\section{Related Work}
\label{sec:related_work}

\noindent\textbf{Synthetic Data for Self-Evolution.}
Synthetic data has become increasingly important for LLM post-training, as it enables models to improve beyond the scale and coverage of manually annotated supervision.
Prior work in this area can be viewed along a progression from offline data construction to learner-aware synthesis.
Early methods generate instruction-following or reasoning corpora from seed demonstrations, teacher models, unlabeled corpora, or problem transformations, including instruction synthesis, instruction evolution, and mathematical data augmentation~\citep{wang2023self,xu2024wizardlm,li2023self,yu2023metamath,luo2025wizardmath}.
Another line of work emphasizes verifiability and data quality, using answer checkers, tool-integrated traces, filtering, or curated large-scale reasoning pipelines to ensure that generated supervision can reliably support post-training~\citep{zelikman2022star,gou2024tora,shao2024deepseekmath,li2024numinamath}.
More recent self-evolving methods place data generation and model learning in a feedback loop, where solver failures, teacher diagnoses, or capability boundaries are used to generate increasingly challenging tasks~\citep{huang2025r,wang2025socratic,yue2026dr,kwan2026scope}.

\noindent\textbf{Feedback-Driven Agentic Data Synthesis.}
Synthetic data for interactive agents introduces additional requirements beyond problem--answer correctness: tasks must be feasible in an environment, executable through tools, consistent with domain policies, and verifiable through state changes or task outcomes.
Existing work addresses these requirements through several complementary forms of feedback.
Environment-grounded methods use exploration, state replay, or valid trajectories to construct executable tasks and broaden interaction coverage~\citep{pahuja2025explorer,sun2025genesis,ramrakhya2025scaling,wang2026trajectory2task,wang2026webaggregator}.
Data-centric methods use checkers, validators, rejection reasons, or learned experience models to improve the validity and quality of tool-use or multi-turn synthetic data~\citep{chen2026eigendata,chen2025scaling}.
A further group of agent-evolution frameworks studies how environments, curricula, workflows, memories, or tool-use experiences can be revised over time~\citep{yang2026coevolve,dong2026agent,dai2026sead,zhai2025agentevolver,wu2025evolver,zhang2026evofsm}.
These directions collectively establish the importance of feedback in agentic data generation, while leaving open how different feedback signals should be converted into reusable synthesis decisions rather than only valid instances, repaired samples, or broader trajectory collections.
We provide a more complete discussion of related work in Appendix~\ref{app:related_work_comparison} and~\ref{app:related_work}.

\section{Problem Formulation}
\label{sec:problem}

We study synthetic data generation for improving language-model agents in interactive tool-use environments.
Let $\mathcal{E}$ denote an environment with domain policies, tool interfaces, state constraints, and an automatic evaluator, and let $\pi_{\theta_r}$ denote the task agent at synthesis round $r$.
At each round, the synthesis process generates a verified dataset $\mathcal{D}_r$ of executable and successful task--trajectory pairs, which is used to update the task agent:
\begin{equation}
\pi_{\theta_{r+1}}
=
\operatorname{Train}
\left(
\pi_{\theta_r},
\mathcal{D}_{\leq r}
\right),
\qquad
\mathcal{D}_{\leq r}
=
\bigcup_{t=1}^{r}\mathcal{D}_t .
\label{eq:agent_update}
\end{equation}
Unlike one-pass synthesis, however, the utility of $\mathcal{D}_r$ depends on the current learner: valid data may provide little learning signal if it does not target the interaction patterns where the agent remains weak. We therefore characterize synthesis at round $r$ by a \emph{revisable synthesis state} $S_r=(\mathcal{O}_r,\alpha_r)$,
where the synthesis space $\mathcal{O}_r$ specifies \emph{what interaction properties can be deliberately synthesized}, and $\alpha_r$ specifies \emph{where synthesis effort is allocated within that space}.
After training and evaluation, learner feedback $F_r$ must be translated into a new synthesis state,
$
S_{r+1}
=
\Psi(S_r,F_r)
=
(\mathcal{O}_{r+1},\alpha_{r+1}),
\label{eq:synthesis_revision}
$
where $\Psi$ denotes the feedback-driven synthesis revision process.
This mapping captures the feedback-to-decision gap: observed learner behavior must be converted into actionable decisions. A complete formulation is provided in Appendix~\ref{app:problem}.

\section{Method}
\label{sec:method}

\begin{figure*}[t]
    \centering
    \includegraphics[width=0.98\textwidth]{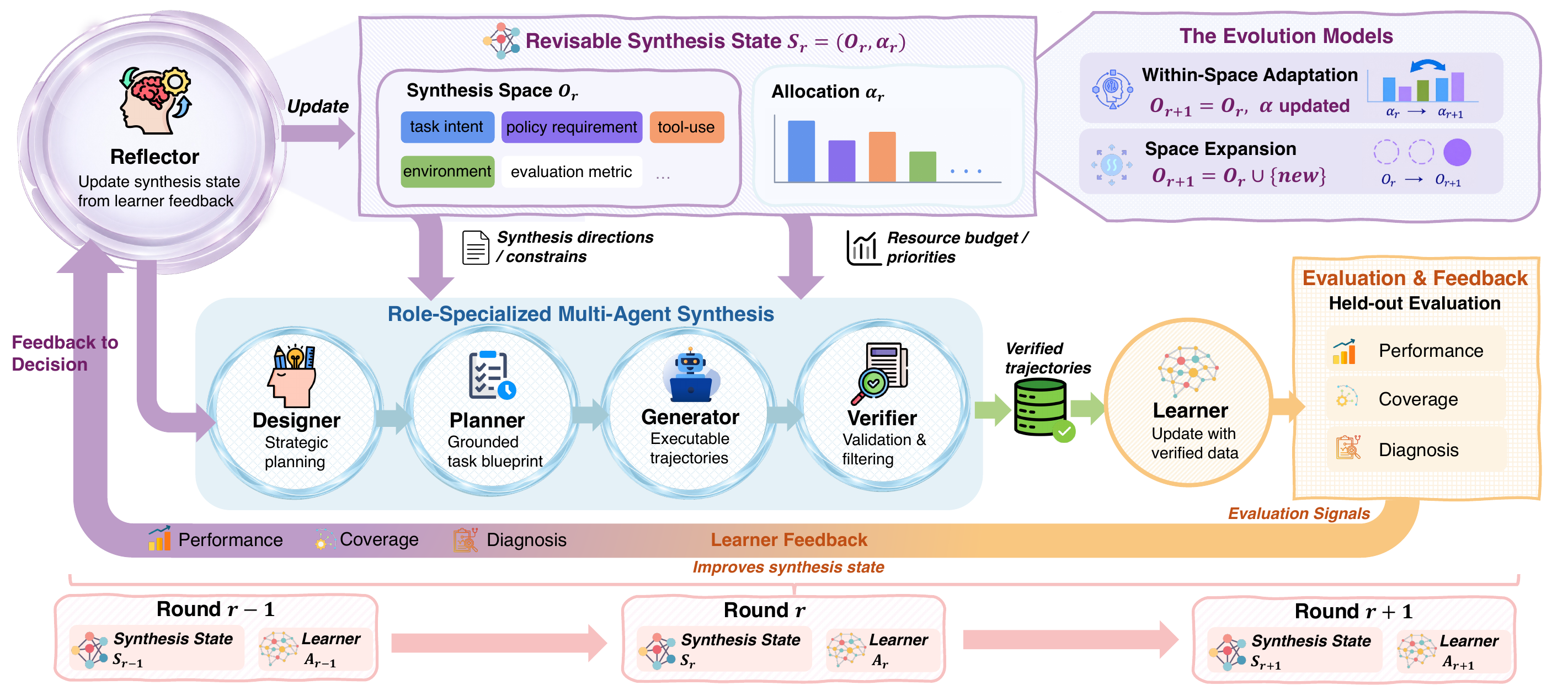}
\caption{
Overview of \textsc{EvoSyn}. Each round instantiates a closed-loop synthesis--learning process in which a revisable synthesis state mediates learner diagnosis and generation control, role-specialized agents realize verified experience, and feedback drives within-space adaptation or synthesis-space expansion to co-evolve synthesis with task-agent learning.
}
    \label{fig:method_overview}
\end{figure*}

\textsc{EvoSyn} is a multi-agent framework for learning-guided, self-evolving agentic data synthesis. As illustrated in Figure~\ref{fig:method_overview}, it maintains a revisable synthesis state that guides grounded data generation and is updated from learner feedback, forming a closed synthesis--learning loop. Additional details of the complete closed-loop framework are provided in Appendix~\ref{app:method_overview_loop}.

\subsection{Revisable Environment-Grounded Synthesis Space}
\label{sec:synthesis_space}

The synthesis state $S_r=(\mathcal{O}_r,\alpha_r)$ requires a structured representation of what experience can be deliberately generated.
Because agentic tasks must respect tool preconditions, domain policies, environment states, and evaluator requirements, \textsc{EvoSyn} instantiates $\mathcal{O}_r$ as a \emph{revisable environment-grounded synthesis space} over controllable interaction factors. Formally,
\begin{equation}
\mathcal{O}_r
=
\left\{
d_1^r,\ldots,d_{m_r}^r
\right\},
\qquad
d_j^r
=
\left(
k_j^r,
V_j^r,
c_j^r
\right),
\label{eq:synthesis_space}
\end{equation}
where $k_j^r$ denotes the factor name, $V_j^r$ its finite set of admissible values, and $c_j^r$ specifies how the factor constrains or guides generation.
Typical factors include task intent, policy requirements, tool-use patterns, expected outcomes, and execution complexity, while $\mathcal{O}_1$ is initialized from environment priors such as tool schemas, domain policies, state constraints, task specifications, and evaluator requirements.

A \emph{synthesis direction} specifies a controllable region of this space:
\begin{equation}
z
=
\left\{
k_j^r=v_j
\right\}_{j\in\mathcal{I}},
\qquad
v_j\in V_j^r,
\label{eq:synthesis_direction}
\end{equation}
where $\mathcal{I}\subseteq\{1,\ldots,m_r\}$ and
$z\in\mathcal{Z}(\mathcal{O}_r)$ represents a high-level interaction target rather than a concrete task. The Designer operates over such directions when allocating synthesis effort, while the Planner grounds each selected direction into an environment-consistent task blueprint.

Importantly, $\mathcal{O}_r$ serves as a shared representation for both \emph{learner diagnosis} and \emph{generation control}: its coordinates provide an interpretable basis for measuring learner performance and data coverage, while also specifying interaction properties that can be deliberately targeted in future synthesis. The space is therefore revisable: when current coordinates are sufficient, \textsc{EvoSyn} updates $\alpha_r$ to redistribute synthesis effort; when learner feedback reveals a capability-relevant property outside the current representation, it expands $\mathcal{O}_r$ with a new controllable factor. 

\subsection{Feedback-Guided Synthesis Evolution}
\label{sec:synthesis_evolution}

The role of learning feedback in \textsc{EvoSyn} is not merely to identify failed examples, but to determine how the synthesis process itself should change.
After each round of task-agent training, we summarize learner behavior and synthesized experience as
\begin{equation}
F_r
=
\operatorname{Feedback}
\left(
\pi_{\theta_{r+1}},
\mathcal{D}_{\leq r},
\mathcal{O}_r,
\mathcal{E}
\right),
\label{eq:feedback_report}
\end{equation}
where $F_r$ combines \emph{performance}, \emph{coverage}, and \emph{failure-diagnostic} signals.
Performance identifies weak interaction regions, coverage reveals whether these regions have received sufficient synthesis effort, and failure diagnostics capture recurrent patterns that may not yet correspond to existing synthesis coordinates. A Reflector Agent converts this evidence into the next synthesis state:
\begin{equation}
\left(
\mathcal{O}_{r+1},
\alpha_{r+1}
\right)
=
\operatorname{Reflector}
\left(
\mathcal{O}_r,
\alpha_r,
F_r
\right).
\label{eq:reflector_update}
\end{equation}
Rather than repairing individual failures, the Reflector operates at the synthesis level, deciding whether the observed weakness can be addressed within the current synthesis space or requires revising the space itself.

\paragraph{Within-Space Adaptation.}
When the learner's weakness is already expressible by factors in $\mathcal{O}_r$, \textsc{EvoSyn} keeps the synthesis space fixed and updates only the allocation:
\begin{equation}
\mathcal{O}_{r+1}
=
\mathcal{O}_r,
\qquad
\alpha_{r+1}
=
\operatorname{Reallocate}
\left(
\alpha_r,
F_r;
\mathcal{O}_r
\right).
\label{eq:within_space_adaptation}
\end{equation}
The Reflector shifts synthesis effort toward directions with poor performance or insufficient coverage, while reducing effort on well-covered or saturated regions.
The resulting allocation is then realized by the Designer Agent in the next synthesis round.

\paragraph{Synthesis-Space Expansion.}
When recurrent failures reveal a capability-relevant property that is not adequately represented in $\mathcal{O}_r$, reallocation alone is insufficient.
The Reflector instead identifies candidate interaction factors from diagnostic patterns in learner failures and synthesized trajectories.
A factor is promoted only if it is \emph{evidence-grounded}, \emph{interpretable and finite-valued}, \emph{non-redundant} with existing coordinates, and \emph{actionable} by the downstream synthesis process.
For an accepted factor
$d_{\mathrm{new}}=(k_{\mathrm{new}},V_{\mathrm{new}},c_{\mathrm{new}})$,
the synthesis space is expanded as
\begin{equation}
\mathcal{O}_{r+1}
=
\mathcal{O}_r
\cup
\left\{
d_{\mathrm{new}}
\right\}.
\label{eq:space_expansion_method}
\end{equation}
The new factor becomes an explicit synthesis coordinate that can be measured, budgeted for, and deliberately instantiated in subsequent rounds.

Together, these two updates allow \textsc{EvoSyn} to adapt both \emph{where} synthesis effort is placed and, when necessary, \emph{what} the synthesis process can explicitly represent.

\subsection{Role-Specialized Multi-Agent Synthesis} 
\label{sec:grounded_synthesis}

The synthesis state $S_r=(\mathcal{O}_r,\alpha_r)$ specifies what interaction factors should be targeted and how synthesis effort should be distributed, but it does not directly define executable training examples.
\textsc{EvoSyn} therefore uses a role-specialized multi-agent process to progressively translate these high-level synthesis decisions into environment-grounded and verified experience.

\paragraph{Designer: Strategic Synthesis Planning.}
The Designer Agent operates over the synthesis space rather than individual examples.
Given $\mathcal{O}_r$ and $\alpha_r$, it constructs a generation plan
\begin{equation}
\mathcal{G}_r
=
\left\{
(z_i,b_i)
\right\}_{i=1}^{K}
=
\operatorname{Designer}
\left(
\mathcal{O}_r,\alpha_r
\right),
\label{eq:designer}
\end{equation}
where $z_i\in\mathcal{Z}(\mathcal{O}_r)$ denotes a synthesis direction and $b_i$ its allocated budget.
The Designer thus operationalizes the revised state into decisions about \emph{what regions to target} and \emph{how much experience to generate}, separating round-level synthesis planning from the Reflector's meta-level revision.

\paragraph{Planner: Grounding Synthesis Intent.}
A synthesis direction specifies desired interaction properties but is not itself an executable task.
The Planner Agent bridges this abstraction gap by producing an environment-grounded blueprint
\begin{equation}
q_i
=
\operatorname{Planner}
\left(
z_i,\mathcal{O}_r,\mathcal{E}
\right),
\label{eq:planner}
\end{equation}
which specifies the user goal, required environment conditions, relevant policy constraints, expected tool interactions, and target outcome.
The Planner therefore serves as a grounding layer between synthesis-level intent and environment-level task realization.

\paragraph{Executable Realization and Verification.}
Given $q_i$, the Generator Agent instantiates a concrete task
$x_i=\operatorname{Generator}(q_i,\mathcal{E})$, and a rollout teacher interacts with the environment to produce a candidate successful trajectory
$\tau_i=\operatorname{RolloutTeacher}(x_i,\mathcal{E})$.
A Verifier Agent then checks executability, tool-call validity, policy compliance, state consistency, and task success, retaining only valid task--trajectory pairs:
\begin{equation}
\mathcal{D}_r
=
\left\{
(x_i,\tau_i)
\mid
\operatorname{Verifier}(x_i,\tau_i,\mathcal{E})=1
\right\}.
\label{eq:verified_synthesis}
\end{equation}

This role specialization forms a coarse-to-fine synthesis process: the Designer performs strategic allocation over the controllable space, the Planner grounds abstract targets into feasible task structures, and the Generator, rollout teacher, and Verifier realize and validate executable experience.

\subsection{Task-Agent Learning and Closing the Loop}
\label{sec:task_agent_learning}

The verified task--trajectory pairs $\mathcal{D}_r$ are converted into supervision for the downstream task agent while preserving user messages, tool calls, environment observations, and target responses.
To incorporate newly synthesized experience without discarding previously covered interactions, \textsc{EvoSyn} trains on accumulated verified data
$\mathcal{T}_{\leq r}=\bigcup_{t=1}^{r}\mathcal{T}_t$
using supervised fine-tuning:
\begin{equation}
\theta_{r+1}
=
\arg\min_{\theta}
\sum_{(s,a)\in\mathcal{T}_{\leq r}}
-\log \pi_{\theta}(a\mid s),
\label{eq:task_agent_sft}
\end{equation}
where $s$ denotes the interaction context and $a$ the target response or tool action.

The updated task agent $\pi_{\theta_{r+1}}$ is then evaluated on a held-out diagnostic set in the target environment.
Its performance, together with synthesis coverage and failure diagnostics, forms the feedback $F_r$ described in Section~\ref{sec:synthesis_evolution}, which the Reflector uses to revise the next synthesis state $S_{r+1}=(\mathcal{O}_{r+1},\alpha_{r+1})$.
The resulting closed loop is
$
S_r
\rightarrow
\mathcal{D}_r
\rightarrow
\pi_{\theta_{r+1}}
\rightarrow
F_r
\rightarrow
S_{r+1}.
\label{eq:evosyn_closed_loop}
$
Thus, each round updates the learner through synthesized experience while using the learner's resulting behavior to reshape future synthesis, enabling the task agent and synthesis process to co-evolve over rounds.

\begin{table*}[t]
\centering
\scriptsize
\setlength{\tabcolsep}{4.0pt}
\renewcommand{\arraystretch}{0.92}
\setlength{\aboverulesep}{1.5pt}
\setlength{\belowrulesep}{1.5pt}
\setlength{\extrarowheight}{0pt}
\definecolor{evoblue}{RGB}{232,242,255}

\caption{
Overall task success rate (\%) on agentic benchmarks.
For a consistent comparison across synthesis approaches, all synthesis methods use Qwen3-8B as the downstream task agent.
\textsc{EvoSyn}-R1 uses first-round data, while \textsc{EvoSyn}-Final-SFT uses accumulated multi-round data.
}
\label{tab:main_results}

\resizebox{0.92\textwidth}{!}{%
\begin{tabular}{llccccc}
\toprule
\multirow{2}{*}{\textbf{Method}}
& \multirow{2}{*}{\textbf{Synthesis LLM}}
& \multicolumn{3}{c}{\textbf{$\tau^2$-Bench}}
& \multirow{2}{*}{\textbf{CAR-Bench}}
& \multirow{2}{*}{\textbf{Avg.}} \\
\cmidrule(lr){3-5}
&
& \textbf{Airline}
& \textbf{Retail}
& \textbf{Telecom}
&
& \\
\midrule

\multicolumn{7}{l}{\emph{Frontier Closed-source Models}} \\[-1pt]
Claude Opus 4.5 & -- & 79.3 & 91.8 & 88.0 & 82.0 & 85.3 \\
GPT-5.2         & -- & 83.0 & 81.6 & 89.7 & 74.0 & 82.1 \\
Gemini-3 Pro    & -- & 80.5 & 75.9 & 91.0 & 67.3 & 78.7 \\

\midrule
\multicolumn{7}{l}{\emph{Open-Source Foundation Models}} \\[-1pt]
Qwen3-8B   & -- & 37.3 & 40.1 & 40.3 & 40.0 & 39.4 \\
Qwen3-30B  & -- & 62.0 & 75.4 & 41.8 & 49.3 & 57.1 \\
Qwen3-235B & -- & 64.7 & 83.3 & 65.8 & 65.3 & 69.8 \\

\midrule
\multicolumn{7}{l}{\emph{General Agentic Synthesis Baselines}} \\[-1pt]
AgentSkills     & GPT-5.2      & 35.3 & 64.3 & 61.4 & 51.3 & 53.0 \\
EnvScaler       & GPT-4.1      & 36.0 & 58.2 & 38.6 & 48.7 & 45.3 \\
KlearAgentForge & Strong LLMs  & 40.7 & 69.3 & 36.3 & 46.7 & 48.2 \\
AgenticQwen     & Qwen3-235B   & 45.3 & 65.5 & 47.1 & 56.0 & 53.4 \\

\midrule
\multicolumn{7}{l}{\emph{Synthesis Variant Baselines}} \\[-1pt]
Monolithic Synthesis & Opus 4.5 & 47.3 & 53.5 & 56.4 & 40.6 & 49.5 \\
Fixed-State Synthesis       & Opus 4.5 & 54.7 & 59.1 & 61.7 & 48.7 & 56.1 \\
R-Zero Style           & Opus 4.5 & 59.3 & 69.6 & 67.3 & 56.7 & 63.2 \\
AutoPlay Style         & Opus 4.5 & 60.7 & 71.3 & 70.8 & 55.3 & 64.5 \\

\midrule
\multicolumn{7}{l}{\emph{Ours}} \\[-1pt]
\textsc{EvoSyn}-R1
& Qwen3-30B & 62.0 & 71.4 & 43.8 & 47.3 & 56.1 \\
\rowcolor{evoblue}
\textsc{EvoSyn}-Final-SFT
& Qwen3-30B & 68.2 & 76.3 & 47.7 & 52.7 & 61.2 \\[-1pt]

\textsc{EvoSyn}-R1
& Qwen3-235B & 65.3 & 80.1 & 63.5 & 58.6 & 66.9 \\
\rowcolor{evoblue}
\textsc{EvoSyn}-Final-SFT
& Qwen3-235B & 70.6 & 84.2 & 68.4 & 63.3 & 71.6 \\[-1pt]

\textsc{EvoSyn}-R1
& Opus 4.5 & 64.8 & 78.3 & 79.8 & 60.7 & 70.9 \\
\rowcolor{evoblue}
\textbf{\textsc{EvoSyn}-Final-SFT}
& Opus 4.5
& \textbf{71.3}
& \textbf{83.6}
& \textbf{83.9}
& \textbf{72.6}
& \textbf{77.8} \\

\bottomrule
\end{tabular}%
}
\end{table*}

\section{Experiments}
\label{app:exps}

\subsection{Experimental Setup}
\label{app:experimental_setup}

\textbf{Datasets and metrics.}
We evaluate \textsc{EvoSyn} on $\tau^2$-Bench and CAR-Bench, both of which provide executable environments and automatic task-level evaluators.
Across all benchmarks, we report task success rate under the corresponding environment evaluator.
\textbf{LLM backbones.}
We include closed-source models Claude Haiku 4.5, Claude Sonnet 4.5, Claude Opus 4.5, GPT-5.2, Gemini-3 Flash and Gemini-3 Pro, and open-source Qwen3 models including 8B, 14B and 30B.
Qwen3-8B is the default task-agent backbone, Qwen3-14B and Qwen3-30B are used for task-agent scaling. Qwen3-30B, Qwen3-235B and Claude Opus 4.5 are used as synthesis LLMs.
\textbf{Experimental protocol.}
Full experimental details, hyperparameters and settings are provided in Appendix~\ref{app:exp_setup}.

\subsection{RQ1: Overall Effectiveness}

Table~\ref{tab:main_results} shows that \textsc{EvoSyn} consistently outperforms multiple classes of competing approaches across all evaluated environments. With Opus 4.5 as the synthesis LLM, \textsc{EvoSyn}-Final-SFT achieves an average success rate of 77.8\%, substantially exceeding all general agentic synthesis baselines as well as the stronger synthesis variants, including Monolithic, Fixed-State, R-Zero-style, and AutoPlay-style synthesis. It also outperforms all directly evaluated open-source foundation models despite training only a Qwen3-8B task agent, highlighting the value of targeted synthetic experience beyond model scale alone. The advantage is already visible after the first synthesis round, where \textsc{EvoSyn}-R1 reaches 70.9\%, while feedback-guided multi-round evolution further improves performance to 77.8\%, demonstrating additional gains from continually revising what experience the learner receives. Importantly, the same trend holds with both Qwen3-30B and Qwen3-235B synthesis backends, indicating that the framework is not tied to a particular frontier generator. Moreover, Appendix~\ref{app:task_agent_scaling} Table~\ref{tab:task_agent_scaling} shows that the benefit persists as the downstream learner scales: \textsc{EvoSyn}-Final-SFT reaches 80.9\% with Qwen3-14B and 86.0\% with Qwen3-30B task agents. 

\subsection{RQ2: Ablation Study}
\label{sec:rq3_ablation}

Table~\ref{tab:rq3_ablation_two_domains} shows that the full \textsc{EvoSyn} achieves the strongest downstream performance and intrinsic data quality across both Airline and CAR-Bench, consistently leading all reported metrics. Removing learning feedback reduces success from 0.713 to 0.660 on Airline and from 0.726 to 0.613 on CAR-Bench, while trajectory diversity also drops markedly, indicating that feedback reshapes the structural support of synthesized experience rather than merely filtering individual samples. Removing the Reflector consistently degrades both task performance and data quality, with CAR-Bench factor coverage falling from 0.893 to 0.803 and trajectory diversity from 0.938 to 0.800. This supports the role of synthesis-space expansion: recurrent learner failures reveal useful interaction factors that cannot be recovered by allocation over the existing space alone. Removing the Designer exposes a complementary failure mode, reducing CAR-Bench success to 0.540 and factor coverage to 0.634, showing that discovering useful factors is insufficient unless synthesis effort is actively redirected toward them. The accompanying degradation in balance, interaction coverage, and trajectory diversity further suggests that allocation affects not only which regions are emphasized, but also the overall composition of the resulting training experience. Finally, Single-Agent Synthesis performs worst overall, with Airline trajectory diversity collapsing from 0.769 to 0.077 and CAR-Bench success falling to 0.406, demonstrating that role specialization is critical for producing broad, diverse, and effective agentic supervision.

\begin{table*}[t]
\centering
\scriptsize
\setlength{\tabcolsep}{3.2pt}
\renewcommand{\arraystretch}{1.12}
\definecolor{evoblue}{RGB}{232,242,255}

\caption{
Ablation study on $\tau^2$-Bench Airline and CAR-Bench Base.
All variants use Opus 4.5 as the synthesis LLM, Qwen3-8B as the task agent, Final-SFT as the training protocol, and the same candidate generation budget.
Intrinsic data-quality metrics are computed only on the final verified synthetic samples and normalized to $[0,1]$, where higher is better.
}
\label{tab:rq3_ablation_two_domains}

\resizebox{\textwidth}{!}{%
\begin{tabular}{lccccc|ccccc}
\toprule
\multirow{2}{*}{\textbf{Method}}
& \multicolumn{5}{c|}{\textbf{$\tau^2$-Bench Airline}}
& \multicolumn{5}{c}{\textbf{CAR-Bench Base}} \\
\cmidrule(lr){2-6}
\cmidrule(lr){7-11}
& \makecell{\textbf{Success}\\\textbf{Rate}}
& \makecell{\textbf{Factor}\\\textbf{Coverage}}
& \makecell{\textbf{Factor}\\\textbf{Balance}}
& \makecell{\textbf{Interaction}\\\textbf{Coverage}}
& \makecell{\textbf{Trajectory}\\\textbf{Diversity}}
& \makecell{\textbf{Success}\\\textbf{Rate}}
& \makecell{\textbf{Factor}\\\textbf{Coverage}}
& \makecell{\textbf{Factor}\\\textbf{Balance}}
& \makecell{\textbf{Interaction}\\\textbf{Coverage}}
& \makecell{\textbf{Trajectory}\\\textbf{Diversity}} \\
\midrule

\rowcolor{evoblue}
\textsc{EvoSyn}
& \textbf{0.713}
& \textbf{0.891}
& \textbf{0.450}
& \textbf{0.455}
& \textbf{0.769}
& \textbf{0.726}
& \textbf{0.893}
& \textbf{0.508}
& \textbf{0.373}
& \textbf{0.938} \\

\midrule

w/o Learning Feedback
& 0.660
& 0.778
& 0.368
& 0.403
& 0.538
& 0.613
& 0.787
& 0.468
& \underline{0.351}
& 0.750 \\

w/o Reflector
& \underline{0.673}
& \underline{0.862}
& \underline{0.413}
& \underline{0.437}
& \underline{0.615}
& \underline{0.633}
& \underline{0.803}
& \underline{0.483}
& 0.350
& \underline{0.800} \\

 w/o Designer
& 0.613
& 0.766
& 0.387
& 0.412
& 0.462
& 0.540
& 0.634
& 0.450
& 0.339
& 0.733 \\

Single-Agent Synthesis
& 0.547
& 0.365
& 0.165
& 0.133
& 0.077
& 0.406
& 0.465
& 0.425
& 0.287
& 0.467 \\

\bottomrule
\end{tabular}%
}
\end{table*}

\begin{figure*}[t]
    \centering
    \includegraphics[width=\textwidth]{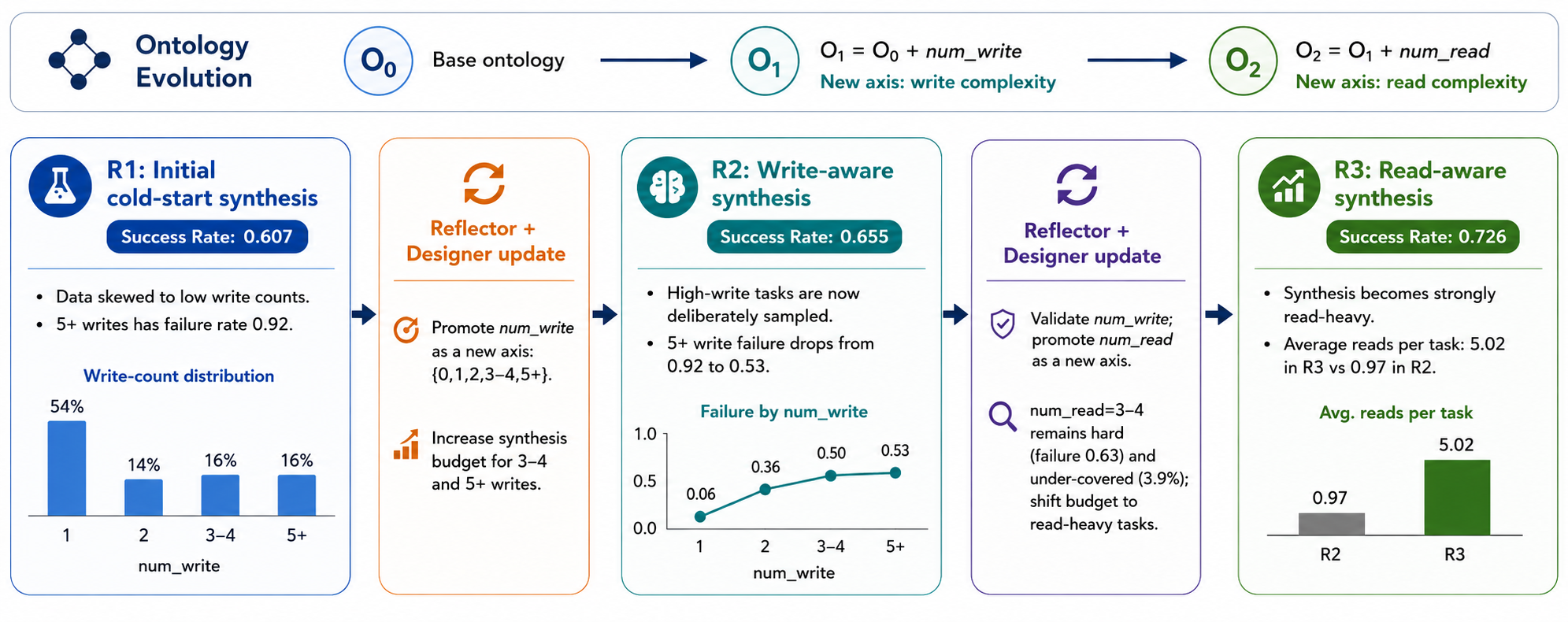}
    \caption{
Closed-loop synthesis-space evolution on CAR-Bench.
As learner weaknesses shift across rounds, \textsc{EvoSyn} promotes newly discovered interaction factors into explicit generation controls and redirects synthesis toward the evolving capability boundary, progressing from write-intensive execution to increasingly read-intensive decision making.
}
    \label{fig:carbench_round_evolution}
\end{figure*}

\subsection{RQ3: How Does Learner Feedback Evolve the Synthesis Space?}
\label{sec:rq3_self_evolution}

Figure~\ref{fig:carbench_round_evolution} illustrates how \textsc{EvoSyn} revises its synthesis focus as the learner's capability boundary changes. After R1, failures concentrate on execution-heavy tasks with many state-changing actions, prompting the Reflector to introduce \texttt{num\_write} and the Designer to shift R2 toward high-write trajectories. After R2, the residual weakness moves to read-heavy tasks, leading to the introduction of \texttt{num\_read} and a corresponding R3 shift toward read--decide--act interactions. Two controls further support this mechanism. The feedback-conditioned Reflector recovers both subsequently adopted factors ($2/2$), whereas a no-feedback control proposes six plausible dimensions but recovers neither ($0/6$; Appendix~\ref{app:rq5_nofeedback_control}, Table~\ref{tab:app_rq5_nofeedback_control}). The promoted coordinates are also operational generation controls: requested and realized factor values agree with $99.6$--$100\%$ exact accuracy and Spearman $\rho=0.997$--$1.000$ across R2 synthesis runs (Appendix~\ref{app:factor_controllability}, Table~\ref{tab:factor_controllability}). These results indicate that \textsc{EvoSyn} performs a learner-grounded synthesis-space revision rather than a generic expansion.

\begin{figure*}[t]
    \centering
    \includegraphics[width=\textwidth]{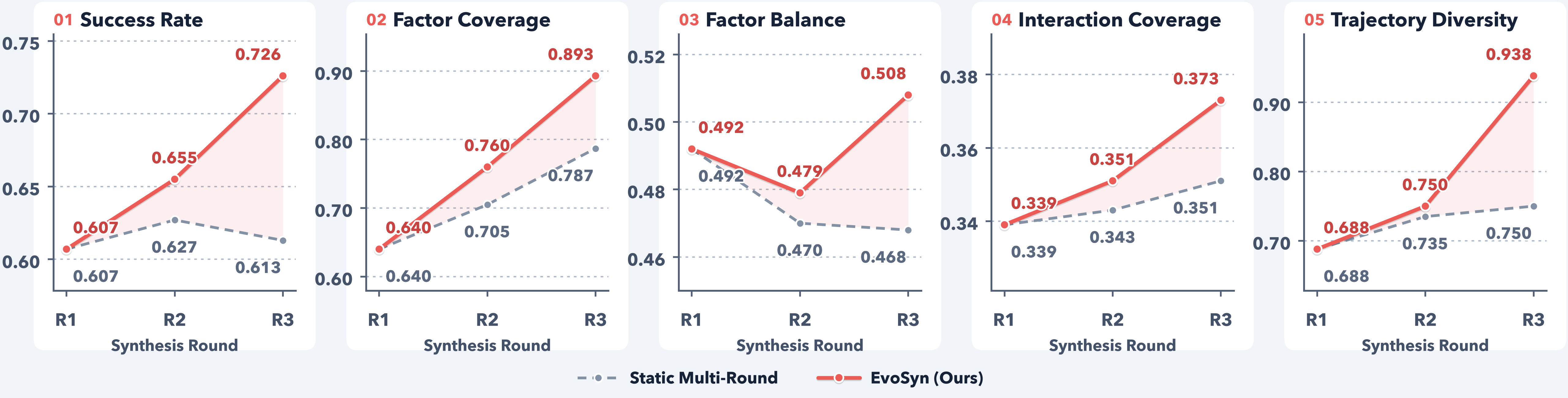}
    \caption{
Closed-loop evolution of task performance and synthetic-data quality on CAR-Bench.
Unlike static accumulation, \textsc{EvoSyn} progressively improves both downstream success and the structural quality of synthesized experience across successive synthesis rounds.
}
    \label{fig:carbench_data_quality_rounds}
\end{figure*}

Figure~\ref{fig:carbench_data_quality_rounds} further shows that this evolution yields more useful experience than simply accumulating static samples. Starting from the same R1 data, Static Multi-Round changes from 0.607 to 0.627 and then 0.613, whereas \textsc{EvoSyn} improves monotonically from 0.607 to 0.655 and 0.726. Its final data also achieves higher factor coverage (0.893 vs.\ 0.787) and trajectory-pattern diversity (0.938 vs.\ 0.750). Thus, later rounds are beneficial not merely because they add data, but because feedback changes both what can be synthesized and where synthesis effort is allocated. Detailed CAR-Bench evidence is provided in Appendix~\ref{app:carbench_case}, with a $\tau^2$-Bench case in Appendix~\ref{app:airline_case}.

\subsection{RQ4: \textsc{EvoSyn} Adapts Synthesis Toward Capability Gaps}
\label{sec:rq4_capability_gap}

\begin{figure*}[t]
    \centering
    \includegraphics[width=\textwidth]{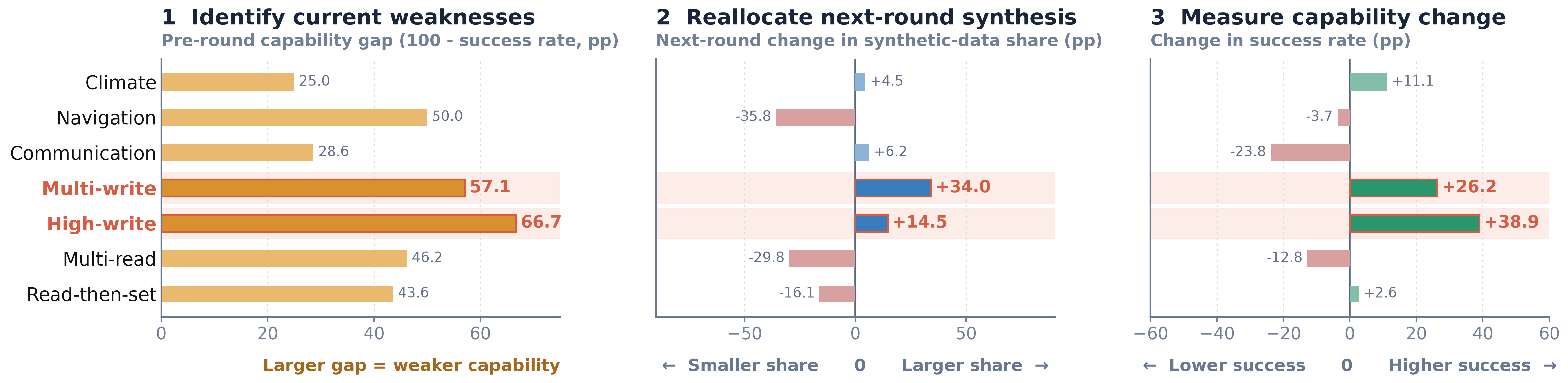}
    \caption{
Capability-gap-driven synthesis adaptation on CAR-Bench.
The three panels trace the closed-loop transition from pre-round capability gaps to next-round synthesis reallocation and subsequent capability changes, showing how \textsc{EvoSyn} redirects synthesis toward learner weaknesses.
}
    \label{fig:rq4_car_capability_evolution}
\end{figure*}

We further examine whether the closed loop in \textsc{EvoSyn} produces targeted capability evolution, rather than merely changing the data distribution or improving average performance. 
Figure~\ref{fig:rq4_car_capability_evolution} illustrates the R1$\rightarrow$R2 transition on CAR-Bench Base by aligning the task agent's pre-round capability gap, the next round's verified-data proportion shift, and the post-round capability gain over the same interaction slices.
The resulting pattern reveals a two-stage self-correcting process.
After R1, the largest actionable weakness lies in write-heavy execution; \textsc{EvoSyn} correspondingly shifts R2 synthesis toward multi-write and high-write tasks, and these targeted slices improve after training.
After R2, the residual weakness moves from execution-heavy writing to read-heavy decision making; \textsc{EvoSyn} redirects R3 synthesis toward multi-read, navigation, and read-then-set interactions, which also improve in the next evaluation.
The corresponding R2$\rightarrow$R3 transition and detailed statistics are provided in Appendix~\ref{app:rq4_targeted_evolution}, showing that the system reallocates synthesis toward the current capability boundary and pivots as that boundary evolves.

\subsection{RQ5: How Does \textsc{EvoSyn} Differ from Existing Synthetic Corpora?}
\label{sec:rq5_data_comparison}

We further compare \textsc{EvoSyn} with publicly released synthetic corpora constructed for the same environment. As shown in Figure~\ref{fig:baseline_unique_support}, under matched data budgets on Airline, \textsc{EvoSyn} provides the largest unique decisive interaction support ($73.0\%$), compared with AReaL-SEA ($69.0\%$), APIGen-MT ($49.8\%$), and WRIT ($25.5\%$). This complementarity also grows through synthesis-space evolution, with baseline-unseen structural support increasing from $13.6\%$ in R1 to $79.4\%$ in later rounds. Appendix~\ref{app:data_comparison} further shows substantially higher order-sensitive contrastive support and more balanced synthesis across interaction structures. These results show that feedback-guided evolution changes both the support and distribution of synthesized experience, rather than merely generating additional variants within a fixed synthesis space, as learner weaknesses evolve through later rounds.

\section{Conclusion}
\label{sec:conclusion}

In this paper, we present \textsc{EvoSyn}, a closed-loop multi-agent framework for self-evolving agentic data synthesis. By coupling environment-grounded generation with task-agent learning, \textsc{EvoSyn} uses post-training feedback to both reallocate synthesis effort within known interaction factors and expand the synthesis space with newly discovered factors. Experiments on $\tau^2$-Bench and CAR-Bench show consistent improvements in task-agent performance and synthetic-data quality, while demonstrating that learner feedback reveals useful interaction factors missed by feedback-free synthesis. Further analyses show that this iterative evolution progressively broadens the structural support of synthesized experience beyond that covered by existing synthesis pipelines.

\bibliography{iclr2027_conference}

\begin{thebibliography}{44}
\providecommand{\natexlab}[1]{#1}
\providecommand{\url}[1]{\texttt{#1}}
\expandafter\ifx\csname urlstyle\endcsname\relax
  \providecommand{\doi}[1]{doi: #1}\else
  \providecommand{\doi}{doi: \begingroup \urlstyle{rm}\Url}\fi

\bibitem[Cao et~al.(2025)Cao, Zhang, Li, Zhang, Liu, Duan, Zhang, and Chen]{cao2025condor}
Maosong Cao, Taolin Zhang, Mo~Li, Chuyu Zhang, Yunxin Liu, Haodong Duan, Songyang Zhang, and Kai Chen.
\newblock Condor: Enhance llm alignment with knowledge-driven data synthesis and refinement.
\newblock \emph{arXiv preprint arXiv:2501.12273}, 2\penalty0 (4), 2025.

\bibitem[Chen et~al.(2026)Chen, Qi, Gao, Wang, Wang, and Jin]{chen2026eigendata}
Jiaao Chen, Jingyuan Qi, Mingye Gao, Wei-Chen Wang, Hanrui Wang, and Di~Jin.
\newblock Eigendata: A self-evolving multi-agent platform for function-calling data synthesis, auditing, and repair.
\newblock \emph{arXiv preprint arXiv:2603.05553}, 2026.

\bibitem[Chen et~al.(2025)Chen, Zhao, Zhang, Liu, Qi, Wu, Kalluri, Cao, Xiong, Tong, et~al.]{chen2025scaling}
Zhaorun Chen, Zhuokai Zhao, Kai Zhang, Bo~Liu, Qi~Qi, Yifan Wu, Tarun Kalluri, Sara Cao, Yuanhao Xiong, Haibo Tong, et~al.
\newblock Scaling agent learning via experience synthesis.
\newblock \emph{arXiv preprint arXiv:2511.03773}, 2025.

\bibitem[Cheng et~al.(2025)Cheng, Tang, Li, Huo, Duan, Huang, Jing, and Yan]{cheng2025evolving}
Yuhao Cheng, Liang Tang, Shuxian Li, Yukang Huo, Tiaonan Duan, Kaer Huang, Yanzhe Jing, and Yiqiang Yan.
\newblock Evolving in tasks: Empowering the multi-modality large language model as the computer use agent.
\newblock \emph{arXiv preprint arXiv:2508.04037}, 2025.

\bibitem[Dai et~al.(2026)Dai, Gao, Zhang, Wang, Luo, Wang, Wang, Wu, and Wang]{dai2026sead}
Yuqin Dai, Ning Gao, Wei Zhang, Jie Wang, Zichen Luo, Jinpeng Wang, Yujie Wang, Ruiyuan Wu, and Chaozheng Wang.
\newblock Sead: Self-evolving agent for multi-turn service dialogue.
\newblock \emph{arXiv preprint arXiv:2602.03548}, 2026.

\bibitem[Dong et~al.(2026)Dong, Lu, Huang, Zhong, Liu, Huang, Li, Zhao, Song, Li, et~al.]{dong2026agent}
Guanting Dong, Junting Lu, Junjie Huang, Wanjun Zhong, Longxiang Liu, Shijue Huang, Zhenyu Li, Yang Zhao, Xiaoshuai Song, Xiaoxi Li, et~al.
\newblock Agent-world: Scaling real-world environment synthesis for evolving general agent intelligence.
\newblock \emph{arXiv preprint arXiv:2604.18292}, 2026.

\bibitem[Gao et~al.(2026)Gao, Chen, He, Wang, Xu, Wang, Jin, and Wu]{gao2026self}
Jiaxuan Gao, Jiaao Chen, Chuyi He, Wei-Chen Wang, Shusheng Xu, Hanrui Wang, Di~Jin, and Yi~Wu.
\newblock From self-evolving synthetic data to verifiable-reward rl: Post-training multi-turn interactive tool-using agents.
\newblock \emph{arXiv preprint arXiv:2601.22607}, 2026.

\bibitem[Gou et~al.(2024)Gou, Shao, Gong, Yang, Huang, Duan, Chen, et~al.]{gou2024tora}
Zhibin Gou, Zhihong Shao, Yeyun Gong, Yujiu Yang, Minlie Huang, Nan Duan, Weizhu Chen, et~al.
\newblock Tora: A tool-integrated reasoning agent for mathematical problem solving.
\newblock In \emph{International Conference on Learning Representations}, volume 2024, pp.\  48362--48395, 2024.

\bibitem[Gu et~al.(2026)Gu, Han, and Zhou]{gu2026writ}
Hengrui Gu, Xiaotian Han, and Kaixiong Zhou.
\newblock Writ: Write-read intensive trajectory synthesis for multi-turn user-facing agents.
\newblock \emph{arXiv preprint arXiv:2606.02908}, 2026.

\bibitem[He et~al.(2025)He, Fan, Kuang, Xiaoqing, Song, Zhou, and Qiu]{he2025fine}
Junliang He, Ziyue Fan, Shaohui Kuang, Li~Xiaoqing, Kai Song, Yaqian Zhou, and Xipeng Qiu.
\newblock Fine: Filtering and improving noisy data elaborately with large language models.
\newblock In \emph{Proceedings of the 2025 Conference of the Nations of the Americas Chapter of the Association for Computational Linguistics: Human Language Technologies (Volume 1: Long Papers)}, pp.\  8686--8707, 2025.

\bibitem[Huang et~al.(2025)Huang, Yu, Wang, Zhang, Li, Li, Huang, Mi, and Yu]{huang2025r}
Chengsong Huang, Wenhao Yu, Xiaoyang Wang, Hongming Zhang, Zongxia Li, Ruosen Li, Jiaxin Huang, Haitao Mi, and Dong Yu.
\newblock R-zero: Self-evolving reasoning llm from zero data.
\newblock \emph{arXiv preprint arXiv:2508.05004}, 2025.

\bibitem[Kwan et~al.(2026)Kwan, Gema, Leang, and Minervini]{kwan2026scope}
Wai-Chung Kwan, Aryo~Pradipta Gema, Joshua Ong~Jun Leang, and Pasquale Minervini.
\newblock Scope: Self-play via co-evolving policies for open-ended tasks.
\newblock \emph{arXiv preprint arXiv:2605.31433}, 2026.

\bibitem[Li et~al.(2024)Li, Beeching, Tunstall, Lipkin, Soletskyi, Huang, Rasul, Yu, Jiang, Shen, et~al.]{li2024numinamath}
Jia Li, Edward Beeching, Lewis Tunstall, Ben Lipkin, Roman Soletskyi, Shengyi Huang, Kashif Rasul, Longhui Yu, Albert~Q Jiang, Ziju Shen, et~al.
\newblock Numinamath: The largest public dataset in ai4maths with 860k pairs of competition math problems and solutions.
\newblock \emph{Hugging Face repository}, 13\penalty0 (9):\penalty0 9, 2024.

\bibitem[Li et~al.(2023)Li, Yu, Zhou, Schick, Levy, Zettlemoyer, Weston, and Lewis]{li2023self}
Xian Li, Ping Yu, Chunting Zhou, Timo Schick, Omer Levy, Luke Zettlemoyer, Jason~E Weston, and Mike Lewis.
\newblock Self-alignment with instruction backtranslation.
\newblock In \emph{The twelfth international conference on learning representations}, 2023.

\bibitem[Li et~al.(2026)Li, Wu, Li, Hu, and Liu]{li2026less}
Zhongzhi Li, Xuansheng Wu, Yijiang Li, Lijie Hu, and Ninghao Liu.
\newblock Less is enough: synthesizing diverse data in feature space of llms.
\newblock \emph{arXiv e-prints}, pp.\  arXiv--2602, 2026.

\bibitem[Liang et~al.(2026)Liang, Li, Gong, Wang, Zhang, Wu, Chen, et~al.]{liang2026sws}
Xiao Liang, Zhong-Zhi Li, Yeyun Gong, Yang Wang, Hengyuan Zhang, Ying~Nian Wu, Weizhu Chen, et~al.
\newblock Sws: Self-aware weakness-driven problem synthesis in reinforcement learning for llm reasoning.
\newblock \emph{Advances in Neural Information Processing Systems}, 38:\penalty0 56801--56839, 2026.

\bibitem[Luo et~al.(2025)Luo, Sun, Xu, Zhao, Lou, Tao, Geng, Lin, Chen, Tang, et~al.]{luo2025wizardmath}
Haipeng Luo, Qingfeng Sun, Can Xu, Pu~Zhao, Jian-Guang Lou, Chongyang Tao, Xiubo Geng, Qingwei Lin, Shifeng Chen, Yansong Tang, et~al.
\newblock Wizardmath: Empowering mathematical reasoning for large language models via reinforced evol-instruct.
\newblock In \emph{International Conference on Learning Representations}, volume 2025, pp.\  49573--49609, 2025.

\bibitem[Lyu et~al.(2026)Lyu, Wang, Zheng, Yue, Yan, Wang, and Huang]{lyu2026agenticqwen}
Yuanjie Lyu, Chengyu Wang, Haonan Zheng, Yuanhao Yue, Junbing Yan, Ming Wang, and Jun Huang.
\newblock Agenticqwen: Training small agentic language models with dual data flywheels for industrial-scale tool use.
\newblock In \emph{Proceedings of the 64th Annual Meeting of the Association for Computational Linguistics (ACL 2026)}, pp.\  535--551, 2026.

\bibitem[Pahuja et~al.(2025)Pahuja, Lu, Rosset, Gou, Mitra, Whitehead, Su, and Hassan]{pahuja2025explorer}
Vardaan Pahuja, Yadong Lu, Corby Rosset, Boyu Gou, Arindam Mitra, Spencer Whitehead, Yu~Su, and Ahmed Hassan.
\newblock Explorer: Scaling exploration-driven web trajectory synthesis for multimodal web agents.
\newblock In \emph{Findings of the Association for Computational Linguistics: ACL 2025}, pp.\  6300--6323, 2025.

\bibitem[Prabhakar et~al.(2026)Prabhakar, Liu, Zhu, Zhang, Awalgaonkar, Wang, Liu, Chen, Hoang, Niebles, et~al.]{prabhakar2026apigen}
Akshara Prabhakar, Zuxin Liu, Ming Zhu, Jianguo Zhang, Tulika~Manoj Awalgaonkar, Shiyu Wang, Zhiwei Liu, Haolin Chen, Thai Hoang, Juan~Carlos Niebles, et~al.
\newblock Apigen-mt: Agentic pipeline for multi-turn data generation via simulated agent-human interplay.
\newblock \emph{Advances in Neural Information Processing Systems}, 38, 2026.

\bibitem[Ramrakhya et~al.(2025)Ramrakhya, Szot, Attia, Yang, Nguyen, Mazoure, Gan, Agrawal, and Toshev]{ramrakhya2025scaling}
Ram Ramrakhya, Andrew Szot, Omar Attia, Yuhao Yang, Anh Nguyen, Bogdan Mazoure, Zhe Gan, Harsh Agrawal, and Alexander Toshev.
\newblock Scaling synthetic task generation for agents via exploration.
\newblock \emph{arXiv preprint arXiv:2509.25047}, 2025.

\bibitem[Shao et~al.(2024)Shao, Wang, Zhu, Xu, Song, Bi, Zhang, Zhang, Li, Wu, et~al.]{shao2024deepseekmath}
Zhihong Shao, Peiyi Wang, Qihao Zhu, Runxin Xu, Junxiao Song, Xiao Bi, Haowei Zhang, Mingchuan Zhang, YK~Li, Yang Wu, et~al.
\newblock Deepseekmath: Pushing the limits of mathematical reasoning in open language models.
\newblock \emph{arXiv preprint arXiv:2402.03300}, 2024.

\bibitem[Song et~al.(2026)Song, Chang, Dong, Zhu, Wen, and Dou]{song2026envscaler}
Xiaoshuai Song, Haofei Chang, Guanting Dong, Yutao Zhu, Ji-Rong Wen, and Zhicheng Dou.
\newblock Envscaler: Scaling tool-interactive environments for llm agent via programmatic synthesis.
\newblock In \emph{Findings of the Association for Computational Linguistics: ACL 2026}, pp.\  8326--8357, 2026.

\bibitem[Sun et~al.(2025)Sun, Cheng, Ding, Jin, Wang, Xu, Wu, Jia, Chen, Liu, et~al.]{sun2025genesis}
Qiushi Sun, Kanzhi Cheng, Zichen Ding, Chuanyang Jin, Yian Wang, Fangzhi Xu, Zhenyu Wu, Chengyou Jia, Liheng Chen, Zhoumianze Liu, et~al.
\newblock Os-genesis: Automating gui agent trajectory construction via reverse task synthesis.
\newblock In \emph{Proceedings of the 63rd Annual Meeting of the Association for Computational Linguistics (Volume 1: Long Papers)}, pp.\  5555--5579, 2025.

\bibitem[Sun et~al.(2026)Sun, Ji, Cai, Wang, Wang, Li, and Chen]{sun2026agentskiller}
Zexu Sun, Bokai Ji, Hengyi Cai, Shuaiqiang Wang, Lei Wang, Guangxia Li, and Xu~Chen.
\newblock Agentskiller: Scaling generalist agent intelligence through semantically integrated cross-domain data synthesis.
\newblock \emph{arXiv preprint arXiv:2602.09372}, 2026.

\bibitem[Tang et~al.(2025)Tang, Pang, Liu, Tang, Ye, Jin, Dong, Wang, and Chen]{tang2025synthesizing}
Shuo Tang, Xianghe Pang, Zexi Liu, Bohan Tang, Rui Ye, Tian Jin, Xiaowen Dong, Yanfeng Wang, and Siheng Chen.
\newblock Synthesizing post-training data for llms through multi-agent simulation.
\newblock In \emph{Proceedings of the 63rd Annual Meeting of the Association for Computational Linguistics (Volume 1: Long Papers)}, pp.\  23306--23335, 2025.

\bibitem[Tu et~al.(2025)Tu, Meng, He, Yao, Qi, Liu, and Li]{tu2025resofilter}
Zeao Tu, Xiangdi Meng, Yu~He, Zihan Yao, Tianyu Qi, Jun Liu, and Ming Li.
\newblock Resofilter: Fine-grained synthetic data filtering for large language models through data-parameter resonance analysis.
\newblock In \emph{Findings of the Association for Computational Linguistics: NAACL 2025}, pp.\  5414--5428, 2025.

\bibitem[Wang et~al.(2025{\natexlab{a}})Wang, Zhang, Fu, Fu, Liu, Zhang, Sun, Jiang, Tang, Ji, et~al.]{wang2025klear}
Qi~Wang, Hongzhi Zhang, Jia Fu, Kai Fu, Yahui Liu, Tinghai Zhang, Chenxi Sun, Gangwei Jiang, Jingyi Tang, Xingguang Ji, et~al.
\newblock Klear-agentforge: Forging agentic intelligence through posttraining scaling.
\newblock \emph{arXiv preprint arXiv:2511.05951}, 2025{\natexlab{a}}.

\bibitem[Wang et~al.(2026{\natexlab{a}})Wang, Zhang, Ma, Zhang, Wang, Chen, Xue, Fang, Zhang, Zhang, et~al.]{wang2026webaggregator}
Rui Wang, Ce~Zhang, Jun-Yu Ma, Jianshu Zhang, Hongru Wang, Yi~Chen, Boyang Xue, Tianqing Fang, Zhisong Zhang, Hongming Zhang, et~al.
\newblock Webaggregator: Enhancing compositional reasoning capabilities of deep research agent foundation models.
\newblock In \emph{Proceedings of the 64th Annual Meeting of the Association for Computational Linguistics (Volume 1: Long Papers)}, pp.\  24486--24517, 2026{\natexlab{a}}.

\bibitem[Wang et~al.(2025{\natexlab{b}})Wang, Jiao, Zhang, Peng, Ze, Yang, Wang, Wei, and Zhang]{wang2025socratic}
Shaobo Wang, Zhengbo Jiao, Zifan Zhang, Yilang Peng, Xu~Ze, Boyu Yang, Wei Wang, Hu~Wei, and Linfeng Zhang.
\newblock Socratic-zero: Bootstrapping reasoning via data-free agent co-evolution.
\newblock \emph{arXiv preprint arXiv:2509.24726}, 2025{\natexlab{b}}.

\bibitem[Wang et~al.(2023)Wang, Kordi, Mishra, Liu, Smith, Khashabi, and Hajishirzi]{wang2023self}
Yizhong Wang, Yeganeh Kordi, Swaroop Mishra, Alisa Liu, Noah~A Smith, Daniel Khashabi, and Hannaneh Hajishirzi.
\newblock Self-instruct: Aligning language models with self-generated instructions.
\newblock In \emph{Proceedings of the 61st annual meeting of the association for computational linguistics (volume 1: long papers)}, pp.\  13484--13508, 2023.

\bibitem[Wang et~al.(2026{\natexlab{b}})Wang, Lu, Zhang, Chen, Dong, Huang, Gesi, Tang, Luo, Liu, et~al.]{wang2026trajectory2task}
Ziyi Wang, Yuxuan Lu, Yimeng Zhang, Pei Chen, Ziwei Dong, Jing Huang, Jiri Gesi, Xianfeng Tang, Chen Luo, Qun Liu, et~al.
\newblock Trajectory2task: Training robust tool-calling agents with synthesized yet verifiable data for complex user intents.
\newblock In \emph{Proceedings of the 64th Annual Meeting of the Association for Computational Linguistics (Volume 1: Long Papers)}, pp.\  44021--44044, 2026{\natexlab{b}}.

\bibitem[Wu et~al.(2025)Wu, Wang, Mei, Cai, Fu, Yang, Wen, Yang, Shen, Wang, et~al.]{wu2025evolver}
Rong Wu, Xiaoman Wang, Jianbiao Mei, Pinlong Cai, Daocheng Fu, Cheng Yang, Licheng Wen, Xuemeng Yang, Yufan Shen, Yuxin Wang, et~al.
\newblock Evolver: Self-evolving llm agents through an experience-driven lifecycle.
\newblock \emph{arXiv preprint arXiv:2510.16079}, 2025.

\bibitem[Xia et~al.(2025)Xia, Zeng, Liu, Qin, Wu, Zhou, Xiong, and Yao]{xia2025agent0}
Peng Xia, Kaide Zeng, Jiaqi Liu, Can Qin, Fang Wu, Yiyang Zhou, Caiming Xiong, and Huaxiu Yao.
\newblock Agent0: Unleashing self-evolving agents from zero data via tool-integrated reasoning.
\newblock \emph{arXiv preprint arXiv:2511.16043}, 2025.

\bibitem[Xu et~al.(2024)Xu, Sun, Zheng, Geng, Zhao, Feng, Tao, Lin, and Jiang]{xu2024wizardlm}
Can Xu, Qingfeng Sun, Kai Zheng, Xiubo Geng, Pu~Zhao, Jiazhan Feng, Chongyang Tao, Qingwei Lin, and Daxin Jiang.
\newblock Wizardlm: Empowering large pre-trained language models to follow complex instructions.
\newblock In \emph{International Conference on Learning Representations}, volume 2024, pp.\  30745--30766, 2024.

\bibitem[Yang et~al.(2025{\natexlab{a}})Yang, Le, Xing, An, Chen, Zhao, Song, and Zhang]{yang2025toolmind}
Chen Yang, Ran Le, Yun Xing, Zhenwei An, Zongchao Chen, Wayne~Xin Zhao, Yang Song, and Tao Zhang.
\newblock Toolmind technical report: A large-scale, reasoning-enhanced tool-use dataset.
\newblock \emph{arXiv preprint arXiv:2511.15718}, 2025{\natexlab{a}}.

\bibitem[Yang et~al.(2026)Yang, Ma, Huang, Hu, Wang, and Chu]{yang2026coevolve}
Shidong Yang, Ziyu Ma, Tongwen Huang, Yiming Hu, Yong Wang, and Xiangxiang Chu.
\newblock Coevolve: Training llm agents via agent-data mutual evolution.
\newblock In \emph{Proceedings of the 64th Annual Meeting of the Association for Computational Linguistics (Volume 1: Long Papers)}, pp.\  23015--23036, 2026.

\bibitem[Yang et~al.(2025{\natexlab{b}})Yang, Weng, Pang, Cao, Ping, Zhang, Li, Zhao, Yang, Wang, et~al.]{yang2025toward}
Wei Yang, Muyan Weng, Jiacheng Pang, Defu Cao, Heng Ping, Peiyu Zhang, Shixuan Li, Yue Zhao, Qiang Yang, Mengdi Wang, et~al.
\newblock Toward evolutionary intelligence: Llm-based agentic systems with multi-agent reinforcement learning.
\newblock \emph{Available at SSRN 5819182}, 2025{\natexlab{b}}.

\bibitem[Yu et~al.(2023)Yu, Jiang, Shi, Yu, Liu, Zhang, Kwok, Li, Weller, and Liu]{yu2023metamath}
Longhui Yu, Weisen Jiang, Han Shi, Jincheng Yu, Zhengying Liu, Yu~Zhang, James~T Kwok, Zhenguo Li, Adrian Weller, and Weiyang Liu.
\newblock Metamath: Bootstrap your own mathematical questions for large language models.
\newblock \emph{arXiv preprint arXiv:2309.12284}, 2023.

\bibitem[Yue et~al.(2024)Yue, Qu, Zhang, Fu, Huang, Sun, Su, and Chen]{yue2024mammoth}
Xiang Yue, Xingwei Qu, Ge~Zhang, Yao Fu, Wenhao Huang, Huan Sun, Yu~Su, and Wenhu Chen.
\newblock Mammoth: Building math generalist models through hybrid instruction tuning.
\newblock In \emph{International Conference on Learning Representations}, volume 2024, pp.\  40320--40341, 2024.

\bibitem[Yue et~al.(2026)Yue, Upasani, Yang, Ge, Nie, Mao, Liu, and Wang]{yue2026dr}
Zhenrui Yue, Kartikeya Upasani, Xianjun Yang, Suyu Ge, Shaoliang Nie, Yuning Mao, Zhe Liu, and Dong Wang.
\newblock Dr. zero: Self-evolving search agents without training data.
\newblock \emph{arXiv preprint arXiv:2601.07055}, 2026.

\bibitem[Zelikman et~al.(2022)Zelikman, Wu, Mu, and Goodman]{zelikman2022star}
Eric Zelikman, Yuhuai Wu, Jesse Mu, and Noah Goodman.
\newblock Star: Bootstrapping reasoning with reasoning.
\newblock \emph{Advances in Neural Information Processing Systems}, 35:\penalty0 15476--15488, 2022.

\bibitem[Zhai et~al.(2025)Zhai, Tao, Chen, Zou, Chen, Fu, Mai, Yu, Deng, Cao, et~al.]{zhai2025agentevolver}
Yunpeng Zhai, Shuchang Tao, Cheng Chen, Anni Zou, Ziqian Chen, Qingxu Fu, Shinji Mai, Li~Yu, Jiaji Deng, Zouying Cao, et~al.
\newblock Agentevolver: Towards efficient self-evolving agent system.
\newblock \emph{arXiv preprint arXiv:2511.10395}, 2025.

\bibitem[Zhang et~al.(2026)Zhang, Yuan, Guo, Yu, Xu, Chen, Li, Yang, Guan, Tang, et~al.]{zhang2026evofsm}
Shuo Zhang, Chaofa Yuan, Ryan Guo, Xiaomin Yu, Rui Xu, Zhangquan Chen, Zinuo Li, Zhi Yang, Shuhao Guan, Zhenheng Tang, et~al.
\newblock Evofsm: Controllable self-evolution for deep research with finite state machines.
\newblock \emph{arXiv preprint arXiv:2601.09465}, 2026.

\end{thebibliography}
\bibliographystyle{iclr2027_conference}

\clearpage

\appendix
\section*{Appendix}

\section{Method}

\subsection{Problem Formulation}
\label{app:problem}

\paragraph{Interactive Agentic Synthesis.}
We consider synthetic data generation for improving language-model agents in interactive tool-use environments.
Let $\mathcal{E}$ denote an environment that specifies domain policies, tool interfaces, state constraints, and an automatic task evaluator, and let $\pi_{\theta_r}$ denote the task agent at synthesis round $r$.
A synthesized training example consists of a task $x$ and an interaction trajectory $\tau$, where $\tau$ contains the sequence of language responses, tool calls, and environment observations produced while solving $x$.
We denote by
\[
V_{\mathcal{E}}(x,\tau)\in\{0,1\}
\]
the environment verifier, which determines whether the task--trajectory pair is executable, policy-consistent, state-consistent, and successful under the environment evaluator.

At round $r$, a synthesis process generates a candidate set
$\widetilde{\mathcal{D}}_r=\{(x_i,\tau_i)\}_{i=1}^{N_r}$,
from which verified supervision is retained as
\begin{equation}
\mathcal{D}_r
=
\left\{
(x_i,\tau_i)\in\widetilde{\mathcal{D}}_r
\mid
V_{\mathcal{E}}(x_i,\tau_i)=1
\right\}.
\label{eq:verified_data}
\end{equation}
The task agent is trained on accumulated verified experience,
\begin{equation}
\pi_{\theta_{r+1}}
=
\operatorname{Train}
\left(
\pi_{\theta_r},
\mathcal{D}_{\leq r}
\right),
\qquad
\mathcal{D}_{\leq r}
=
\bigcup_{t=1}^{r}\mathcal{D}_t .
\label{eq:agent_update}
\end{equation}
We measure the resulting agent by its held-out task success
$J(\pi_{\theta_{r+1}})$ in the target environment.
Unlike one-pass synthesis, however, the value of newly synthesized data depends on the current learner: an example can be fully valid while providing little useful experience for an agent that has already mastered the corresponding interaction pattern.

\paragraph{Revisable Synthesis State.}
To make this learner dependence explicit, we characterize synthesis at round $r$ by a state
\begin{equation}
S_r = \left(\mathcal{O}_r,\alpha_r\right),
\label{eq:synthesis_state}
\end{equation}
which separates two complementary decisions.
The \emph{synthesis space} $\mathcal{O}_r$ specifies \emph{what interaction properties can be deliberately synthesized}, while $\alpha_r$ specifies \emph{where synthesis effort is allocated within that space}.
We use $\mathcal{Z}(\mathcal{O}_r)$ to denote the set of synthesis directions expressible under $\mathcal{O}_r$.
A direction $z\in\mathcal{Z}(\mathcal{O}_r)$ corresponds to a controllable region of the synthesis space, whose concrete structure is introduced in Section~\ref{sec:method}.

Given a synthesis budget $B_r$, we view
\begin{equation}
\alpha_r:
\mathcal{Z}(\mathcal{O}_r)
\rightarrow
\mathbb{R}_{\geq 0},
\qquad
\sum_{z\in\mathcal{Z}(\mathcal{O}_r)}
\alpha_r(z)
=
B_r,
\label{eq:synthesis_allocation}
\end{equation}
as an allocation of generation effort across synthesis directions.
The synthesis process can therefore be written abstractly as
\[
\widetilde{\mathcal{D}}_r
\sim
\mathcal{G}
\left(
\mathcal{E},
\mathcal{O}_r,
\alpha_r
\right),
\]
where $\mathcal{G}$ realizes high-level synthesis decisions as concrete agentic tasks and trajectories.
This decomposition distinguishes changing the distribution of experience within a known space from changing the space of experience that can be deliberately generated in the first place.

\paragraph{Feedback-to-Decision Revision.}
After training, the updated task agent is evaluated to obtain learner feedback
\begin{equation}
F_r
=
\operatorname{Feedback}
\left(
\pi_{\theta_{r+1}},
\mathcal{D}_{\leq r},
\mathcal{O}_r,
\mathcal{E}
\right),
\label{eq:learner_feedback}
\end{equation}
which summarizes evidence about the learner's current capabilities and the synthesized experience it has received.
The central problem is not merely to observe $F_r$, but to translate it into an actionable revision of future synthesis.
We formalize this step through a synthesis revision operator
\begin{equation}
S_{r+1}
=
\Psi(S_r,F_r),
\qquad
(\mathcal{O}_{r+1},\alpha_{r+1})
=
\Psi(\mathcal{O}_r,\alpha_r,F_r).
\label{eq:synthesis_revision}
\end{equation}
This revision map captures the \textbf{feedback-to-decision gap}: learner feedback is expressed through observed behavior, whereas future synthesis requires decisions over controllable experience.

Importantly, feedback can imply two fundamentally different forms of revision.
First, when a remaining weakness is already expressible by the current synthesis coordinates, the mismatch is \emph{distributional}.
The synthesis space remains fixed while generation effort is redistributed:
\begin{equation}
\underbrace{\mathcal{O}_{r+1}=\mathcal{O}_r}_{\text{same synthesis space}},
\qquad
\underbrace{\alpha_{r+1}\neq\alpha_r}_{\text{revised allocation}}.
\label{eq:within_space}
\end{equation}
We refer to this case as \textbf{within-space adaptation}.
It captures settings in which the appropriate interaction factor is already controllable, but the learner has received insufficient, imbalanced, or poorly targeted experience along that factor.

Second, learner feedback may expose a systematic weakness that cannot be adequately represented by any existing synthesis coordinate.
In this case, reallocating $\alpha_r$ within the current space cannot directly target the underlying capability gap.
The synthesis representation itself must be revised:
\begin{equation}
\mathcal{O}_{r+1}
=
\mathcal{O}_r
\cup
\Delta\mathcal{O}_r,
\qquad
\Delta\mathcal{O}_r\neq\varnothing,
\label{eq:space_expansion}
\end{equation}
where $\Delta\mathcal{O}_r$ contains newly introduced controllable interaction factors.
We refer to this case as \textbf{synthesis-space expansion}.
The allocation $\alpha_{r+1}$ is then defined over the expanded set of synthesis directions
$\mathcal{Z}(\mathcal{O}_{r+1})$.
Conventional adaptation over a fixed synthesis space is a special case in which
$\mathcal{O}_{r+1}=\mathcal{O}_r$ for all rounds.

\paragraph{Closed-Loop Objective.}
The overall synthesis--learning process can be summarized as
\begin{equation}
S_r
\xrightarrow{\;\text{synthesize}\;}
\mathcal{D}_r
\xrightarrow{\;\text{train}\;}
\pi_{\theta_{r+1}}
\xrightarrow{\;\text{evaluate}\;}
F_r
\xrightarrow{\;\text{revise}\;}
S_{r+1}.
\label{eq:closed_loop_problem}
\end{equation}
Given a total synthesis budget $B$, our goal is to construct a sequence of feedback-conditioned synthesis states that maximizes the final task-agent performance,
\begin{equation}
\max_{S_{1:R}}
\quad
J\!\left(\pi_{\theta_{R+1}}\right)
\qquad
\text{s.t.}
\qquad
\sum_{r=1}^{R} B_r \leq B,
\quad
S_{r+1}=\Psi(S_r,F_r).
\label{eq:overall_objective}
\end{equation}
The resulting problem therefore requires deciding, at each round, whether learner feedback should change \emph{where} synthesis effort is placed within the current space or \emph{what} the synthesis process is capable of explicitly controlling.
Section~\ref{sec:method} introduces \textsc{EvoSyn}, which instantiates this revision-and-realization process through a role-specialized multi-agent architecture.

\subsection{Comparison with Existing Data Synthesis Frameworks}
\label{app:related_work_comparison}

Table~\ref{tab:related_work_capabilities} reveals a clear progression in existing data-synthesis frameworks, but also a persistent separation between environment grounding, learner adaptation, and synthesis-space evolution. Reasoning-oriented methods such as R-Zero, Socratic-Zero, and SwS make generation learner-aware by targeting solver failures or capability boundaries, yet operate outside executable, stateful agent environments. Agentic synthesis methods such as AReaL-SEA, Explorer, AutoPlay, WebAggregator, AgentSkiller, and EnvScaler address the complementary problem: they generate executable and verifiable interaction experience, but synthesis is largely driven by environment structure, coverage, or data quality rather than by the trained learner's post-training response. More tightly coupled systems, including Agent-World, DreamGym, SEAD, Agent0, and AgenticQwen, further incorporate task-agent feedback and closed-loop resynthesis. However, their adaptation remains confined to already available task representations, predefined dimensions, or free-form generation, without explicitly revising the synthesis space when the learner exposes a previously unrepresented capability gap.

\textsc{EvoSyn} closes this gap by making the synthesis representation itself part of the learning loop. Under the conservative criteria in Table~\ref{tab:related_work_capabilities}, it is the only framework that jointly supports all eight capabilities, and in particular the only one that combines \emph{learning-effect feedback} with explicit \emph{new-factor discovery}. Post-training behavior is mapped into a structured and revisable synthesis space that serves simultaneously as a diagnostic representation and a generation-control interface. When a weakness is already expressible in the current space, \textsc{EvoSyn} performs known-factor adaptation by reallocating synthesis effort toward weak or under-covered regions. When the weakness cannot be represented by existing coordinates, it instead expands the space by promoting a new reusable and controllable interaction factor. This distinction is critical: rather than merely generating harder instances, repairing failed samples, or resampling weak categories, \textsc{EvoSyn} can change both \emph{where} synthesis is concentrated and \emph{what} the synthesis process is capable of explicitly controlling. The resulting loop therefore connects executable data generation, task-agent training, post-training diagnosis, synthesis-space revision, and targeted resynthesis within a unified closed-loop framework.

\begin{table*}[t]
\centering
\scriptsize
\setlength{\tabcolsep}{3.4pt}
\renewcommand{\arraystretch}{1.15}

\caption{
Comparison of representative data-synthesis methods for post-training LLMs and agents.
\cmark\ indicates that the capability is explicitly supported and demonstrated by the method,
while \nomark\ indicates that it is absent or not demonstrated.
Partial or implicit support is conservatively marked as \nomark.
}
\label{tab:related_work_capabilities}

\resizebox{\textwidth}{!}{%
\begin{tabular}{lcccccccc}
\toprule
\textbf{Method}
&
\makecell{\textbf{Interactive}\\\textbf{Setting}}
&
\makecell{\textbf{Executable \&}\\\textbf{Verifiable}}
&
\makecell{\textbf{Task-Agent}\\\textbf{Feedback}}
&
\makecell{\textbf{Learning-Effect}\\\textbf{Feedback}}
&
\makecell{\textbf{Structured}\\\textbf{Task Space}}
&
\makecell{\textbf{Known-Factor}\\\textbf{Adaptation}}
&
\makecell{\textbf{New-Factor}\\\textbf{Discovery}}
&
\makecell{\textbf{Closed-Loop}\\\textbf{Evolution}}
\\
\midrule

\multicolumn{9}{l}{
\textit{Reasoning and instruction synthesis}
}
\\[1pt]

MATRIX / MATRIX-Gen
& \nomark
& \nomark
& \nomark
& \nomark
& \cmark
& \nomark
& \nomark
& \nomark
\\

Socratic-Zero
& \nomark
& \nomark
& \cmark
& \nomark
& \nomark
& \cmark
& \nomark
& \cmark
\\

R-Zero
& \nomark
& \nomark
& \cmark
& \nomark
& \nomark
& \cmark
& \nomark
& \cmark
\\

SwS
& \nomark
& \nomark
& \cmark
& \nomark
& \cmark
& \cmark
& \nomark
& \nomark
\\

\midrule

\multicolumn{9}{l}{
\textit{Interactive and agentic data synthesis}
}
\\[1pt]

AReaL-SEA / EigenData
& \cmark
& \cmark
& \nomark
& \nomark
& \nomark
& \nomark
& \nomark
& \nomark
\\

SCOPE
& \cmark
& \nomark
& \cmark
& \nomark
& \nomark
& \cmark
& \nomark
& \cmark
\\

Agent-World
& \cmark
& \cmark
& \cmark
& \nomark
& \cmark
& \cmark
& \nomark
& \cmark
\\

WebAggregator
& \cmark
& \cmark
& \nomark
& \nomark
& \cmark
& \nomark
& \nomark
& \nomark
\\

DreamGym
& \cmark
& \cmark
& \cmark
& \nomark
& \nomark
& \cmark
& \nomark
& \cmark
\\

Explorer
& \cmark
& \cmark
& \nomark
& \nomark
& \nomark
& \nomark
& \nomark
& \nomark
\\

AutoPlay
& \cmark
& \cmark
& \nomark
& \nomark
& \cmark
& \nomark
& \nomark
& \nomark
\\

SEAD
& \cmark
& \cmark
& \cmark
& \nomark
& \cmark
& \cmark
& \nomark
& \cmark
\\

Agent0
& \cmark
& \cmark
& \cmark
& \nomark
& \nomark
& \cmark
& \nomark
& \cmark
\\

AgentSkiller
& \cmark
& \cmark
& \nomark
& \nomark
& \cmark
& \nomark
& \nomark
& \nomark
\\

EnvScaler
& \cmark
& \cmark
& \nomark
& \nomark
& \cmark
& \nomark
& \nomark
& \nomark
\\

Klear-AgentForge
& \cmark
& \cmark
& \cmark
& \nomark
& \nomark
& \nomark
& \nomark
& \nomark
\\

AgenticQwen
& \cmark
& \cmark
& \cmark
& \nomark
& \cmark
& \cmark
& \nomark
& \cmark
\\

\midrule

\rowcolor{gray!15}
\textbf{Ours}
& \cmark
& \cmark
& \cmark
& \cmark
& \cmark
& \cmark
& \cmark
& \cmark
\\

\bottomrule
\end{tabular}%
}

\vspace{3pt}

\begin{minipage}{0.99\textwidth}
\footnotesize
\textbf{Criterion definitions.}
\textit{Interactive Setting} denotes synthesis for multi-turn interaction with
users, tools, or stateful environments.
\textit{Executable \& Verifiable} requires synthesized tasks or trajectories
to be executable in an environment and automatically verifiable.
\textit{Task-Agent Feedback} means that synthesis is informed by the current task agent's behavior, performance, or failures.
\textit{Learning-Effect Feedback} further evaluates how previously synthesized data changes the task agent after training and uses this learning effect to guide subsequent synthesis.
\textit{Structured Task Space} denotes an explicit representation based
on task factors, schemas, states, or an ontology.
\textit{Known-Factor Adaptation} adjusts difficulty, coverage, proportions, or
generation budgets over already represented factors.
\textit{New-Factor Discovery} requires discovering a previously unrepresented
reusable synthesis factor and incorporating it into future synthesis;
generating new instances, trajectories, or environment states alone does not
satisfy this criterion.
\textit{Closed-Loop Evolution} requires repeated synthesis, learner training,
evaluation, and feedback-driven resynthesis.
\end{minipage}

\end{table*}

\subsection{Closed-Loop Role-Specialized Synthesis}
\label{app:method_overview_loop}

Agentic data synthesis spans multiple levels of abstraction.
The system must reason about what experience the current learner needs, allocate synthesis effort accordingly, translate abstract synthesis targets into feasible tasks under environment constraints, realize successful tool-use trajectories, and verify that the resulting supervision is valid.
We therefore factorize \textsc{EvoSyn} by \emph{decision abstraction}, assigning specialized agents to distinct roles in the synthesis--learning loop rather than forcing a single generator to jointly solve these heterogeneous decisions.

At round $r$, task-agent feedback $F_r$ first enters a \emph{backward synthesis-revision path}.
A Reflector Agent reasons over learner performance, data coverage, and failure structure to revise the synthesis state,
\begin{equation}
    F_r
    \xrightarrow{\;\textsc{Reflector}\;}
    S_{r+1}
    =
    (\mathcal{O}_{r+1},\alpha_{r+1}),
\label{eq:reflective_revision_overview}
\end{equation}
where $\mathcal{O}_{r+1}$ determines which interaction factors can be explicitly controlled and $\alpha_{r+1}$ determines how synthesis effort is allocated over them.
Importantly, the Reflector revises the synthesis process itself rather than directly repairing individual failed examples.

The revised synthesis state is then realized through a \emph{forward grounded-synthesis path}.
A Designer Agent operates at the strategic level, translating the synthesis state into generation directions and budgets.
A Planner Agent grounds each abstract direction into an environment-consistent task blueprint, bridging synthesis-level intent and executable task construction.
The Generator Agent and rollout teacher instantiate these blueprints into concrete tasks and successful interaction trajectories, while a Verifier Agent filters invalid supervision according to environment and task constraints.
The resulting verified experience is used to train and evaluate the task agent, producing new feedback that initiates the next round of synthesis revision.

This decomposition gives the major components of \textsc{EvoSyn} distinct but complementary roles.
The Reflector performs \emph{meta-level synthesis revision}, the Designer performs \emph{round-level synthesis planning}, and the Planner performs \emph{instance-level grounding}, while the Generator, rollout teacher, and Verifier realize and validate the resulting experience in the environment.
Together, these components form a closed loop in which learner behavior reshapes the synthesis process, and the revised synthesis process determines the experience from which the learner evolves.

\section{Experiments}

This appendix provides additional experimental details and analyses supporting the main results. We first describe the complete experimental setup, including datasets, synthesis backends, verification, task-agent training, controlled baselines, intrinsic data-quality metrics, initial synthesis spaces, and representative agent prompts. We then examine the robustness and properties of \textsc{EvoSyn} through task-agent scaling, Designer ablations, and comparative structural analyses against existing synthetic corpora. A series of mechanism-focused experiments further investigates capability-gap-driven reallocation, feedback-dependent factor discovery, controllability of evolved synthesis factors, structural-support expansion, residual learner weaknesses, and observed learning gains across rounds. Finally, detailed CAR-Bench and $\tau^2$-Bench Airline case studies trace how learner feedback is converted into synthesis-space revisions and subsequent changes in the generated experience.

\subsection{Experimental Details}
\label{app:exp_setup}

\paragraph{Datasets and Evaluation.}
We evaluate \textsc{EvoSyn} on $\tau^2$-Bench and CAR-Bench, both of which provide executable interactive environments and automatic task-level evaluators. For $\tau^2$-Bench, we consider the Airline, Retail, and Telecom domains, while CAR-Bench is evaluated under the Base setting. We follow the official dataset splits and evaluate on held-out benchmark tasks using the benchmark-native binary task-success reward. Unless otherwise specified, reported results are averaged over three random seeds.

\paragraph{Synthesis Backends.}
We instantiate \textsc{EvoSyn} with three synthesis backends: Claude Opus 4.5, Qwen3-30B-A3B-Thinking-2507, and Qwen3-235B-A22B-Thinking-2507. The open-source Qwen3 models are served with vLLM. Within each synthesis run, the Designer, Planner, Generator, Repair module, and Reflector share the selected backend. Claude Opus 4.5 is used as the user simulator during both synthesis and evaluation, with thinking disabled to match the benchmark-standard simulator configuration. Unless otherwise specified, the Designer, Planner, and Generator use a sampling temperature of 1.0. For brevity, we refer to Qwen3-30B-A3B-Thinking-2507 and Qwen3-235B-A22B-Thinking-2507 as Qwen3-30B and Qwen3-235B throughout the paper.

\paragraph{Synthesis and Verification.}
Each round generates a fixed budget of synthesis blueprints within the corresponding experimental arm and retains only environment-grounded trajectories that pass deterministic replay and achieve the benchmark reward of 1. Main synthesis runs typically use approximately 1,200--1,500 blueprint proposals per round, with domain-specific adjustments where necessary. When a proposed pseudo-ground-truth trajectory fails replay, the Repair module is invoked for at most two attempts, with every repaired trajectory replayed from the original environment state. The synthesis process runs for at most three rounds and terminates early after two consecutive rounds without a synthesis-space update. To keep synthesis-space evolution gradual and interpretable, the Reflector introduces at most one new synthesis factor per round. To construct learning feedback, we sample diagnostic examples from each synthesis round and accumulate them across rounds, allowing the Reflector to analyze the learner against both newly generated experience and the evolving synthesis history.

\paragraph{Task-Agent Training.}
Qwen3-8B serves as the default downstream task agent, while Qwen3-14B and Qwen3-30B are additionally evaluated for task-agent scaling. We perform full-parameter supervised fine-tuning with LLaMA-Factory and DeepSpeed ZeRO-3 using the same training recipe across model scales and synthesis rounds. We use a maximum sequence length of 16,384, six epochs, a learning rate of $1\times10^{-5}$ with cosine decay, a warm-up ratio of 0.1, BF16 precision, and FlashAttention-2. The per-device batch size is 1 with two gradient-accumulation steps over eight GPUs, yielding an effective global batch size of 16. Training examples are exported in ShareGPT format from verified successful trajectories, and multi-round \textsc{EvoSyn} trains on the accumulated verified experience synthesized up to the corresponding round.

\paragraph{Controlled Comparisons.}
For synthesis variants and ablations, we reuse the same environment, tool schema, domain policy, verifier, task-agent initialization, SFT recipe, and evaluation harness, changing only the synthesis mechanism under study. All generated trajectories are subject to the same environment replay and reward-based verification procedure, enabling controlled comparisons of feedback, synthesis-space evolution, and role specialization.

\paragraph{Evaluation Protocol.}
Learner feedback used for synthesis revision is computed on a held-out diagnostic set that is disjoint from the official test set. The official held-out benchmark tasks are reserved for final evaluation and reporting, using a rollout temperature of 1.0 and the corresponding native evaluator. Test labels are never used to drive synthesis or training. Analyses involving newly promoted synthesis factors use the learner state before the corresponding factor is introduced, preventing post-hoc leakage from subsequent factor-guided synthesis or training.

\paragraph{Baselines.}
We compare against four recent agentic data synthesis and post-training methods.
\textbf{AgentSkiller}~\citep{sun2026agentskiller} constructs executable environments and synthesizes high-fidelity multi-turn interaction trajectories across semantically connected domains, combining structured domain modeling, cross-domain composition, and execution-based validation to train generalist agents.
\textbf{EnvScaler}~\citep{song2026envscaler} programmatically synthesizes tool-interactive environments and further generates scenarios, tasks, verification functions, and interaction trajectories for supervised fine-tuning and reinforcement learning.
\textbf{Klear-AgentForge}~\citep{wang2025klear} builds general-purpose agentic models through large-scale supervised fine-tuning followed by multi-turn reinforcement learning, using trajectories distilled or synthesized from strong LLMs together with filtering and verification.
\textbf{AgenticQwen}~\citep{lyu2026agenticqwen} adopts iterative data flywheels to progressively increase training difficulty, including an agentic flywheel that expands workflows into multi-branch behavior structures and converts the resulting branches into new executable tasks for subsequent reinforcement learning. For these baselines, we evaluate the reported synthetic data or training pipeline using Qwen3-8B as the downstream task agent for a consistent comparison.

We additionally construct four synthesis-variant baselines under the same environment, verifier, task-agent initialization, training recipe, and evaluation harness.
\textbf{Monolithic Synthesis} removes the role-specialized decomposition and directly generates tasks and trajectories with a single synthesis agent.
\textbf{Fixed-State Synthesis} freezes the initial synthesis space and allocation, disabling feedback-driven revision across rounds.
\textbf{R-Zero Style}~\citep{huang2025r} uses learner failures as unstructured synthesis guidance without an explicit revisable synthesis space.
\textbf{AutoPlay Style}~\citep{ramrakhya2025scaling} uses environment-grounded trajectory exploration without learner-feedback-driven synthesis-space evolution.

\paragraph{Intrinsic Data-Quality Metrics.}
We evaluate the verified synthetic data using four deterministic intrinsic metrics, all normalized to $[0,1]$. 
Metrics are computed only on samples whose ground-truth trajectories successfully replay to full environment reward. 
Importantly, all structural factors are re-derived from the realized ground-truth tool actions rather than from the synthesis plan, avoiding inflation from intended but unrealized generation targets.
We consider eight observed factors, including task family, tool-use pattern, expected outcome, state-change type, computation requirement, numbers of write and read actions, and policy constraints.

\textbf{Factor Coverage.}
Let $\mathcal{V}_k$ denote the fixed reference value set of factor $k$, and
$\mathcal{O}_k$ the distinct values observed in the verified data.
We measure the fraction of each factor's reference space that is realized:
\begin{equation}
\mathrm{FactorCov}
=
\frac{1}{|\mathcal{K}|}
\sum_{k\in\mathcal{K}}
\frac{|\mathcal{O}_k \cap \mathcal{V}_k|}
{|\mathcal{V}_k|}.
\end{equation}
Higher values therefore indicate broader marginal coverage of controllable interaction factors.
For CAR-Bench, where marginal factor values saturate across methods, we instead use combinatorial coverage over the joint factor configuration space, computed as the fraction of distinct observed factor cells relative to the union of cells across the compared conditions.

\textbf{Factor Balance.}
Coverage alone does not reveal whether synthesis is concentrated on only a few values.
For each factor $k$, let $p_k$ denote the empirical distribution over values in $\mathcal{V}_k$.
We compute normalized Shannon entropy,
\begin{equation}
\mathrm{FactorBal}
=
\frac{1}{|\mathcal{K}|}
\sum_{k\in\mathcal{K}}
\frac{H(p_k)}
{\log |\mathcal{V}_k|},
\end{equation}
where $H(\cdot)$ is Shannon entropy.
A larger value indicates that synthesis mass is distributed more evenly across the factor values.

\textbf{Interaction Coverage.}
To capture compositional structure beyond individual factors, we measure coverage over predefined pairs of interaction factors.
For each template $(a,b)\in\mathcal{P}$, let $\mathcal{O}_{a,b}$ denote the distinct observed value pairs.
We compute
\begin{equation}
\mathrm{InteractCov}
=
\frac{1}{|\mathcal{P}|}
\sum_{(a,b)\in\mathcal{P}}
\frac{|\mathcal{O}_{a,b}|}
{|\mathcal{V}_a|\,|\mathcal{V}_b|}.
\end{equation}
The templates capture complementary interaction structure such as task family $\times$ tool pattern, task family $\times$ outcome, tool pattern $\times$ state change, and computation $\times$ write complexity.
Higher values indicate richer joint coverage of interaction configurations.

\textbf{Trajectory Diversity.}
We further characterize the structural diversity of executable trajectories independently of surface language.
Each ground-truth tool action is mapped to a functional symbol corresponding to \emph{read}, \emph{write}, \emph{compute}, or \emph{transfer}, and consecutive identical symbols are collapsed to form an abstract trajectory pattern.
Let $p_{\mathrm{traj}}$ denote the empirical distribution of these patterns and $\mathcal{P}_{\mathrm{union}}$ the union of patterns observed across all compared conditions.
We define
\begin{equation}
\mathrm{TrajDiv}
=
\frac{H(p_{\mathrm{traj}})}
{\log |\mathcal{P}_{\mathrm{union}}|}.
\end{equation}
All compared methods and rounds are evaluated against the same
$\mathcal{P}_{\mathrm{union}}$, ensuring that the normalized entropy is directly comparable.
Higher values indicate a broader and more evenly distributed set of execution structures.

\paragraph{Initial Task Ontologies}
\label{app:initial_ontologies}

We initialize each domain with a compact, domain-specific task ontology.
Airline, Retail, and Telecom share the same six ontology axes but use
different domain-specific values. CAR-Bench uses seven native axes to
better capture vehicle-state execution, infeasibility, and ambiguity.

\begin{tcolorbox}[
    title={Airline Initial Ontology $\mathcal{O}_0$},
    colback=gray!3,
    colframe=gray!60,
    fonttitle=\bfseries,
    breakable
]
\small\ttfamily
\{ \\
\quad "task\_family": [ \\
\qquad "flight\_booking", \\
\qquad "flight\_change", \\
\qquad "flight\_cancellation", \\
\qquad "refund\_or\_compensation", \\
\qquad "seat\_or\_baggage\_service", \\
\qquad "loyalty\_or\_membership\_service" \\
\quad ], \\
\quad "expected\_outcome": [ \\
\qquad "fulfill", \\
\qquad "refuse" \\
\quad ], \\
\quad "policy\_or\_constraint\_focus": [ \\
\qquad "fare\_rule\_constraint", \\
\qquad "change\_or\_cancellation\_fee", \\
\qquad "refund\_eligibility", \\
\qquad "compensation\_eligibility", \\
\qquad "passenger\_or\_identity\_constraint", \\
\qquad "time\_window\_constraint", \\
\qquad "payment\_or\_credit\_constraint", \\
\qquad "loyalty\_status\_constraint" \\
\quad ], \\
\quad "tool\_use\_pattern": [ \\
\qquad "search\_compare\_then\_book", \\
\qquad "lookup\_verify\_then\_modify", \\
\qquad "multi\_entity\_retrieval", \\
\qquad "policy\_check\_before\_action", \\
\qquad "price\_or\_fee\_recalculation", \\
\qquad "state\_change\_with\_confirmation", \\
\qquad "partial\_update\_or\_split\_action" \\
\quad ], \\
\quad "interaction\_challenge": [ \\
\qquad "missing\_information", \\
\qquad "ambiguous\_preference", \\
\qquad "user\_false\_claim", \\
\qquad "preference\_change", \\
\qquad "emotional\_pressure", \\
\qquad "multi\_intent\_interleaving", \\
\qquad "constraint\_hidden\_until\_late" \\
\quad ], \\
\quad "difficulty\_profile": ["easy", "medium", "hard"] \\
\}
\end{tcolorbox}

\begin{tcolorbox}[
    title={Retail Initial Ontology $\mathcal{O}_0$},
    colback=gray!3,
    colframe=gray!60,
    fonttitle=\bfseries,
    breakable
]
\small\ttfamily
\{ \\
\quad "task\_family": [ \\
\qquad "order\_cancellation", \\
\qquad "order\_item\_modification", \\
\qquad "order\_logistics\_modification", \\
\qquad "return\_delivered\_order", \\
\qquad "exchange\_delivered\_order", \\
\qquad "profile\_or\_inquiry" \\
\quad ], \\
\quad "expected\_outcome": ["fulfill", "refuse"], \\
\quad "policy\_or\_constraint\_focus": [ \\
\qquad "order\_status\_precondition", \\
\qquad "identity\_authentication", \\
\qquad "once\_per\_order\_action\_lock", \\
\qquad "product\_variant\_match", \\
\qquad "payment\_or\_giftcard\_balance", \\
\qquad "confirmation\_requirement", \\
\qquad "cancellation\_reason\_validity", \\
\qquad "scope\_or\_escalation\_boundary" \\
\quad ], \\
\quad "tool\_use\_pattern": [ \\
\qquad "authenticate\_then\_act", \\
\qquad "lookup\_verify\_then\_modify", \\
\qquad "multi\_entity\_retrieval", \\
\qquad "policy\_check\_before\_action", \\
\qquad "price\_or\_refund\_calculation", \\
\qquad "state\_change\_with\_confirmation", \\
\qquad "batched\_multi\_item\_action" \\
\quad ], \\
\quad "interaction\_challenge": [ \\
\qquad "missing\_information", \\
\qquad "ambiguous\_request", \\
\qquad "user\_false\_claim", \\
\qquad "preference\_change", \\
\qquad "emotional\_pressure", \\
\qquad "multi\_intent\_interleaving", \\
\qquad "constraint\_hidden\_until\_late" \\
\quad ], \\
\quad "difficulty\_profile": ["easy", "medium", "hard"] \\
\}
\end{tcolorbox}

\begin{tcolorbox}[
    title={Telecom Initial Ontology $\mathcal{O}_0$},
    colback=gray!3,
    colframe=gray!60,
    fonttitle=\bfseries,
    breakable
]
\small\ttfamily
\{ \\
\quad "task\_family": [ \\
\qquad "plan\_change", \\
\qquad "billing\_issue", \\
\qquad "device\_upgrade", \\
\qquad "line\_management", \\
\qquad "service\_troubleshooting", \\
\qquad "account\_profile\_update" \\
\quad ], \\
\quad "expected\_outcome": ["fulfill", "refuse"], \\
\quad "policy\_or\_constraint\_focus": [ \\
\qquad "eligibility\_check", \\
\qquad "contract\_or\_fee\_constraint", \\
\qquad "billing\_or\_payment\_constraint", \\
\qquad "identity\_or\_authorization\_constraint", \\
\qquad "service\_compatibility\_constraint", \\
\qquad "time\_window\_constraint" \\
\quad ], \\
\quad "tool\_use\_pattern": [ \\
\qquad "lookup\_then\_update", \\
\qquad "multi\_hop\_retrieval", \\
\qquad "compare\_options\_then\_apply", \\
\qquad "verify\_before\_action", \\
\qquad "diagnose\_then\_resolve", \\
\qquad "state\_change\_with\_confirmation" \\
\quad ], \\
\quad "interaction\_challenge": [ \\
\qquad "missing\_information", \\
\qquad "ambiguous\_request", \\
\qquad "user\_false\_claim", \\
\qquad "preference\_change", \\
\qquad "emotional\_pressure", \\
\qquad "multi\_intent\_interleaving" \\
\quad ], \\
\quad "difficulty\_profile": ["easy", "medium", "hard"] \\
\}
\end{tcolorbox}

\begin{tcolorbox}[
    title={CAR-Bench Initial Ontology $\mathcal{O}_0$},
    colback=gray!3,
    colframe=gray!60,
    fonttitle=\bfseries,
    breakable
]
\small\ttfamily
\{ \\
\quad "task\_family": [ \\
\qquad "climate\_control", \\
\qquad "window\_and\_body", \\
\qquad "navigation", \\
\qquad "charging", \\
\qquad "communication", \\
\qquad "info\_query" \\
\quad ], \\
\quad "task\_nature": [ \\
\qquad "base\_execute", \\
\qquad "hallucination\_missing\_tool", \\
\qquad "hallucination\_missing\_param", \\
\qquad "hallucination\_missing\_result", \\
\qquad "disambiguation\_internal", \\
\qquad "disambiguation\_user" \\
\quad ], \\
\quad "expected\_behavior": [ \\
\qquad "fulfill\_reach\_goal", \\
\qquad "decline\_infeasible", \\
\qquad "resolve\_from\_preferences", \\
\qquad "clarify\_with\_user" \\
\quad ], \\
\quad "tool\_use\_pattern": [ \\
\qquad "single\_set", \\
\qquad "read\_then\_set", \\
\qquad "multi\_step\_sequence", \\
\qquad "cross\_domain\_compose", \\
\qquad "confirmation\_required" \\
\quad ], \\
\quad "state\_precondition": [ \\
\qquad "no\_precondition", \\
\qquad "ordering\_dependency", \\
\qquad "policy\_guarded", \\
\qquad "occupancy\_or\_context\_gated" \\
\quad ], \\
\quad "interaction\_challenge": [ \\
\qquad "cooperative\_clear", \\
\qquad "persona\_indirect", \\
\qquad "underspecified\_value", \\
\qquad "distractor\_context" \\
\quad ], \\
\quad "difficulty\_profile": ["easy", "medium", "hard"] \\
\}
\end{tcolorbox}

For Airline, Retail, and Telecom, each task selects exactly one
\texttt{task\_family}, one \texttt{expected\_outcome}, one or two
\texttt{policy\_or\_constraint\_focus} values, one to three
\texttt{tool\_use\_pattern} values, zero to two
\texttt{interaction\_challenge} values, and one
\texttt{difficulty\_profile}. The difficulty levels correspond to
increasing tool-use and compositional requirements: easy tasks typically
require 2--3 tool calls, medium tasks 4--5 calls, and hard tasks 6 or
more calls. For the CAR-Bench Base experiments, \texttt{task\_nature} is
restricted to \texttt{base\_execute}, while
\texttt{expected\_behavior} is restricted to
\texttt{fulfill\_reach\_goal} and
\texttt{resolve\_from\_preferences}.

\paragraph{Representative Agent Prompts.}
To keep the experimental-details section concise, we defer the representative prompts for the Designer, Planner, and Reflector to Appendix~\ref{app:agent_prompts}.

\subsection{Additional Overall Performance}
\label{app:task_agent_scaling}

We further examine whether the benefits of \textsc{EvoSyn} persist as the downstream learner scales and how the resulting task agents compare with directly evaluated foundation models. As shown in Table~\ref{tab:task_agent_scaling}, scaling the base Qwen3 backbone from 8B to 14B and 30B raises the average success rate from 39.4\% to 48.6\% and 57.1\%, respectively, while leaving substantial room for post-training. Training on evolved \textsc{EvoSyn} experience consistently closes this gap, with Final-SFT reaching 77.8\%, 80.9\%, and 86.0\% across the three task-agent scales. Crucially, multi-round synthesis remains beneficial even as the learner becomes substantially stronger: relative to R1, feedback-guided evolution provides additional gains of 6.9, 7.6, and 5.7 points for Qwen3-8B, Qwen3-14B, and Qwen3-30B, respectively. These results show that task-agent scaling and synthesis evolution are complementary rather than interchangeable: larger backbones improve the learner itself, while evolving the synthesized experience continues to provide additional supervision beyond model scale alone. Most notably, the Qwen3-30B task agent trained with \textsc{EvoSyn} reaches an average success rate of 86.0\%, surpassing the strongest directly evaluated foundation-model average in the table, Claude Opus 4.5 at 85.3\%. This demonstrates that the utility of \textsc{EvoSyn} is not confined to compensating for a weak learner, but persists across substantially different model capacities and can elevate open-source task agents to frontier-level performance.

\begin{table*}[t]
\centering
\scriptsize
\setlength{\tabcolsep}{4.0pt}
\renewcommand{\arraystretch}{1.05}
\definecolor{evoblue}{RGB}{232,242,255}

\caption{
Direct foundation-model performance and task-agent scaling results of \textsc{EvoSyn}.
For \textsc{EvoSyn}, Claude Opus 4.5 is used as the synthesis LLM, while Qwen3-8B, Qwen3-14B, and Qwen3-30B serve as downstream task agents.
\textsc{EvoSyn}-R1 uses first-round synthesized data, whereas \textsc{EvoSyn}-Final-SFT trains on accumulated multi-round data.
}
\label{tab:task_agent_scaling}

\resizebox{0.96\textwidth}{!}{%
\begin{tabular}{lllccccc}
\toprule
\multirow{2}{*}{\textbf{Method}}
& \multirow{2}{*}{\textbf{Synthesis LLM}}
& \multirow{2}{*}{\textbf{Task Agent}}
& \multicolumn{3}{c}{\textbf{$\tau^2$-Bench}}
& \multirow{2}{*}{\textbf{CAR-Bench}}
& \multirow{2}{*}{\textbf{Avg.}} \\
\cmidrule(lr){4-6}
&
&
& \textbf{Airline}
& \textbf{Retail}
& \textbf{Telecom}
&
& \\
\midrule

\multicolumn{8}{l}{\emph{Frontier Closed-source Models}} \\[-1pt]
Claude Haiku 4.5  & -- & -- & 60.0 & 75.1 & 42.4 & 53.3 & 57.7 \\
Claude Sonnet 4.5 & -- & -- & 83.3 & 88.3 & 62.0 & 71.3 & 76.2 \\
Claude Opus 4.5   & -- & -- & 79.3 & 91.8 & 88.0 & 82.0 & 85.3 \\
GPT-5.2            & -- & -- & 83.0 & 81.6 & 89.7 & 74.0 & 82.1 \\
Gemini-3 Flash     & -- & -- & 82.5 & 76.8 & 91.2 & 69.3 & 80.1 \\
Gemini-3 Pro       & -- & -- & 80.5 & 75.9 & 91.0 & 67.3 & 78.7 \\

\midrule
\multicolumn{8}{l}{\emph{Open-Source Foundation Models}} \\[-1pt]
Qwen3-8B   & -- & -- & 37.3 & 40.1 & 40.3 & 40.0 & 39.4 \\
Qwen3-14B  & -- & -- & 40.7 & 61.4 & 43.0 & 49.3 & 48.6 \\
Qwen3-30B  & -- & -- & 62.0 & 75.4 & 41.8 & 49.3 & 57.1 \\
Qwen3-235B & -- & -- & 64.7 & 83.3 & 65.8 & 65.3 & 69.8 \\

\midrule
\multicolumn{8}{l}{\emph{\textsc{EvoSyn} Task-Agent Scaling}} \\[-1pt]

\textsc{EvoSyn}-R1
& Opus 4.5 & Qwen3-8B
& 64.8 & 78.3 & 79.8 & 60.7 & 70.9 \\

\rowcolor{evoblue}
\textsc{EvoSyn}-Final-SFT
& Opus 4.5 & Qwen3-8B
& 71.3 & 83.6 & 83.9 & 72.6 & 77.8 \\[-1pt]

\textsc{EvoSyn}-R1
& Opus 4.5 & Qwen3-14B
& 64.5 & 83.3 & 82.5 & 62.7 & 73.3 \\

\rowcolor{evoblue}
\textsc{EvoSyn}-Final-SFT
& Opus 4.5 & Qwen3-14B
& 73.3 & 86.5 & 90.4 & 73.3 & 80.9 \\[-1pt]

\textsc{EvoSyn}-R1
& Opus 4.5 & Qwen3-30B
& 75.3 & 87.7 & 89.5 & 68.7 & 80.3 \\

\rowcolor{evoblue}
\textbf{\textsc{EvoSyn}-Final-SFT}
& Opus 4.5 & Qwen3-30B
& \textbf{84.0}
& \textbf{89.8}
& \textbf{93.3}
& \textbf{76.7}
& \textbf{86.0} \\

\bottomrule
\end{tabular}%
}
\end{table*}

\subsection{Additional Ablation: Why Is a Designer Necessary?}
\label{app:designer_ablation}

\begin{figure}[t]
    \centering
    \includegraphics[width=0.8\linewidth]{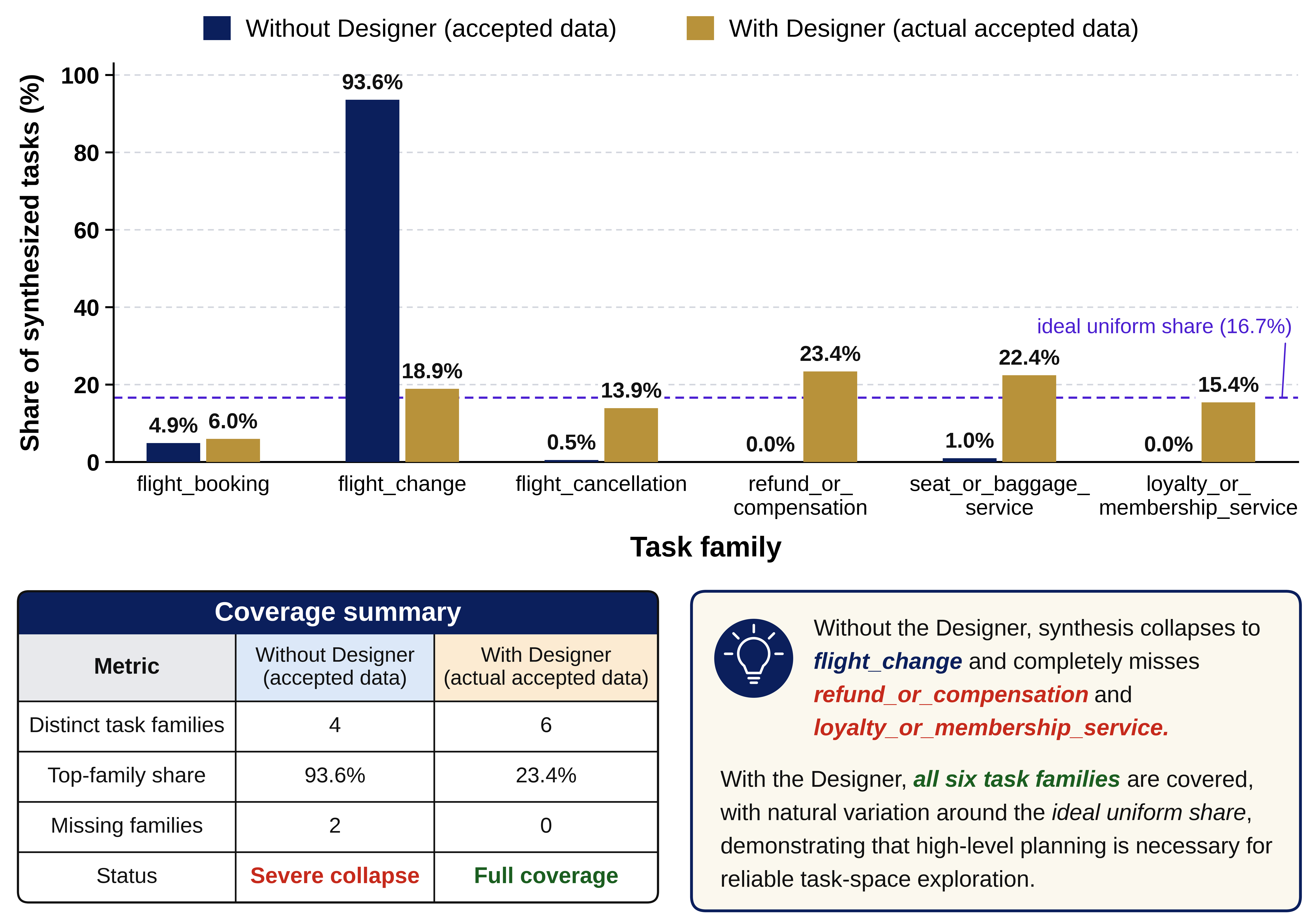}
    \caption{
    Role of the Designer in task-space exploration on $\tau^2$-Bench Airline.
    A Planner--Generator pipeline without explicit high-level allocation collapses toward \texttt{flight\_change}, whereas introducing the Designer separates global synthesis allocation from local task construction and restores coverage across all six task families.
    }
    \label{fig:designer_ablation}
\end{figure}

Many agentic synthesis pipelines primarily rely on a Planner--Generator decomposition, where the Planner determines how a task should be constructed and the Generator realizes it in the environment. We therefore further investigate whether this conventional decomposition is sufficient for self-evolving synthesis, or whether an explicit Designer is necessary to separate \emph{where to synthesize} from \emph{how to instantiate} each task. Without the Designer, the Planner must simultaneously perform global task-space allocation and local task construction. As shown in Figure~\ref{fig:designer_ablation}, this coupling causes a severe distributional collapse on Airline: \texttt{flight\_change} accounts for $93.6\%$ of accepted data, while two of the six task families are not covered at all. Introducing the Designer explicitly separates high-level synthesis allocation from environment-grounded task realization. The resulting data covers all six task families, while the largest family share drops to $23.4\%$, yielding substantially more balanced exploration of the task space. This result clarifies why the Designer is important beyond ordinary Planner--Generator synthesis: it provides the global allocation mechanism required to intentionally explore selected regions of the synthesis space, which is especially important in a self-evolving pipeline where learner feedback continually changes what should be synthesized next.

\subsection{Comparative Structural Analysis of Synthetic Data}
\label{app:data_comparison}

Beyond downstream task performance, we directly examine how the experience synthesized by \textsc{EvoSyn} differs from existing synthetic corpora at the data-structure level. We conduct three complementary analyses under matched data budgets: \emph{unique structural support} measures whether \textsc{EvoSyn} reaches interaction structures absent from existing corpora, \emph{order-sensitive contrastive support} evaluates local structural variation among similar action compositions, and \emph{balanced structural diversity} examines how broadly and evenly synthesis mass is distributed across the resulting interaction space. Together, these analyses characterize not only whether \textsc{EvoSyn} generates different data, but whether synthesis-space evolution progressively expands, enriches, and redistributes the structural support of the training experience.

\subsubsection{Unique and Expanding Structural Support}
\label{app:baseline_structural_support}

\begin{figure}[t]
    \centering
    \includegraphics[width=0.9\linewidth]{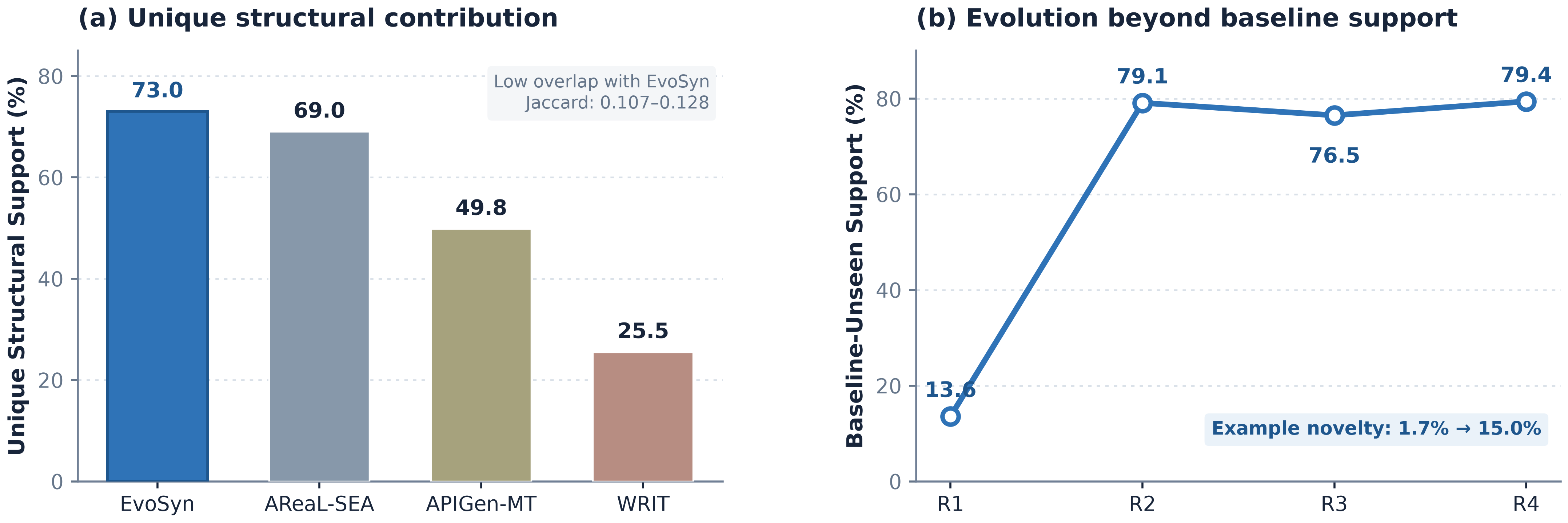}
    \caption{
    Structural complementarity of \textsc{EvoSyn} relative to domain-matched
    synthetic-data baselines on Airline.
    \textbf{Left:} fraction of each method's distinct decisive interaction support
    that is absent from the pooled support of all other methods under matched data
    budgets. \textsc{EvoSyn} provides the largest unique structural contribution.
    \textbf{Right:} fraction of \textsc{EvoSyn}'s structural support that is absent
    from the union of all matched baseline corpora across synthesis rounds.
    Iterative synthesis-space evolution progressively moves generation toward
    interaction structures not represented by existing synthesis pipelines.
    }
    \label{fig:baseline_unique_support}
\end{figure}

We further compare the structural support of \textsc{EvoSyn} with publicly released synthetic corpora constructed for the same environment. We focus on Airline, where \textsc{EvoSyn}, AReaL-SEA~\citep{gao2026self}, APIGen-MT~\citep{prabhakar2026apigen}, and WRIT~\citep{gu2026writ} all provide directly domain-matched synthetic data, and perform all comparisons under matched data budgets. To account for differences in read-action logging across released datasets, we canonicalize each example by its \emph{decisive} tool-interaction sequence, retaining write, compute, and transfer actions. We define the \emph{unique-support fraction} as the proportion of a method's distinct structural signatures that are absent from the pooled support of all other methods. As shown in Figure~\ref{fig:baseline_unique_support}, \textsc{EvoSyn} contributes the largest unique structural support at $73.0\%$, compared with $69.0\%$ for AReaL-SEA, $49.8\%$ for APIGen-MT, and $25.5\%$ for WRIT. Its pairwise overlap with these corpora is also small, with Jaccard similarities of only $0.107$--$0.128$. This indicates that \textsc{EvoSyn} is not simply reproducing structural patterns already present in existing corpora, but contributes a largely complementary set of executable interaction structures under the same data budget.

More importantly, this complementarity emerges progressively through synthesis-space evolution. Relative to the union of all matched baseline corpora, the fraction of \textsc{EvoSyn}'s support that is baseline-unseen increases sharply from $13.6\%$ in R1 to $79.1\%$ in R2 and remains consistently high through R3 and R4 ($76.5\%$ and $79.4\%$). Meanwhile, the fraction of individual examples realizing baseline-unseen structures increases from $1.7\%$ to $15.0\%$. The sharp transition after the initial round suggests that the gain is not explained by the initial synthesis prior alone: once learner feedback begins revising the synthesis space, subsequent rounds increasingly move generation toward interaction structures absent from existing pipelines. These results provide direct data-side evidence that feedback-guided evolution does more than generate additional variants within a fixed interaction distribution. Instead, it expands the realized structural support of the training experience and continually redirects synthesis toward previously underrepresented regions of the agentic interaction space.

\subsubsection{Contrastive Support for Order-Sensitive Interaction Structure}
\label{app:contrastive_support}

\begin{figure}[t]
    \centering
    \includegraphics[width=\linewidth]{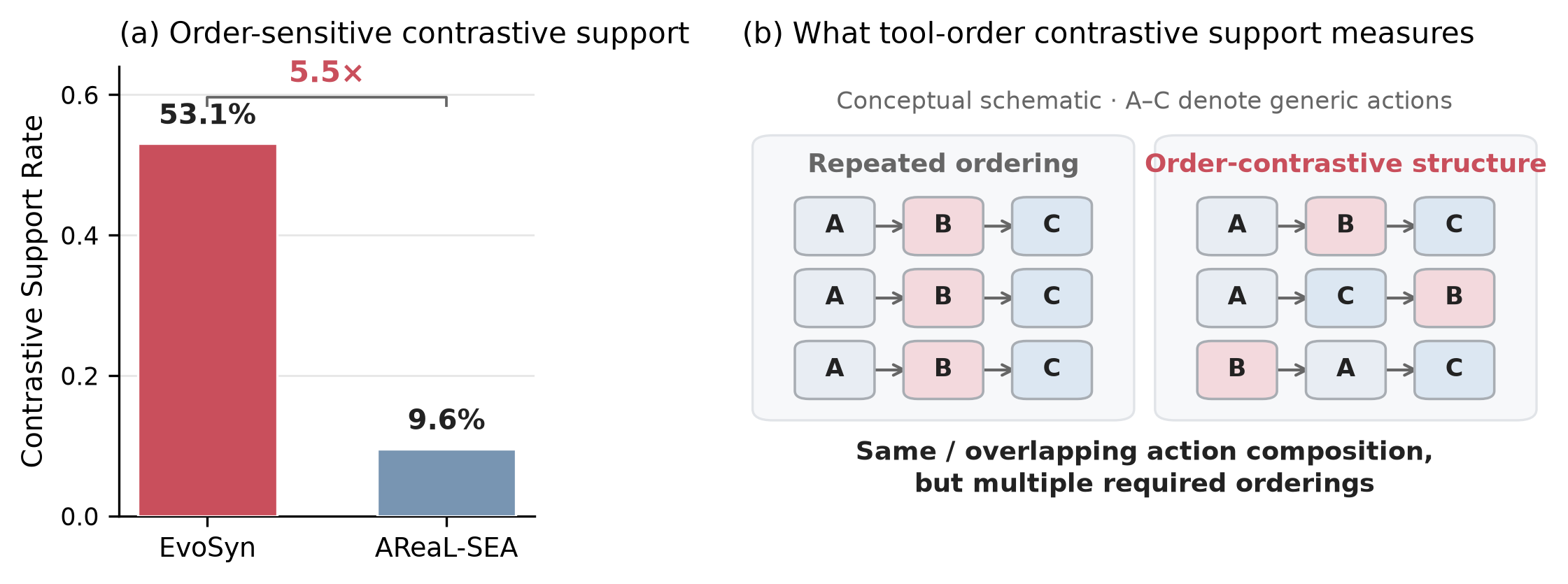}
    \caption{
    Order-sensitive contrastive structure in synthetic data.
    \textbf{Left:} On Telecom, \textsc{EvoSyn} exhibits substantially higher
    tool-order contrastive support than the domain-matched AReaL-SEA baseline
    ($53.1\%$ vs.\ $9.6\%$, $5.5\times$).
    \textbf{Right:} Conceptual illustration of the metric. Repeated use of the
    same ordering provides little order-level contrast, whereas multiple
    orderings over the same or overlapping action composition create richer
    order-sensitive structural variation.
    }
    \label{fig:contrastive_tool_order}
\end{figure}

Structural coverage alone does not reveal whether a synthetic corpus captures meaningful variation within similar interaction compositions. In agentic tasks, examples involving the same or overlapping sets of actions may still require different execution orders, making order-sensitive contrast an important source of training signal. We therefore measure whether structurally similar examples exhibit distinct decisive-action orderings. Figure~\ref{fig:contrastive_tool_order} reports the clearest domain-matched comparison on Telecom, where decisive-action traces are directly comparable across \textsc{EvoSyn} and AReaL-SEA without the read-logging asymmetry present in other domains. \textsc{EvoSyn} achieves a tool-order contrastive support rate of $53.1\%$, compared with only $9.6\%$ for AReaL-SEA, corresponding to a $5.5\times$ increase. This indicates that \textsc{EvoSyn} is substantially more likely to realize alternative order-sensitive interaction structures rather than repeatedly reproducing a single canonical ordering over similar action compositions. Complementing the unique-support analysis above, this result shows that synthesis-space evolution improves not only the global breadth of structural support, but also the local structural variation within that support, exposing execution distinctions that cannot be captured by tool-set coverage alone.

\subsubsection{Balanced Structural Diversity Across Interaction Patterns}
\label{app:balanced_structural_diversity}

\begin{figure*}[t]
    \centering
    \includegraphics[width=0.96\textwidth]{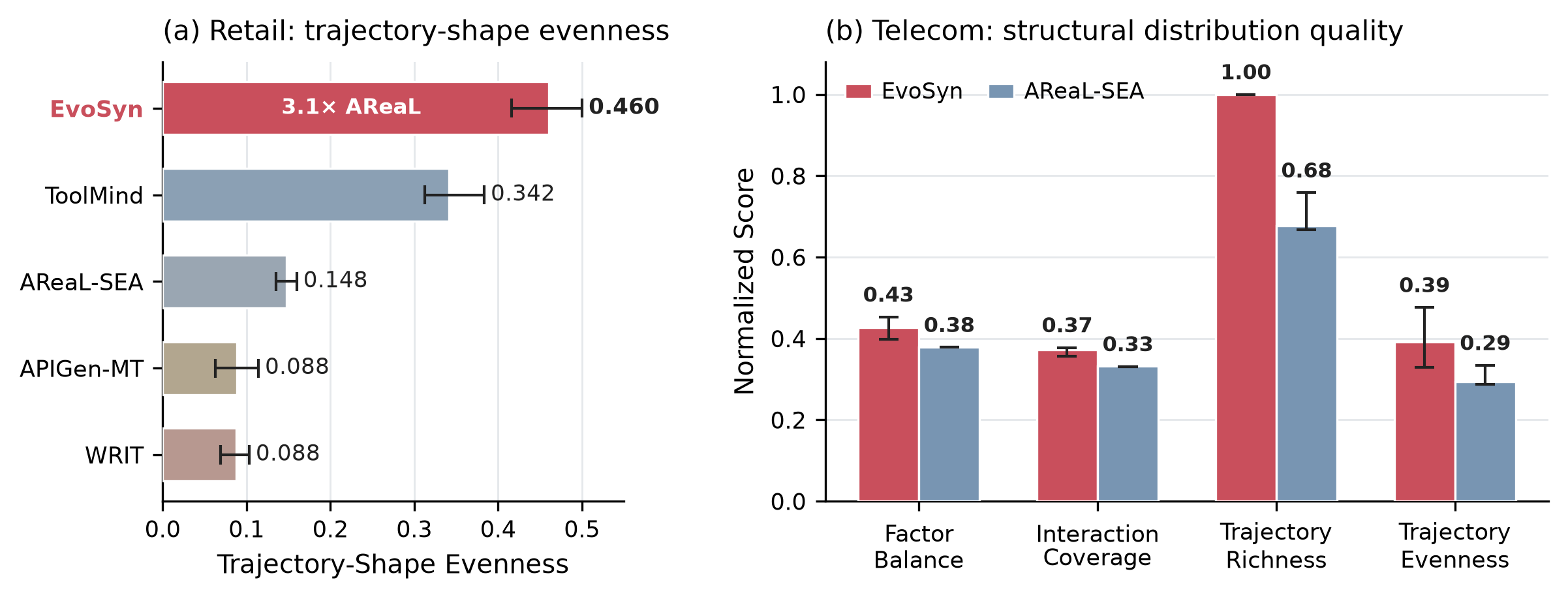}
    \caption{
    Balanced structural diversity of synthetic agent data.
    \textbf{Left:} On Retail, \textsc{EvoSyn} achieves the highest
    trajectory-shape evenness among five domain-matched synthetic corpora under
    the same data budget, exceeding the strongest competing method by
    $1.35\times$ and AReaL-SEA by $3.12\times$.
    \textbf{Right:} On Telecom, where the compared ground-truth traces have
    directly comparable decisive-action structure, \textsc{EvoSyn} provides
    higher factor balance, interaction coverage, trajectory richness, and
    trajectory evenness than AReaL-SEA.
    Together, these results show that \textsc{EvoSyn} distributes synthesis
    mass across a broader and more balanced set of interaction structures rather
    than concentrating on a small number of dominant trajectory patterns.
    }
    \label{fig:balanced_structural_diversity}
\end{figure*}

Broad structural support does not necessarily imply a useful synthetic distribution: a corpus may contain many distinct interaction patterns while still concentrating most of its mass on a small number of dominant modes. We therefore complement the support-based analyses above with distributional measures that quantify how evenly synthesized experience is spread across factors and trajectory structures under matched data budgets. As shown in Figure~\ref{fig:balanced_structural_diversity}, \textsc{EvoSyn} exhibits a particularly strong advantage on Retail, achieving a trajectory-shape evenness of $0.460$, compared with $0.342$ for ToolMind~\citep{yang2025toolmind}, $0.148$ for AReaL-SEA, and $0.088$ for both APIGen-MT and WRIT. This represents a $1.35\times$ improvement over the strongest competing method and a $3.12\times$ improvement over AReaL-SEA, with \textsc{EvoSyn}'s $95\%$ interval ($[0.416,0.500]$) remaining disjoint from those of all four baselines. This result indicates that the additional structures introduced by \textsc{EvoSyn} receive substantial synthesis mass rather than appearing only as isolated long-tail examples.

The same distributional advantage is reflected on Telecom, where the compared ground-truth trajectories admit directly comparable decisive-action structure. We additionally report trajectory richness, defined as the fraction of distinct trajectory patterns covered relative to the shared pattern union. Relative to AReaL-SEA, \textsc{EvoSyn} achieves higher factor balance ($0.427$ vs.\ $0.378$), interaction coverage ($0.372$ vs.\ $0.331$), trajectory richness ($1.000$ vs.\ $0.678$), and trajectory evenness ($0.392$ vs.\ $0.293$).

\subsection{Targeted Slice Statistics for Capability-Gap Adaptation}
\label{app:rq4_targeted_evolution}

\begin{figure*}[t]
    \centering
    \includegraphics[width=\textwidth]{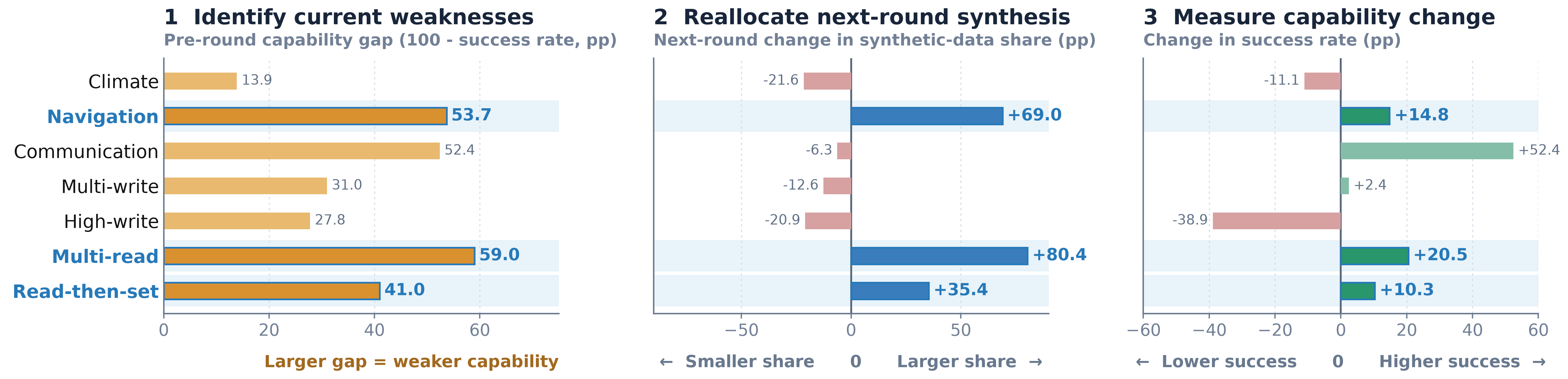}
    \caption{
Capability-gap-driven synthesis adaptation on CAR-Bench from R2 to R3.
As the learner's residual weakness shifts toward read-intensive decision making, \textsc{EvoSyn} substantially increases the verified-data share of \texttt{navigation}, \texttt{multi\_read}, and \texttt{read\_then\_set}, all of which subsequently improve after training.
}
\label{fig:rq4_car_capability_evolution_r2_r3}
\end{figure*}

Table~\ref{tab:rq4_car_targeted_evolution} provides the slice-level statistics underlying the two capability-gap transitions visualized in Figure~\ref{fig:rq4_car_capability_evolution} and Figure~\ref{fig:rq4_car_capability_evolution_r2_r3}. For each transition, we examine interaction slices associated with the promoted synthesis factor or explicitly emphasized by the next-round Designer allocation. During R1$\rightarrow$R2, the learner exhibits large gaps on write-heavy interactions: \texttt{multi\_write} and \texttt{high\_write} have capability gaps of $0.571$ and $0.667$. Their verified-data proportions subsequently increase by $0.340$ and $0.145$, followed by success-rate gains of $0.262$ and $0.389$, respectively. After this write-heavy weakness is reduced, R2$\rightarrow$R3 shifts toward read-intensive decision making, as shown in Figure~\ref{fig:rq4_car_capability_evolution_r2_r3}. \texttt{navigation}, \texttt{multi\_read}, and \texttt{read\_then\_set} receive substantial increases in synthesis allocation and all improve after training, yielding gains of $0.148$, $0.205$, and $0.103$.

The aggregate statistics further quantify this feedback-to-decision alignment. In R2$\rightarrow$R3, pre-round capability gaps are strongly aligned with next-round synthesis shifts (Spearman $\rho=0.964$), indicating that synthesis effort is systematically redirected toward regions where the learner remains weak. All directly upweighted slices improve after the corresponding training round, giving targeted improvement rates of $2/2$ for R1$\rightarrow$R2 and $3/3$ for R2$\rightarrow$R3. The positive allocation--gain association in R2$\rightarrow$R3 ($\rho=0.750$) provides additional descriptive evidence that these targeted shifts coincide with subsequent capability improvement. These slice-level results show that \textsc{EvoSyn} follows a moving capability boundary rather than repeatedly oversampling a fixed notion of task difficulty.

\begin{table*}[t]
\centering
\small
\setlength{\tabcolsep}{5.2pt}
\renewcommand{\arraystretch}{1.08}
\caption{
Targeted capability evolution on CAR-Bench Base.
We report the promoted or directly targeted slices in each transition.
$\Delta$Prop measures the change in verified synthetic-data proportion from the current round to the next round, and Gain measures the corresponding held-out success-rate change after training.
}
\label{tab:rq4_car_targeted_evolution}
\resizebox{\textwidth}{!}{%
\begin{tabular}{llccccc}
\toprule
\textbf{Transition}
& \textbf{Targeted Slice}
& \textbf{Pre-round SR}
& \textbf{Capability Gap}
& \textbf{$\Delta$Prop}
& \textbf{Post-round Gain}
& \textbf{Targeted Improvement} \\
\midrule

\multirow{2}{*}{R1 $\rightarrow$ R2}
& \texttt{multi\_write}
& 0.429 & 0.571 & +0.340 & +0.262 & \checkmark \\
& \texttt{high\_write}
& 0.333 & 0.667 & +0.145 & +0.389 & \checkmark \\
\addlinespace[2pt]

\multirow{3}{*}{R2 $\rightarrow$ R3}
& \texttt{navigation}
& 0.463 & 0.537 & +0.690 & +0.148 & \checkmark \\
& \texttt{multi\_read}
& 0.410 & 0.590 & +0.804 & +0.205 & \checkmark \\
& \texttt{read\_then\_set}
& 0.590 & 0.410 & +0.354 & +0.103 & \checkmark \\

\midrule
\multicolumn{2}{l}{R1 $\rightarrow$ R2 aggregate}
& \multicolumn{2}{c}{Allocation--Gain $\rho=0.571$}
& \multicolumn{2}{c}{Targeted Improvement Rate $=2/2$}
& \\
\multicolumn{2}{l}{R2 $\rightarrow$ R3 aggregate}
& \multicolumn{2}{c}{Gap--Allocation $\rho=0.964$}
& \multicolumn{2}{c}{Allocation--Gain $\rho=0.750$, TIR $=3/3$}
& \\

\bottomrule
\end{tabular}%
}
\end{table*}

\subsection{No-Feedback Control for Synthesis-Space Expansion}
\label{app:rq5_nofeedback_control}

We further test whether the factors introduced during synthesis-space expansion are genuinely driven by learner feedback or could be recovered through generic reasoning over the synthesis space and environment alone. For each CAR-Bench transition, we compare the feedback-conditioned Reflector with a no-feedback control that receives the same synthesis interface, environment description, and tool schema, but no post-training failures, coverage statistics, diagnostic probes, or learner-feedback summaries. As shown in Table~\ref{tab:app_rq5_nofeedback_control}, the feedback-conditioned Reflector identifies both factors subsequently adopted by \textsc{EvoSyn}: \texttt{num\_write} after R1 and \texttt{num\_read} after R2. These factors are also operationally reflected in the next-round data, with the verified \texttt{multi\_write} share increasing from $0.232$ to $0.571$ and the \texttt{multi\_read} share from $0.056$ to $0.860$.

In contrast, the no-feedback control proposes six semantically plausible dimensions, including temporal reasoning, entity resolution, constraint interaction, information synthesis, and target resolution, but none recovers the write- or read-volume factors that are actually exposed by the learner's measured failures. The resulting adopted-factor recovery is therefore $2/2$ with feedback versus $0/6$ without feedback. This control shows that synthesis-space expansion is not equivalent to asking a strong LLM to invent additional task dimensions: learner feedback identifies the specific missing coordinate along which the current task agent fails, allowing the Reflector to convert that weakness into a reusable and actionable generation control.

\begin{table*}[t]
\centering
\small
\setlength{\tabcolsep}{4.0pt}
\renewcommand{\arraystretch}{1.12}
\caption{
Detailed no-feedback control for Reflector breadth on CAR-Bench Base.
The feedback-conditioned Reflector uses empirical failure and coverage evidence, while the no-feedback control receives only the ontology, environment description, and tool schema.
}
\label{tab:app_rq5_nofeedback_control}
\resizebox{\textwidth}{!}{%
\begin{tabular}{p{0.10\linewidth} p{0.20\linewidth} p{0.17\linewidth} p{0.17\linewidth} p{0.25\linewidth} p{0.12\linewidth}}
\toprule
\textbf{Transition}
& \textbf{Empirical Feedback Signal}
& \textbf{Feedback-Conditioned Discovery}
& \textbf{Adoption Evidence}
& \textbf{No-Feedback Proposals}
& \textbf{Control Outcome} \\
\midrule

R1 $\rightarrow$ R2
&
Tasks with 5+ write actions have high failure, indicating that write-action quantity is a missing difficulty coordinate beyond the coarse tool\_use\_pattern dimension.
&
Promotes num\_write with buckets \{0, 1, 2, 3--4, 5+\}.
The factor is novel, evidence-grounded, actionable, and adopted.
&
num\_write is added to the next synthesis interface.
The verified multi\_write share increases from 0.232 to 0.571.
&
temporal\_reasoning\_depth, entity\_resolution\_complexity, constraint\_interaction.
&
Misses the adopted write-volume factor; 0/3 proposals recover the adopted write-volume factor.
\\

\addlinespace[2pt]

R2 $\rightarrow$ R3
&
Tasks with 3--4 read actions have high failure and low coverage, exposing an under-sampled read-heavy decision region.
&
Promotes num\_read with buckets \{0, 1, 2, 3--4, 5+\}.
The factor is novel, evidence-grounded, actionable, and adopted.
&
num\_read is added to the next synthesis interface.
The verified multi\_read share increases from 0.056 to 0.860.
&
temporal\_reasoning, information\_synthesis\_depth, target\_resolution.
&
Misses the adopted read-volume factor; 0/3 proposals recover the adopted read-volume factor.
\\

\midrule

\textbf{Overall}
&
\multicolumn{2}{p{0.39\linewidth}}{
Feedback-conditioned Reflector: 2 proposed factors, both evidence-grounded and both adopted.
}
&
\multicolumn{2}{p{0.42\linewidth}}{
No-feedback control: 6 proposed factors, none grounded in observed failure or coverage evidence, and none matching the adopted write/read-volume factors.
}
&
Adopted-factor recovery: 2/2 vs.\ 0/6.
\\

\bottomrule
\end{tabular}%
}
\end{table*}

\subsection{Controllability of Evolved Synthesis Factors}
\label{app:factor_controllability}

A revisable synthesis space is useful only if its factors serve as actionable generation controls rather than post-hoc descriptors of generated data. We therefore test whether a factor value specified before generation is faithfully realized in the resulting environment-grounded trajectory. We use \texttt{num\_write} as a representative directly measurable factor: for each blueprint--task pair, we recover the requested value from the synthesis plan and independently recompute the realized value from the ground-truth tool trajectory. To isolate generation controllability from round-wise evolution, we evaluate R2 synthesis runs across multiple domains and synthesis backbones. As shown in Table~\ref{tab:factor_controllability}, requested and realized values exhibit near-perfect agreement in every setting, with exact-match accuracy of $99.6$--$100\%$ and Spearman $\rho=0.997$--$1.000$ across thousands of blueprint--task pairs. Importantly, this agreement is not an artifact of the Verifier filtering non-conforming generations: the explicit axis-conformance check rejects only $0.15\%$ of candidates in Retail and $1.0\%$ in CAR-Bench, while candidates rejected for unrelated reasons still retain requested--realized agreement of $100\%$ and $98.3\%$, respectively. These results show that the synthesis pipeline can reliably translate an abstract factor specification into the intended interaction structure, supporting the synthesis space as an operational control interface for feedback-driven allocation and evolution.

\begin{table}[t]
\centering
\small
\setlength{\tabcolsep}{6.0pt}
\renewcommand{\arraystretch}{1.08}
\caption{
Controllability of the synthesis coordinate across R2 synthesis runs.
We compare the value requested in the synthesis plan with the value independently recomputed from the realized ground-truth trajectory.
}
\label{tab:factor_controllability}
\begin{tabular}{llcc}
\toprule
\textbf{Domain} &
\textbf{Synthesis LLM} &
\textbf{Exact Match} &
\textbf{Spearman $\rho$} \\
\midrule
CAR-Bench & Claude Opus 4.5 & 99.6\%  & 0.998 \\
Retail    & Claude Opus 4.5 & 100.0\% & 1.000 \\
Airline   & Qwen3-30B       & 100.0\% & 1.000 \\
CAR-Bench & Qwen3-235B      & 99.7\%  & 0.997 \\
\bottomrule
\end{tabular}
\end{table}

\subsection{Capability-Gap Alignment of Synthesis Reallocation}
\label{app:gap_alignment}

\begin{figure}[t]
    \centering
    \includegraphics[width=0.7\linewidth]{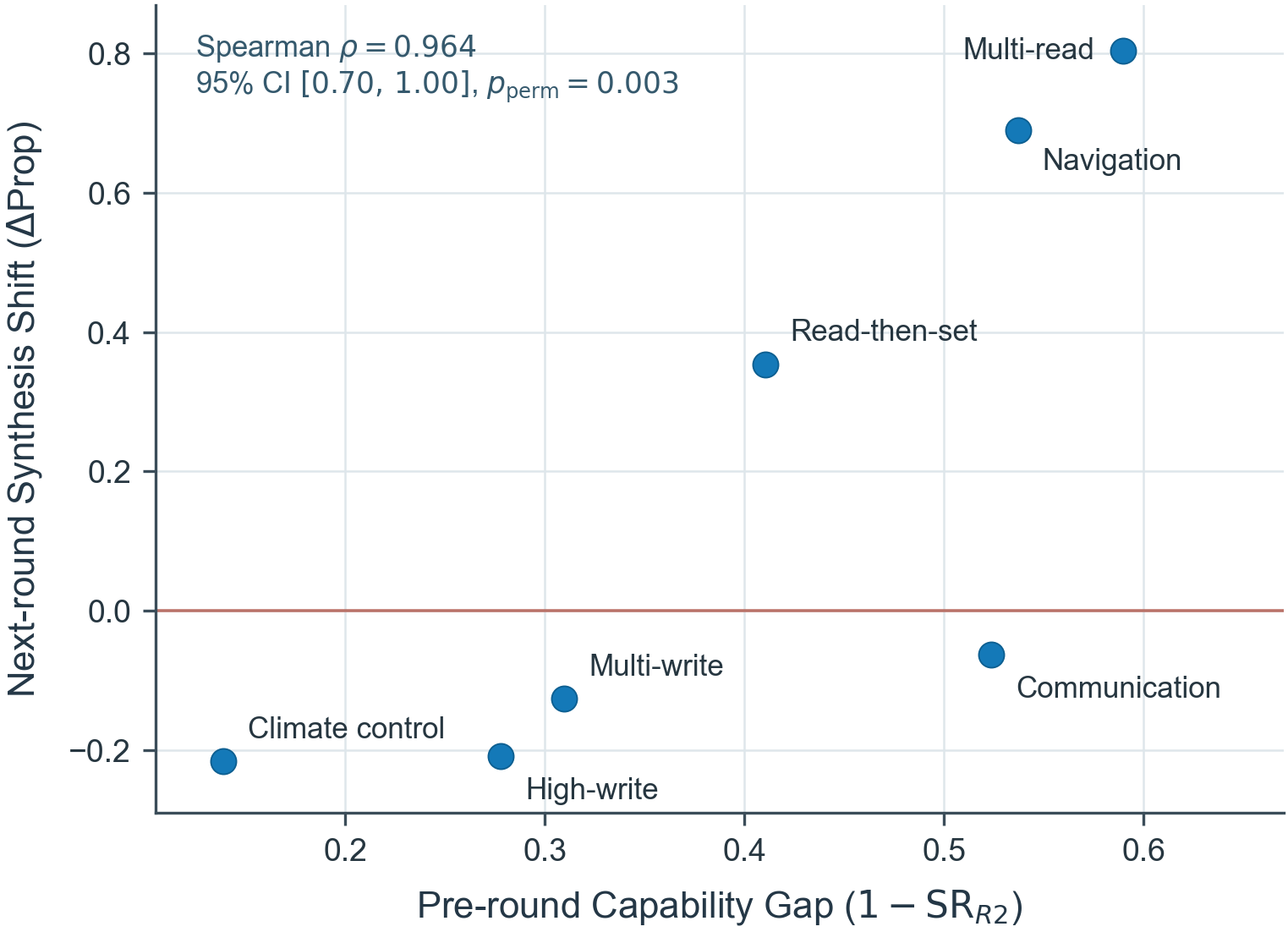}
    \caption{
    Capability-gap-driven synthesis reallocation on CAR-Bench from R2 to R3.
    Each point denotes an interaction slice. The pre-round capability gap
    ($1-\mathrm{SR}_{R2}$) is strongly aligned with the subsequent change in
    synthesis allocation (Spearman $\rho=0.964$, 95\% CI $[0.70,1.00]$,
    permutation $p=0.003$), showing that \textsc{EvoSyn} shifts synthesis effort
    toward regions where the current learner remains weak.
    }
    \label{fig:gap_allocation}
\end{figure}

We further examine whether feedback-driven synthesis reallocation is systematically aligned with the learner's residual capability gaps rather than being driven by a small number of selected cases. We focus on the CAR-Bench R2$\rightarrow$R3 transition and evaluate all seven measurable interaction slices. For each slice, we define the pre-round capability gap as $1-\mathrm{SR}_{R2}$ and quantify the next-round synthesis shift by the change in its proportion of verified synthetic data. As shown in Figure~\ref{fig:gap_allocation}, the two quantities are strongly correlated (Spearman $\rho=0.964$, 95\% CI $[0.70,1.00]$, permutation $p=0.003$). Interaction slices with larger residual gaps, such as \texttt{multi\_read} and \texttt{navigation}, receive substantially greater synthesis emphasis, whereas regions in which the learner is already stronger are generally downweighted. In contrast, under the Fixed-State Synthesis baseline, the same learner-side capability gaps are negatively correlated with subsequent changes in data proportion ($\rho=-0.964$), indicating that simply accumulating additional synthetic data does not reproduce this learner-aligned redistribution. The observed reallocation is also consistent with subsequent learning outcomes. All three slices that receive increased synthesis emphasis in R3, namely \texttt{multi\_read}, \texttt{navigation}, and \texttt{read\_then\_set}, improve after training by $+0.205$, $+0.148$, and $+0.103$, respectively. These results provide direct evidence that \textsc{EvoSyn} translates residual learner weaknesses into targeted synthesis decisions and continuously redirects future experience as the learner's capability boundary changes.

\subsection{Structural Support Expansion Across Synthesis Rounds}
\label{app:structural_support}

We next examine whether synthesis-space evolution expands the structural support of the generated experience or simply produces additional samples within an already covered interaction regime. To abstract away surface-level linguistic variation, we represent each verified example by its ground-truth tool-call sequence and measure the cumulative number of distinct interaction sequences. As shown in Table~\ref{tab:structural_support}, \textsc{EvoSyn} exhibits sustained structural growth across synthesis rounds, whereas Static synthesis increasingly saturates. On CAR-Bench, the number of distinct sequences grows from 270 in R1 to 497 in R2 and 952 in R3 under \textsc{EvoSyn}, while Static reaches only 415 by R3. The contrast is even sharper on Airline, where \textsc{EvoSyn} expands from 28 to 111 and 122 sequences across rounds, compared with only 36 for Static at R3. This divergence indicates that later \textsc{EvoSyn} rounds continue to introduce new executable interaction structures rather than repeatedly sampling from the support established in earlier rounds.

\begin{figure}[t]
    \centering
    \includegraphics[width=0.65\linewidth]{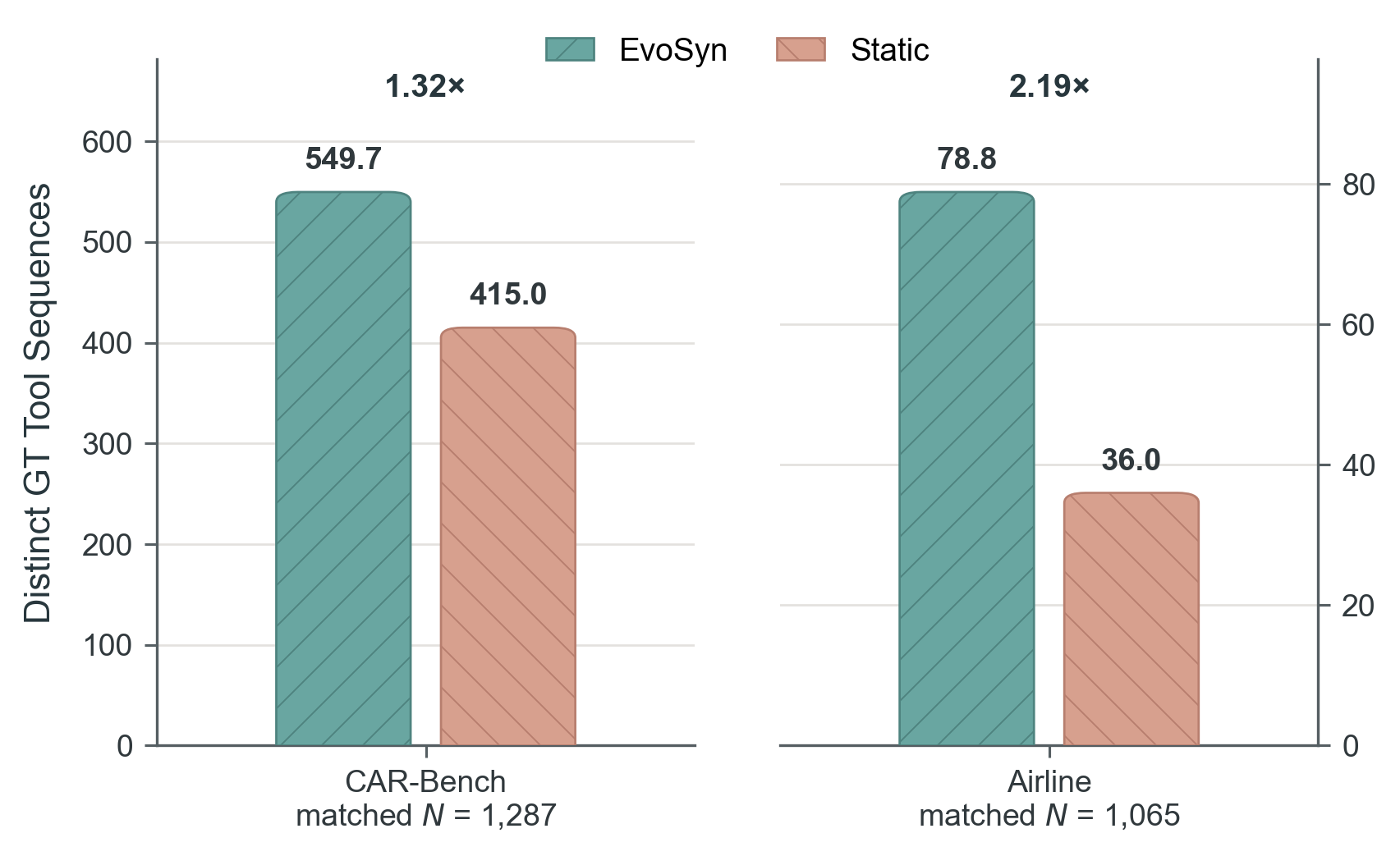}
    \caption{
    Size-matched structural support at the final synthesis round.
    The cumulative \textsc{EvoSyn} pool is subsampled to match the number of
    accepted examples in the Static baseline, and structural support is measured
    by the number of distinct ground-truth tool-call sequences.
    After controlling for data volume, \textsc{EvoSyn} still covers
    $1.32\times$ more interaction sequences on CAR-Bench and
    $2.19\times$ more on Airline.
    }
    \label{fig:size_matched_support}
\end{figure}

Because the cumulative \textsc{EvoSyn} pools contain more accepted examples, we further control for data volume through size-matched subsampling at the final round. For each domain, we subsample the \textsc{EvoSyn} pool to exactly match the number of accepted examples in Static and recompute structural support. Figure~\ref{fig:size_matched_support} shows that the advantage remains substantial: at the same $N=1{,}287$, \textsc{EvoSyn} covers 549.7 distinct sequences on CAR-Bench compared with 415.0 for Static, a $1.32\times$ increase. On Airline, it covers 78.8 sequences versus 36.0 at the same $N=1{,}065$, corresponding to a $2.19\times$ increase. The persistence of this gap after controlling for sample count shows that the structural expansion cannot be explained by data accumulation alone. Instead, revising the synthesis state changes the support of the generated experience, enabling controllable interaction factors to be realized through a broader set of executable behaviors.

\begin{table}[t]
\centering
\small
\setlength{\tabcolsep}{7pt}
\renewcommand{\arraystretch}{1.08}
\caption{
Structural-support growth across synthesis rounds.
We report the cumulative number of distinct ground-truth tool-call sequences
covered by the verified synthetic data.
}
\label{tab:structural_support}
\begin{tabular}{llccc}
\toprule
\textbf{Domain} &
\textbf{Method} &
\textbf{R1} &
\textbf{R2} &
\textbf{R3} \\
\midrule
CAR-Bench
& \textsc{EvoSyn} & 270 & 497 & \textbf{952} \\
& Static          & 276 & 398 & 415 \\
\midrule
Airline
& \textsc{EvoSyn} & 28  & 111 & \textbf{122} \\
& Static          & 23  & 27  & 36 \\
\bottomrule
\end{tabular}
\end{table}

\subsection{Do Promoted Factors Capture Previously Unmodeled Learner Weakness?}
\label{app:factor_failure}

We next examine whether factors introduced during synthesis-space expansion capture genuine learner weaknesses that are not adequately represented by the current synthesis space. For each promoted factor, we retrospectively annotate the corresponding interaction property on held-out evaluation tasks and analyze the learner \emph{before} that factor is added, so that the diagnosis is not confounded by subsequent factor-guided synthesis or training. To quantify the additional failure signal carried by the promoted coordinate, we compute the conditional mutual information $I(Y_{\mathrm{fail}}; d_{\mathrm{new}} \mid \mathcal{O}_r)$, where $\mathcal{O}_r$ denotes the synthesis space before promotion. This quantity measures how much information about learner failure remains attributable to the new factor after conditioning on the interaction factors already represented in the current synthesis space.

As shown in Table~\ref{tab:factor_failure}, the promoted factors retain substantial additional failure information across representative expansion cases. On Airline, \texttt{num\_write} contributes 0.122 nats beyond the existing synthesis coordinates. On CAR-Bench, the corresponding values are 0.168 nats for \texttt{num\_write} and 0.134 nats for \texttt{num\_read}. These positive values indicate that the promoted factors are not redundant with the pre-existing synthesis representation. Instead, they capture residual failure-relevant structure that remains visible even after accounting for the factors already available to the synthesizer.

\begin{figure}[t]
    \centering
    \includegraphics[width=0.6\linewidth]{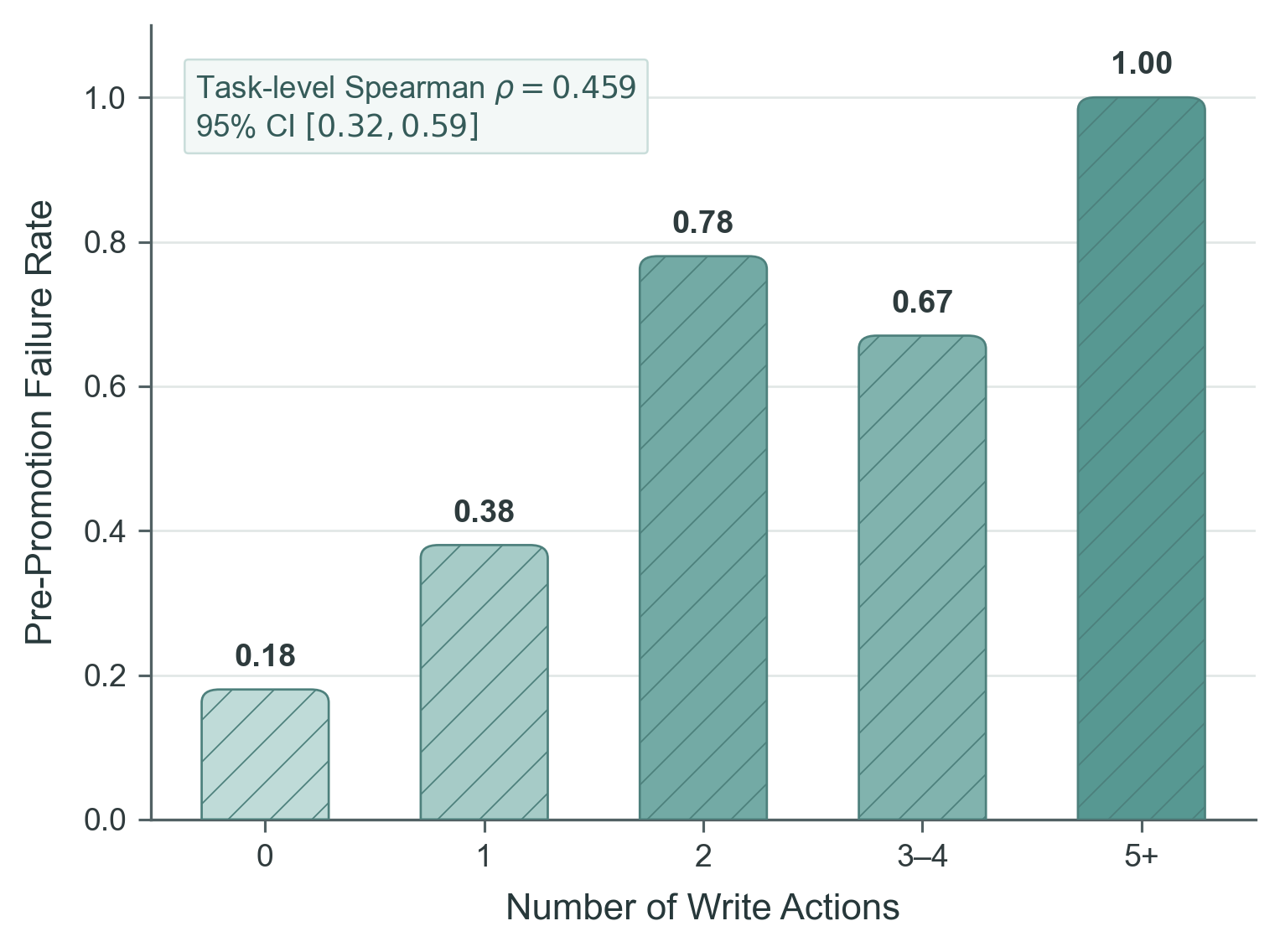}
    \caption{
    Pre-promotion learner failure rate across \texttt{num\_write} buckets on
    Airline. Tasks requiring more state-changing write actions are substantially
    more failure-prone before \texttt{num\_write} is introduced as an explicit
    synthesis coordinate, with a positive task-level association between write
    complexity and learner failure (Spearman $\rho=0.459$, 95\% CI
    $[0.32,0.59]$).
    }
    \label{fig:airline_num_write_failure}
\end{figure}

Figure~\ref{fig:airline_num_write_failure} provides an intuitive illustration on Airline. Before \texttt{num\_write} is introduced as an explicit synthesis coordinate, the learner's failure rate rises sharply with the number of required write actions, increasing from 0.18 for tasks with no write action to 1.00 for tasks with five or more, with a positive task-level rank association of $\rho=0.459$. Because all measurements are taken from the pre-promotion learner, this pattern reflects a capability gap that precedes and motivates the later synthesis-space revision. The result therefore supports a stronger interpretation than post-hoc relabeling: promoted factors expose previously unmodeled failure structure and convert it into actionable synthesis coordinates for subsequent rounds.

\begin{table}[t]
\centering
\small
\setlength{\tabcolsep}{7pt}
\renewcommand{\arraystretch}{1.08}
\caption{
Residual learner-failure information captured by promoted synthesis factors.
All statistics are computed using the learner before the corresponding factor is
introduced. Conditional mutual information measures the additional failure
information provided by the promoted factor after conditioning on the existing
synthesis space $\mathcal{O}_r$.
}
\label{tab:factor_failure}
\begin{tabular}{llc}
\toprule
\textbf{Domain} &
\textbf{Promoted Factor} &
$\mathbf{I(Y_{\mathrm{fail}};d_{\mathrm{new}}\mid\mathcal{O}_r)}$ \textbf{(nats)} \\
\midrule
Airline   & \texttt{num\_write} & 0.122 \\
CAR-Bench & \texttt{num\_write} & \textbf{0.168} \\
CAR-Bench & \texttt{num\_read}  & 0.134 \\
\bottomrule
\end{tabular}
\end{table}

\subsection{Do Promoted Factors Capture Failure Structure Beyond Generic Complexity?}
\label{app:factor_complexity}

A natural alternative explanation for synthesis-space expansion is that the promoted factors merely recover generic task difficulty. For example, tasks requiring more reads or writes may simply involve longer tool-use trajectories and therefore be harder for the learner. We test this hypothesis by using the total number of tool calls, discretized into tertile-based buckets, as a domain-agnostic complexity variable $g$. For each promoted factor $z$, we measure its dependence on generic complexity using $\mathrm{NMI}(z,g)$ and quantify the residual failure information it provides after conditioning on $g$ through $I(Y_{\mathrm{fail}};z\mid g)$.

As shown in Table~\ref{tab:factor_complexity}, Airline \texttt{num\_write} is only weakly associated with overall tool count ($\mathrm{NMI}=0.216$), yet retains $0.241$ nats of failure information after generic complexity is known, compared with only $0.032$ nats carried by tool count itself. CAR-Bench \texttt{num\_read} shows a moderate association with tool count ($\mathrm{NMI}=0.420$), but still contributes $0.075$ nats of residual failure information beyond it. These results show that the promoted factors are not redundant proxies for trajectory length or overall interaction complexity. Instead, they identify more specific failure-relevant interaction properties, supporting synthesis-space expansion as a mechanism for exposing structured capability gaps rather than merely generating globally harder tasks.

\begin{table}[t]
\centering
\small
\setlength{\tabcolsep}{6.5pt}
\renewcommand{\arraystretch}{1.08}
\caption{
Promoted synthesis factors capture learner-failure structure beyond generic
interaction complexity. We use the total number of tool calls as a
domain-agnostic complexity variable $g$. All information quantities are reported
in nats.
}
\label{tab:factor_complexity}
\begin{tabular}{llccc}
\toprule
\textbf{Domain} &
\textbf{Promoted Factor $z$} &
$\mathbf{\mathrm{NMI}(z,g)}$ &
$\mathbf{I(Y_{\mathrm{fail}};g)}$ &
$\mathbf{I(Y_{\mathrm{fail}};z\mid g)}$ \\
\midrule
Airline
& \texttt{num\_write}
& 0.216
& 0.032
& \textbf{0.241} \\

CAR-Bench
& \texttt{num\_read}
& 0.420
& 0.109
& \textbf{0.075} \\
\bottomrule
\end{tabular}
\end{table}

\subsection{Observed Marginal Learning Gain Across Synthesis Rounds}
\label{app:marginal_utility}

We further examine how downstream improvement changes as new verified experience is introduced across synthesis rounds. Because both the learner state and training-data composition evolve over time, this analysis is descriptive rather than a causal estimate of per-example utility. We normalize the observed change in task success by the amount of newly added verified training data. As shown in Table~\ref{tab:marginal_utility}, the CAR-Bench R1$\rightarrow$R2 transition improves success rate by $+0.020$ with approximately $0.97$K additional examples, corresponding to an observed gain of $+0.021$ per 1K examples. In comparison, R2$\rightarrow$R3 improves success by $+0.120$ with only $0.78$K additional examples, yielding $+0.153$ per 1K examples. This is approximately seven times the normalized gain observed in the preceding round. Notably, R2$\rightarrow$R3 is also the transition in which synthesis is redirected toward the learner's newly exposed read-heavy capability gaps, linking the strongest synthesis-space revision with the largest observed learning return from newly added experience.

The endpoint comparison provides a complementary view after accounting for cumulative training-set size. Relative to the base Qwen3-8B agent, \textsc{EvoSyn} R3 gains $0.400$ success-rate points using 1,996 verified examples, corresponding to $+0.200$ per 1K examples, compared with $+0.210$ and $+0.163$ per 1K for Static. While these ratios should not be interpreted as intrinsic sample-level utility, they provide additional evidence that feedback-guided synthesis can improve not only the amount and structural coverage of training experience, but also the observed learning return associated with later synthesis rounds.

\begin{table}[t]
\centering
\small
\setlength{\tabcolsep}{6pt}
\renewcommand{\arraystretch}{1.08}
\caption{
Observed marginal learning gain on CAR-Bench.
Round-wise gain is normalized by the number of newly added verified examples;
endpoint efficiency is measured relative to the base Qwen3-8B agent
($\mathrm{SR}=0.360$).
}
\label{tab:marginal_utility}
\begin{tabular}{lccc}
\toprule
\textbf{Transition} &
$\mathbf{\Delta}$\textbf{SR} &
\textbf{Added Examples} &
$\mathbf{\Delta}$\textbf{SR / 1K} \\
\midrule
R1$\rightarrow$R2 & +0.020 & 0.97K & +0.021 \\
R2$\rightarrow$R3 & \textbf{+0.120} & 0.78K & \textbf{+0.153} \\
\midrule
\multicolumn{4}{c}{\textit{Final endpoint efficiency relative to the base agent}} \\
\midrule
\textsc{EvoSyn} R3 & +0.400 & 1.996K & \textbf{+0.200} \\
Static R3          & +0.210 & 1.287K & +0.163 \\
\bottomrule
\end{tabular}
\end{table}

\subsection{Evolution Case Study}

\subsubsection{Car-Bench}
\label{app:carbench_case}

Table~\ref{tab:rq2_data_quality_carbench} quantitatively verifies the multi-round evolution illustrated in Figure~\ref{fig:carbench_data_quality_rounds}.
Both Static Multi-Round and \textsc{EvoSyn} start from the same R1 data, isolating the effect of feedback-driven synthesis decisions in later rounds.
Static Multi-Round continues to add data under a fixed synthesis interface and allocation, but its downstream performance does not improve consistently, moving from 0.607 to 0.627 and then dropping to 0.613.
In contrast, \textsc{EvoSyn} improves monotonically from 0.607 to 0.655 and finally to 0.726, suggesting that later rounds become increasingly useful for task-agent learning rather than merely larger in size.
The intrinsic metrics show the same pattern.
By R3, \textsc{EvoSyn} achieves substantially higher factor coverage than Static Multi-Round (0.893 vs. 0.787) and much higher trajectory-pattern diversity (0.938 vs. 0.750), while also improving factor balance and interaction coverage.
These results support the mechanism behind \textsc{EvoSyn}: post-training feedback does not only evaluate the current synthetic pool, but changes the controllable synthesis interface and redirects subsequent generation toward interaction regions that static sampling fails to cover effectively.

\begin{table*}[t]
\centering
\small
\setlength{\tabcolsep}{5.2pt}
\renewcommand{\arraystretch}{1.08}
\caption{
Intrinsic quality of synthetic data over rounds on CAR-Bench Base.
Each row is computed on the cumulative data up to the corresponding round.
All intrinsic metrics are normalized to $[0,1]$, where higher is better.
}
\label{tab:rq2_data_quality_carbench}
\resizebox{\textwidth}{!}{%
\begin{tabular}{llccccc}
\toprule
\textbf{Setting}
& \textbf{Round}
& \makecell{\textbf{Factor}\\\textbf{Coverage}}
& \makecell{\textbf{Factor}\\\textbf{Balance}}
& \makecell{\textbf{Interaction}\\\textbf{Coverage}}
& \makecell{\textbf{Trajectory Pattern}\\\textbf{Diversity}}
& \textbf{Success Rate} \\
\midrule

\multirow{3}{*}{Static Multi-Round}
& R1 & 0.640 & 0.492 & 0.339 & 0.688 & 0.607 \\
& R2 & 0.705 & 0.470 & 0.343 & 0.735 & 0.627 \\
& R3 & 0.787 & 0.468 & 0.351 & 0.750 & 0.613 \\

\midrule

\multirow{3}{*}{\textsc{EvoSyn}}
& R1 & 0.640 & 0.492 & 0.339 & 0.688 & 0.607 \\
& R2 & 0.760 & 0.479 & 0.351 & 0.750 & 0.655 \\
& R3 & 0.893 & 0.508 & 0.373 & 0.938 & 0.726 \\

\bottomrule
\end{tabular}%
}
\end{table*}

Table~\ref{tab:rq2_evolution_trace_carbench} provides the detailed evidence behind the CAR-Bench evolution trace shown in Figure~\ref{fig:carbench_round_evolution}.
The table expands each round transition into four components: the post-training feedback observed after the previous round, the Reflector's ontology decision, the Designer's next-round synthesis focus, and the resulting synthesis change.
In the R1$\rightarrow$R2 transition, the key signal is a failure concentration in write-heavy vehicle-control tasks: the 5+ write bucket has a failure rate of 0.92, while the initial interface only represents coarse tool-use patterns and cannot directly control the number of state-changing actions.
The Reflector therefore promotes \texttt{num\_write}, allowing the Designer to allocate synthesis budget toward high-write trajectories.
In the R2$\rightarrow$R3 transition, feedback shows that the write dimension has become meaningful, but a new gap appears in read-heavy tasks: the \texttt{num\_read}=3--4 bucket has a failure rate of 0.63 while covering only 3.9\% of the data.
The Reflector consequently promotes \texttt{num\_read}, shifting the next round toward read-heavy trajectories.
This detailed trace supports the interpretation that \textsc{EvoSyn} evolves by converting post-training feedback into new controllable synthesis dimensions, rather than merely resampling more tasks from a fixed interface.

\begin{table*}[t]
\centering
\small
\setlength{\tabcolsep}{4.2pt}
\renewcommand{\arraystretch}{1.12}
\caption{
Mechanistic trace of learning-guided synthesis evolution on CAR-Bench Base.
Each row reports the concrete feedback evidence used by the Reflector, the resulting ontology decision, the Designer's budget shift, and the observed synthesis change.
}
\label{tab:rq2_evolution_trace_carbench}
\resizebox{\textwidth}{!}{%
\begin{tabular}{p{0.10\linewidth} p{0.27\linewidth} p{0.20\linewidth} p{0.20\linewidth} p{0.19\linewidth}}
\toprule
\textbf{Transition}
& \textbf{Feedback Evidence}
& \textbf{Reflector Decision}
& \textbf{Designer Focus}
& \textbf{Observed Synthesis Change} \\
\midrule

R1 $\rightarrow$ R2
&
R1 evaluation shows that the hardest cases concentrate in high-write vehicle-control tasks: the 5+ write bucket has failure rate 0.92. 
The R1 data is skewed toward lower write counts (1 write: 54\%; 2 writes: 14\%; 3--4 writes: 16\%; 5+ writes: 16\%). 
The existing \texttt{tool\_use\_pattern} dimension captures action-sequence shape, but not the quantity of state-changing writes.
&
Promote \texttt{num\_write\_actions} into a new controllable dimension \texttt{num\_write}. 
This separates write-action quantity from the existing tool-pattern dimension and introduces a new axis for execution-heavy tasks.
&
Increase synthesis budget for high-write regions, especially \texttt{num\_write}=3--4 and 5+, under the base-only CAR environment with a weak macro prior.
&
R2 generates multi-write vehicle-control tasks that the initial interface could not directly target. 
The round accepts 971 verified tasks, bringing the cumulative verified set to 1212.
\\

\addlinespace[2pt]

R2 $\rightarrow$ R3
&
After R2, the \texttt{num\_write} axis is validated: failure on 5+ write tasks drops from 0.92 to 0.53, and the failure gradient becomes monotonic across write-count buckets. 
New feedback shows that read-heavy tasks remain both difficult and under-covered: \texttt{num\_read}=3--4 has failure rate 0.63 but only 3.9\% coverage.
&
Promote \texttt{num\_write} from experimental to standard. 
Promote \texttt{num\_read} into a new controllable axis with buckets 0/1/2/3--4/5+. 
The new axis is considered orthogonal to \texttt{num\_write}, since read actions correspond to information gathering and decision preparation, while write actions correspond to execution.
&
Shift macro budget toward the under-covered high-failure read region, especially \texttt{num\_read}=3--4, while keeping \texttt{num\_write} as a standard seed dimension.
&
R3 produces read-heavy trajectories: the average number of reads increases to 5.02 per task, compared with 0.97 in R2. 
The round accepts 784 verified tasks and expands the data toward read--decide--act interactions.
\\

\bottomrule
\end{tabular}%
}
\end{table*}

\subsubsection{Tau-Bench Airline}
\label{app:airline_case}

The Airline domain provides a complementary case of learning-guided synthesis evolution in a customer-service environment with richer policy constraints and state-changing operations. 
As shown in Table~\ref{tab:rq2_evolution_trace_airline}, \textsc{EvoSyn} first identifies a strong distributional and capability mismatch after R1: the verified data is dominated by \texttt{flight\_change} tasks, while high-write interactions remain under-represented and difficult. 
The Reflector therefore promotes \texttt{num\_write} into a controllable dimension, allowing the Designer to explicitly allocate R2 synthesis budget toward multi-write trajectories. 
After this axis becomes operational, the next round reveals a different source of difficulty: numerical or derived tool arguments that require computation before execution. 
The Reflector then promotes \texttt{computation} as a new synthesis dimension, and R3 shifts toward more balanced coverage of computation-aware interactions. 
This trace supports the same mechanism observed in CAR-Bench: feedback does not merely filter generated tasks, but changes the coordinates along which future tasks can be deliberately synthesized.

Table~\ref{tab:rq2_data_quality_airline} further shows that this mechanism translates into stronger multi-round behavior than Static Multi-Round synthesis. 
Both methods start from the same R1 data, but Static Multi-Round only accumulates more samples under a fixed interface and allocation, leading to unstable performance from 0.648 to 0.633 and then 0.660. 
In contrast, \textsc{EvoSyn} improves steadily from 0.648 to 0.693 and finally to 0.713. 
The intrinsic data-quality metrics show a similar divergence: by R3, \textsc{EvoSyn} achieves higher factor coverage, factor balance, interaction coverage, complexity coverage, and trajectory-pattern diversity than the static baseline. 
The largest gap appears in trajectory-pattern diversity, where \textsc{EvoSyn} reaches 0.769 compared with 0.538 for Static Multi-Round, indicating that feedback-guided evolution expands not only the labels attached to synthetic tasks but also the underlying tool-use and state-transition structures. 
Together with the CAR-Bench results in the main text, this Airline case suggests that the self-evolution loop generalizes across domains: later rounds become more useful because the synthesis process itself is revised according to observed failures, coverage gaps, and newly exposed interaction factors.

\begin{table*}[t]
\centering
\small
\setlength{\tabcolsep}{4.8pt}
\renewcommand{\arraystretch}{1.08}
\caption{
Intrinsic quality of synthetic data over rounds on $\tau^2$-Bench Airline.
Each row is computed on the cumulative data up to the corresponding round.
All metrics are normalized to $[0,1]$, where higher is better.
}
\label{tab:rq2_data_quality_airline}

\resizebox{\textwidth}{!}{%
\begin{tabular}{llccccc}
\toprule
\textbf{Setting}
& \textbf{Round}
& \makecell{\textbf{Factor}\\\textbf{Coverage}}
& \makecell{\textbf{Factor}\\\textbf{Balance}}
& \makecell{\textbf{Interaction}\\\textbf{Coverage}}
& \makecell{\textbf{Trajectory Pattern}\\\textbf{Diversity}}
& \textbf{Success Rate} \\
\midrule

\multirow{3}{*}{Static Multi-Round}
& R1 & 0.742 & 0.264 & 0.311 & 0.462 & 0.648 \\
& R2 & 0.760 & 0.309 & 0.330 & 0.538 & 0.633 \\
& R3 & 0.778 & 0.368 & 0.403 & 0.538 & 0.660 \\

\midrule

\multirow{3}{*}{\textsc{EvoSyn}}
& R1 & 0.742 & 0.264 & 0.311 & 0.462 & 0.648 \\
& R2 & 0.785 & 0.453 & 0.399 & 0.692 & 0.693 \\
& R3 & 0.891 & 0.450 & 0.455 & 0.769 & 0.713 \\

\bottomrule
\end{tabular}%
}
\end{table*}

\begin{table*}[t]
\centering
\small
\setlength{\tabcolsep}{4.2pt}
\renewcommand{\arraystretch}{1.12}
\caption{
Mechanistic trace of learning-guided synthesis evolution on $\tau^2$-Bench Airline.
Each row reports the concrete feedback evidence used by the Reflector, the resulting ontology decision, the Designer's budget shift, and the observed synthesis change.
}
\label{tab:rq2_evolution_trace_airline}
\resizebox{\textwidth}{!}{%
\begin{tabular}{p{0.10\linewidth} p{0.27\linewidth} p{0.20\linewidth} p{0.20\linewidth} p{0.19\linewidth}}
\toprule
\textbf{Transition}
& \textbf{Feedback Evidence}
& \textbf{Reflector Decision}
& \textbf{Designer Focus}
& \textbf{Observed Synthesis Change} \\
\midrule

R1 $\rightarrow$ R2
&
R1 data is highly skewed: \texttt{flight\_change} accounts for 93.6\% of verified tasks. 
Failure probes further show that the 3--4 write-action bucket has failure rate 1.00, indicating that high-write state-changing interactions are difficult but not explicitly controllable in the initial interface.
&
Promote \texttt{num\_write\_actions} into a new controllable axis \texttt{num\_write} with buckets 0/1/2/3--4/5+. 
The axis is introduced as a failure-driven experimental dimension because the probe shows strong failure correlation and is generator-realizable.
&
Increase macro budget for the previously under-covered high-write buckets, especially \texttt{num\_write}=3--4 and 5+. 
Lower write-count buckets remain medium priority.
&
R2 blueprints shift strongly toward high-write synthesis: the 3--4 and 5+ buckets together account for 57\% of realized blueprints. 
Accepted tasks span all five write-count buckets, showing that \texttt{num\_write} becomes an operational synthesis control rather than only a diagnostic label.
\\

\addlinespace[2pt]

R2 $\rightarrow$ R3
&
After R2, \texttt{num\_write} is validated as a meaningful factor: the 5+ write bucket remains difficult with failure rate 0.86, while write-count coverage becomes healthier. 
A new probe, \texttt{num\_numeric\_fields}, shows failure rate 1.00 at value 1, revealing difficulty with numerical or derived tool arguments before execution.
&
Promote \texttt{num\_write} from experimental to standard. 
Then promote \texttt{num\_numeric\_fields} into a new semantic axis \texttt{computation}, with values \texttt{none}, \texttt{aggregate}, \texttt{constrained\_opt}, and \texttt{conditional}.
&
Tilt macro budget toward all four buckets of the new \texttt{computation} axis. 
The goal is to systematically explore different forms of computation-before-action behavior while retaining \texttt{num\_write} as a standard dimension.
&
R3 blueprints become approximately balanced across computation types, with each computation bucket covering about one quarter of the realized blueprint distribution. 
The synthesis focus therefore shifts from only controlling write-action complexity to also covering numerical decision-making and conditional tool execution.
\\

\bottomrule
\end{tabular}%
}
\end{table*}

\section{Related Work}
\label{app:related_work}

\subsection{Synthetic Data for Post-Training and Self-Evolution}
\label{app:rw_synthetic_self_evolution}

Synthetic data has become a major mechanism for expanding the supervision available to large language models beyond manually curated datasets.
A broad line of work studies how to bootstrap instruction-following data from seed tasks, teacher models, unlabeled corpora, or model-generated responses, including Self-Instruct~\citep{wang2023self}, WizardLM~\citep{xu2024wizardlm}, and instruction backtranslation~\citep{li2023self}.
Subsequent methods improve the diversity, realism, and usefulness of synthetic corpora through multi-agent simulation, feature-targeted generation, filtering, or autonomous data engineering~\citep{tang2025synthesizing,li2026less,cao2025condor,he2025fine,tu2025resofilter}.
These approaches establish synthetic data as a central ingredient in LLM post-training.
However, most of them treat data generation as an offline construction process: a corpus is generated, filtered, and then used to train a learner, while the synthesis process itself is only weakly coupled to the learner's post-training behavior.

Reasoning domains provide a particularly favorable setting for scalable synthetic supervision because many examples admit verifiable answers or checkable solution traces.
STaR~\citep{zelikman2022star} bootstraps rationales from successful model attempts, while WizardMath~\citep{luo2025wizardmath}, MetaMath~\citep{yu2023metamath}, MAmmoTH~\citep{yue2024mammoth}, and ToRA~\citep{gou2024tora} expand mathematical or tool-integrated reasoning supervision through instruction evolution, problem transformation, and rationale construction.
Large-scale mathematical post-training pipelines such as DeepSeekMath~\citep{shao2024deepseekmath} and NuminaMath~\citep{li2024numinamath} further demonstrate the effectiveness of combining synthesized or curated reasoning data with outcome verification.
More recent learner-aware methods use model behavior to guide synthesis: for example, SwS~\citep{liang2026sws} identifies persistent weaknesses and generates new problems around the corresponding concepts.
Together, these works show that verifiability and learner feedback can substantially improve synthetic data quality.
Nevertheless, their feedback signals are usually mapped to harder problems, corrected rationales, or concept-level variants within a relatively self-contained problem space.

A related line of work studies self-evolving LLMs, where task generation and task solving are placed in a feedback loop~\citep{yang2025toward}.
R-Zero~\citep{huang2025r} co-evolves a Challenger and Solver through generated problems and solution feedback, and Socratic-Zero~\citep{wang2025socratic} introduces Teacher, Solver, and Generator roles to construct an increasingly challenging curriculum from diagnosed learner weaknesses.
Agent0~\citep{xia2025agent0}, Dr. Zero~\citep{yue2026dr}, and SCOPE~\citep{kwan2026scope} extend this paradigm to tool-integrated reasoning, multi-turn search, and document-grounded task construction.
These methods demonstrate that data generation can be coupled with learner improvement rather than treated as a fixed preprocessing step.
Our work shares this closed-loop perspective, but focuses on a different challenge: in interactive agentic environments, useful synthesis must not only generate harder examples, but also discover and control the interaction factors that determine feasibility, tool execution, state transitions, and downstream learning utility.

\subsection{Feedback-Driven Agentic Data Synthesis}
\label{app:rw_agentic_synthesis}

Synthetic data for interactive agents introduces constraints that are less prominent in self-contained reasoning tasks.
A generated example must correspond to a feasible environment state, admit an executable sequence of tool calls, satisfy domain policies, and lead to a verifiable outcome.
Recent environment-grounded synthesis methods address these requirements by using exploration, trajectory reconstruction, or state replay to build executable agentic experience.
Explorer~\citep{pahuja2025explorer} synthesizes diverse web trajectories through large-scale environment exploration, OS-Genesis~\citep{sun2025genesis} derives executable GUI tasks from successful interactions, AutoPlay~\citep{ramrakhya2025scaling} explores application states before producing grounded tasks, and Trajectory2Task~\citep{wang2026trajectory2task} converts valid tool trajectories into user-facing tasks.
WebAggregator~\citep{wang2026webaggregator} further constructs search tasks requiring compositional evidence aggregation.
These methods make important progress on feasibility and executability, but their feedback is primarily used to obtain valid instances or broaden trajectory coverage, rather than to revise an explicit synthesis control space according to the trained agent's learning response.

Other work improves the quality of tool-use and multi-turn synthetic data through checkers, validators, and data-centric refinement.
EigenData~\citep{chen2026eigendata} synthesizes multi-turn tool-use dialogues with executable checkers, while APIGen and APIGen-MT pipelines construct function-calling or tool-use supervision with varying degrees of automatic validation.
DreamGym~\citep{chen2025scaling} scales synthetic tasks and interactions through a learned experience model.
These approaches show that agentic data generation benefits from structured procedures, executable verification, and iterative data-quality control.
However, validation and repair mainly answer whether a generated sample is well formed, executable, or diverse.
They do not by themselves determine which interaction factors should receive more synthesis budget after training, or whether the current set of factors is sufficient to explain the remaining weaknesses of the task agent.

Several recent frameworks further explore continual improvement in agentic settings.
CoEvolve~\citep{yang2026coevolve} uses trajectory-derived signals such as forgetting and uncertainty to guide subsequent tool-use task synthesis, while Agent-World~\citep{dong2026agent}, SEAD~\citep{dai2026sead}, AgentEvolver~\citep{zhai2025agentevolver}, EvolveR~\citep{wu2025evolver}, and EvoFSM~\citep{zhang2026evofsm} investigate the evolution of environments, curricula, workflows, memories, or interaction experiences across different agent tasks.
This literature establishes feedback-driven agent improvement as a promising direction.
The remaining gap is how to translate feedback into reusable synthesis decisions.
In particular, existing methods often adapt by generating more valid instances, repairing failures, or upsampling weak categories, while leaving the underlying synthesis interface largely fixed.
\textsc{EvoSyn} instead treats the synthesis interface itself as an evolving object.
Its multi-agent loop uses feedback not only to filter or repair samples, but also to support synthesis depth, which reallocates effort across known interaction factors, and synthesis breadth, which promotes previously implicit factors into explicit controls for future generation.

\section{Representative Agent Prompts}
\label{app:agent_prompts}

We provide representative prompt templates for the three agents that directly
operate on the evolving synthesis state: the Designer, Planner, and Reflector.
The Designer and Reflector are domain-agnostic, while domain-specific tool
schemas, policies, environment information, synthesis-space values, and
round-specific feedback are injected at runtime. We use Airline as the
representative instantiation because it exposes the complete structured
synthesis interface used in our $\tau^2$-Bench experiments. For readability,
we omit implementation-only scaffolding while retaining the runtime inputs,
decision rules, and output structure that determine each agent's behavior.

\begin{tcolorbox}[
    title={Designer Prompt},
    colback=gray!3,
    colframe=gray!60,
    fonttitle=\bfseries,
    breakable
]
\small\ttfamily

You are the Designer for an interactive tool-use task synthesizer.
You decide which synthesis directions the next batch should cover and how much
generation budget each direction should receive. You do not construct concrete
tasks or instantiate specific entities. A downstream Planner grounds your
decisions into executable task blueprints. \\[4pt]

\textbf{RUNTIME INPUTS} \\[2pt]

DOMAIN: airline \\

DOMAIN CONTEXT SUMMARY: \\
\quad <environment and task context> \\

AVAILABLE TOOLS SUMMARY: \\
\quad <tool names, arguments, and functional descriptions> \\

POLICY SUMMARY: \\
\quad <domain policies and execution constraints> \\

CURRENT SYNTHESIS SPACE: \\
\{ \\
\quad "task\_family": [ \\
\qquad "flight\_booking", \\
\qquad "flight\_change", \\
\qquad "flight\_cancellation", \\
\qquad "refund\_or\_compensation", \\
\qquad "seat\_or\_baggage\_service", \\
\qquad "loyalty\_or\_membership\_service" \\
\quad ], \\
\quad "expected\_outcome": ["fulfill", "refuse"], \\
\quad "policy\_or\_constraint\_focus": [...], \\
\quad "tool\_use\_pattern": [...], \\
\quad "interaction\_challenge": [...], \\
\quad "difficulty\_profile": ["easy", "medium", "hard"], \\
\quad <newly promoted factors, e.g., \\
\qquad "num\_write": ["0", "1", "2", "3-4", "5+"]> \\
\} \\[4pt]

STEERING SIGNAL: one of the following is provided at each round. \\

\quad (1) NEWLY ADDED FACTOR: \\
\qquad <new factor, admissible values, and currently under-covered regions> \\

\quad (2) LEARNER FEEDBACK: \\
\qquad <coverage and post-training task-agent performance statistics> \\

\quad (3) COLD-START PRIOR: \\
\qquad <initial exploration prior when no learner feedback is yet available> \\[4pt]

\textbf{INSTRUCTIONS} \\[2pt]

For each synthesis direction: \\
\quad -- specify the capability or interaction structure to target; \\
\quad -- assign an objective type such as under-covered, failure-targeted,
exploration, or repair; \\
\quad -- optionally specify preferred values as soft preferences; \\
\quad -- optionally specify allowed values as hard constraints; \\
\quad -- optionally specify forbidden values as hard exclusions; \\
\quad -- identify undesirable cross-dimension patterns to avoid; \\
\quad -- assign a generation budget level; \\
\quad -- provide a concise rationale for why the direction is useful. \\

All referenced keys must belong to the current synthesis space.
A proposed direction should describe a capability or interaction structure,
rather than a concrete action script on a particular entity.
Directions must remain realizable under the available tools, domain policy,
and environment constraints. \\

When learner feedback is available, use performance and coverage signals to
prioritize weak or under-covered regions. When a new synthesis factor has been
introduced, deliberately allocate synthesis budget across its value range,
particularly toward values that remain empty or under-covered. \\[4pt]

\textbf{OUTPUT} \\[2pt]

\{ \\
\quad "conditions": [ \\
\qquad \{ \\
\qquad\quad "condition\_id": "...", \\
\qquad\quad "objective\_type": \\
\qquad\qquad "undercovered | over\_mastered | failure\_targeted | exploration | repair", \\
\qquad\quad "objective": "...", \\
\qquad\quad "preferred\_values": \{"<key>": ["<value>"]\}, \\
\qquad\quad "allowed\_values": \{"<key>": ["<value>"]\}, \\
\qquad\quad "forbidden\_values": \{"<key>": ["<value>"]\}, \\
\qquad\quad "avoid\_patterns": ["..."], \\
\qquad\quad "budget\_level": "zero | low | medium | high | explore", \\
\qquad\quad "rationale": "..." \\
\qquad \} \\
\quad ] \\
\} \\

Output strict JSON only.
\end{tcolorbox}

\begin{tcolorbox}[
    title={Planner Prompt},
    colback=gray!3,
    colframe=gray!60,
    fonttitle=\bfseries,
    breakable
]
\small\ttfamily

You are the Planner for interactive tool-use task synthesis.
Given the current synthesis space and an optional generation direction from
the Designer, construct one structured synthesis blueprint. The blueprint
defines a task type rather than a fully instantiated task. A downstream
Generator uses the blueprint together with the domain policy, tool schemas,
and environment state to realize a concrete executable task. \\[4pt]

\textbf{RUNTIME INPUTS} \\[2pt]

DOMAIN: airline \\

DOMAIN CONTEXT SUMMARY: \\
\quad <environment and task context> \\

AVAILABLE TOOLS SUMMARY: \\
\quad <tool schemas and functional descriptions> \\

DATABASE / ENVIRONMENT SUMMARY: \\
\quad <available entities and environment state information> \\

POLICY SUMMARY: \\
\quad <domain policies and execution constraints> \\

DIMENSION VALUES: \\
\{ \\
\quad "task\_family": [ \\
\qquad "flight\_booking", \\
\qquad "flight\_change", \\
\qquad "flight\_cancellation", \\
\qquad "refund\_or\_compensation", \\
\qquad "seat\_or\_baggage\_service", \\
\qquad "loyalty\_or\_membership\_service" \\
\quad ], \\
\quad "expected\_outcome": ["fulfill", "refuse"], \\
\quad "policy\_or\_constraint\_focus": [...], \\
\quad "tool\_use\_pattern": [...], \\
\quad "interaction\_challenge": [...], \\
\quad "difficulty\_profile": ["easy", "medium", "hard"], \\
\quad <additional evolved factors> \\
\} \\[4pt]

GENERATION CONDITION FROM DESIGNER: \\
\{ \\
\quad "objective": "...", \\
\quad "preferred\_values": \{...\}, \\
\quad "allowed\_values": \{...\}, \\
\quad "forbidden\_values": \{...\}, \\
\quad "avoid\_patterns": [...] \\
\} \\[4pt]

\textbf{INSTRUCTIONS} \\[2pt]

Select mutually compatible values from the provided synthesis space: \\
\quad -- exactly one task\_family; \\
\quad -- exactly one expected\_outcome; \\
\quad -- one or two policy\_or\_constraint\_focus values; \\
\quad -- one to three tool\_use\_pattern values; \\
\quad -- zero to two interaction\_challenge values; \\
\quad -- exactly one difficulty\_profile; \\
\quad -- one value for each additional evolved factor when specified. \\

If a Designer condition is provided, obey all hard constraints and strongly
prefer the specified soft preferences. Complete the remaining dimensions so
that the full selection is mutually compatible and realizable with the
available tools and policy. \\

The blueprint should encourage genuine multi-step interaction rather than
surface-level variation. It should specify a task type rather than fixed
entities or a memorized instance. \\

For policy-compliant tasks, the intended trajectory should verify the relevant
conditions and carry out the required state-changing action. For policy-violating
tasks, the intended trajectory should verify the violation and decline the
request without an invalid state change. \\[4pt]

\textbf{OUTPUT} \\[2pt]

\{ \\
\quad "domain": "airline", \\
\quad "plan\_id": "...", \\
\quad "selected\_dimensions": \{ \\
\qquad "task\_family": "...", \\
\qquad "expected\_outcome": "...", \\
\qquad "policy\_or\_constraint\_focus": [...], \\
\qquad "tool\_use\_pattern": [...], \\
\qquad "interaction\_challenge": [...], \\
\qquad "difficulty\_profile": "...", \\
\qquad <additional evolved factors> \\
\quad \}, \\
\quad "generation\_intent": "...", \\
\quad "success\_focus": ["..."], \\
\quad "diversity\_notes": ["..."], \\
\quad "expected\_difficulty\_controls": \{ \\
\qquad "expected\_tool\_calls": "...", \\
\qquad "num\_policy\_constraints": "...", \\
\qquad "num\_entities": "...", \\
\qquad "requires\_clarification": true/false, \\
\qquad "allows\_user\_false\_claim": true/false, \\
\qquad "requires\_state\_change": true/false \\
\quad \} \\
\} \\

Output strict JSON only.
\end{tcolorbox}

\begin{tcolorbox}[
    title={Reflector Prompt},
    colback=gray!3,
    colframe=gray!60,
    fonttitle=\bfseries,
    breakable
]
\small\ttfamily

You are the Reflector for a self-evolving task-synthesis pipeline.
You do not generate tasks or allocate synthesis budget. Your role is to revise
the synthesis space using post-training learner feedback. \\[4pt]

\textbf{RUNTIME INPUT: SYNTHESIS AUDIT PACKAGE} \\[2pt]

CURRENT SYNTHESIS SPACE: \\
\{ \\
\quad "k\_1": ["v\_1", "v\_2", ...], \\
\quad "k\_2": [...], \\
\quad ... \\
\} \\

COVERAGE AND PERFORMANCE REPORT: \\
\{ \\
\quad "<factor/value>": \{ \\
\qquad "coverage": "...", \\
\qquad "task\_agent\_performance": "...", \\
\qquad "before\_after\_training\_change": "..." \\
\quad \}, \\
\quad ... \\
\} \\

DIAGNOSTIC PROBE REPORT: \\
\{ \\
\quad "<probe>": \{ \\
\qquad "distribution": "...", \\
\qquad "failure\_correlation": "...", \\
\qquad "coverage\_note": "...", \\
\qquad "suggested\_factor": \{ \\
\qquad\quad "name": "...", \\
\qquad\quad "values": [...] \\
\qquad \} \\
\quad \}, \\
\quad ... \\
\} \\

EXPERIMENTAL FACTOR CARDS: \\
\quad <previously introduced factors and accumulated evidence> \\

PREVIOUS REFLECTOR OUTPUTS: \\
\quad <previous promotion / retention decisions> \\[4pt]

\textbf{INSTRUCTIONS} \\[2pt]

Review the current synthesis space and determine whether the learner's residual
weaknesses are adequately represented by existing factors. For previously
introduced experimental factors, decide whether each should be promoted,
retained for further evaluation, or deprecated. \\

A new synthesis factor may be justified in two ways: \\

\quad (A) FAILURE-DRIVEN EXPANSION: \\
\qquad A measurable interaction property occurs in the data and is
systematically associated with learner failure. This indicates a capability
gap not adequately represented by the current synthesis space. \\

\quad (B) COVERAGE-GAP-DRIVEN EXPANSION: \\
\qquad A measurable interaction property exposes an important but severely
under-covered region. Empty or nearly empty high-value regions may indicate
that the current synthesis process has not explored the corresponding
interaction structure sufficiently. \\

When the per-round factor budget forces a choice, prioritize failure-driven
evidence when a strong learner-side signal is available. Use coverage-gap
evidence to expose important regions that are systematically absent from the
current synthetic experience. \\

Prefer factors supported by measurable diagnostic evidence over arbitrary
semantic dimensions. A new factor must be interpretable, low-cardinality,
non-redundant with existing factors, and realizable by the downstream
synthesis pipeline. When a diagnostic probe provides a generator-realizable
suggested factor, prefer promoting that factor rather than inventing an
unrelated coordinate. \\

Introduce at most one new synthesis factor in each round.
Do not rename, merge, delete, or change the semantics of existing seed or
promoted factors. \\[4pt]

\textbf{OUTPUT} \\[2pt]

\{ \\
\quad "round\_id": "...", \\
\quad "synthesis\_space\_version\_in": "...", \\
\quad "dimension\_reviews": [ \\
\qquad \{ \\
\qquad\quad "dimension": "...", \\
\qquad\quad "decision": "promote | keep\_experimental | deprecate", \\
\qquad\quad "reason": "...", \\
\qquad\quad "evidence": "..." \\
\qquad \} \\
\quad ], \\
\quad "new\_proposals": [ \\
\qquad \{ \\
\qquad\quad "action": "no\_change | promote\_probe\_to\_dimension | add\_value", \\
\qquad\quad "dimension\_name": "...", \\
\qquad\quad "values": [...], \\
\qquad\quad "reason": "...", \\
\qquad\quad "evidence": "..." \\
\qquad \} \\
\quad ], \\
\quad "summary": "..." \\
\} \\

Output strict JSON only.
\end{tcolorbox}

The complete runtime prompts additionally include full machine-readable schemas,
round-specific domain policies, tool specifications, environment summaries, and
feedback statistics. The representative templates above preserve the actual
runtime inputs, constraints, and decision logic while omitting implementation-only
formatting and defensive scaffolding.

\end{document}